\documentclass{article}

     \PassOptionsToPackage{numbers, compress}{natbib}

\usepackage[preprint]{style}

\usepackage[utf8]{inputenc} 
\usepackage[T1]{fontenc}    
\usepackage{hyperref}       
\usepackage{url}            
\usepackage{booktabs}       
\usepackage{amsfonts}       
\usepackage{nicefrac}       
\usepackage{microtype}      
\usepackage{xcolor}         

\usepackage{graphicx}
\usepackage{caption}
\usepackage{subcaption}

\usepackage{footnote}
\makesavenoteenv{figure}
\makesavenoteenv{table}
\makesavenoteenv{subfigure}

\usepackage{enumitem}

\usepackage{tabularx}
\usepackage{adjustbox}

\usepackage{amsmath}
\usepackage{amssymb}
\usepackage{mathtools}
\usepackage{amsthm}

\theoremstyle{plain}
\newtheorem{theorem}{Theorem}[section]

\theoremstyle{definition}

\theoremstyle{remark}

\newcommand{\ys}[1]{\textcolor{black}{#1}}

\title{Model Collapse in the Self-Consuming Chain of Diffusion Finetuning: A Novel Perspective from Quantitative Trait Modeling}

\author{%
  Youngseok Yoon \\
  UCSB \\
  California, USA \\
  \texttt{youngseok\_yoon@ucsb.edu} \\
  \And
  Dainong Hu \\
  UCSB \\
  California, USA \\
  \texttt{dainong@ucsb.edu} \\
  \And
  Iain Weissburg \\
  UCSB \\
  California, USA \\
  \texttt{ixw@ucsb.edu} \\
  \AND
  Yao Qin \\
  UCSB, Google Deepmind \\
  California, USA \\
  \texttt{yaoqin@ucsb.edu} \\
  \And
  Haewon Jeong \\
  UCSB \\
  California, USA \\
  \texttt{haewon@ucsb.edu} \\
}

\begin{document}

\maketitle

\begin{abstract}
  \textit{Model collapse}, the severe degradation of generative models when iteratively trained on their own outputs, has gained significant attention in recent years. This paper examines Chain of Diffusion, where a pretrained text-to-image diffusion model is finetuned on its own generated images. We demonstrate that severe image quality degradation was universal and identify \textit{CFG scale} as the key factor impacting this model collapse. Drawing on an analogy between the Chain of Diffusion and biological evolution, we then introduce a novel theoretical analysis based on quantitative trait modeling from statistical genetics. Our theoretical analysis aligns with empirical observations of the generated images in the Chain of Diffusion. Finally, we propose Reusable Diffusion Finetuning (ReDiFine), a simple yet effective strategy inspired by genetic mutations. It operates robustly across various scenarios without requiring any hyperparameter tuning, making it a plug-and-play solution for reusable image generation.
\end{abstract}

\section{Introduction}
\label{section:introduction}

\emph{Can state-of-the-art AI models learn from their own outputs and improve themselves?} As generative AI models (e.g., GPT, Diffusion) now churn out uncountable synthetic texts and images, this question piqued curiosity from many researchers in the past couple of years. While some show positive results of self-improving~\cite{huang2022large,gerstgrasser2024model}, most report an undesirable \emph{``model collapse''}—a phenomenon where a model's performance degrades when it goes through multiple cycles of training with the self-generated data~\cite{bertrand2023stability,gillman2024,taori2023data,shumailov2023curse,dohmatob_model_2024,fu2024towards,marchi2024heat,martinez2023towards}. When large language models (LLMs) are trained with their own outputs, it begins to produce low-quality text that has a lot of repetitions~\cite{dohmatob2024tale}, and its linguistic diversity declines rapidly~\cite{guo_curious_2024,briesch2023large}; image models also show quality degradation~\cite{bohacek2023nepotistically,martinez2023combining} and loss of diversity~\cite{alemohammad2023self,hataya_will_2023}.

The goal of this paper is to investigate model collapse in the practical scenario of finetuning pretrained text-to-image diffusion models. End users often finetune the latest model to generate images with a specific style (e.g., creating characters in the style of Pokémon). In fact, hundreds of new finetuned models trained on different styles and tons of generated images are regularly uploaded on platforms like CivitAI\footnote{\url{https://civitai.com/}}. When users scrape the Internet to collect images in the style they want, it becomes almost inevitable that synthetic images will be included in their datasets.

To this end, we conduct a thorough empirical investigation into how various hyperparameters commonly used during diffusion finetuning (e.g., learning rate, diffusion steps, prompts) impact model collapse. We make a crucial observation: the classifier-free guidance (CFG) scale is the most impactful factor affecting the rate of model collapse. Moreover, we observe a fascinating phenomenon that the direction of degradation varies---low CFG results in low-frequency degradation in images, while high CFG leads to high-frequency degradation. The critical role of CFG in both the rate and direction of model collapse is a novel insight previously unrecognized in the literature.

To gain a deeper theoretical understanding, we draw upon the concept of quantitative trait modeling from statistical genetics. Unlike existing studies that attribute model collapse to limited sample sizes and conclude it leads to zero variance~\cite{bertrand2023stability, shumailov2023curse, alemohammad2023self}, we show that a truncation-based selection results in mean drift of quantitative features. \ys{We found the high-frequency power spectra of generated images as one of the features that accurately follow our theoretical modeling.} Moreover, we show that other theoretical work on model collapse can also be connected to statistical genetics models, such as random genetic drift and its variants~\cite{fisher1999genetical,wright1931evolution,kimura1955solution,paris2019inference}. This fresh angle of drawing parallels with statistical genetics offers a new framework to understand the mechanism of model collapse.

While we have shown the crucial role of CFG in mitigating model collapse both empirically and theoretically, searching for the optimal CFG scale over iterations is computationally expensive and cumbersome to implement in practice. We propose a simple solution that bypasses CFG tuning, called Reusable Diffusion Finetuning (ReDiFine). It introduces small modifications in the finetuning and generation—condition drop and CFG scheduling—and we demonstrate that ReDiFine achieves performance comparable to the optimal CFG scale across all hyperparameter settings we tested.

Our contributions can be summarized as follow: 
\begin{itemize}[left=0cm]
    \item We conducted a comprehensive investigation of model collapse when finetuning a diffusion model on its own outputs, testing a wide range of parameters on four datasets (two digital art and two natural images). Our results show that CFG scale is the most critical factor, governing both the rate of model collapse and the type of image degradation (Section~\ref{section:observation}).
    \item We provide a novel theoretical analysis of model collapse based on quantitative trait modeling that can accurately predict how power spectra of generated images evolve over iterations (Section~\ref{section:theory}). 
    \item We propose ReDiFine, a simple yet effective strategy to achieve a near-optimal reusability-fidelity trade-off without any hyperparameter tuning. By combining condition drop finetuning and CFG scheduling, ReDiFine operates robustly across various scenarios—such as synthetic-real mixed datasets and varying initial CFG settings—across all four datasets we evaluated (Section~\ref{section:experiment}).
\end{itemize}

\section{Related Work}
\label{section:related}

The self-consuming training loop and the associated phenomenon known as ``model collapse'' have become significant areas of study in the past two years~\cite{bertrand2023stability, gillman2024, taori2023data, dohmatob_model_2024, fu2024towards, marchi2024heat, martinez2023towards, guo_curious_2024, briesch2023large, bohacek2023nepotistically, martinez2023combining, alemohammad2023self}. Model collapse, defined as ``a degenerative process affecting generations of learned generative models, where generated data end up polluting the training set of the next generation of models'' in \cite{shumailov2023curse}, has been observed in both language and image generative models\footnote{The definition of model collapse is expanding~\cite{stein2024exposing}, but we focus on a fully self-consuming training loop~\cite{alemohammad2023self}.}.

Empirical studies on LLMs~\cite{briesch2023large} reveal that linguistic diversity collapses, especially in high-entropy tasks~\cite{guo_curious_2024}, although this can be mitigated with data accumulation~\cite{gerstgrasser2024model}. In image generation, several works~\cite{bertrand2023stability, martinez2023towards, bohacek2023nepotistically, martinez2023combining, alemohammad2023self, hataya_will_2023} note image degradation when diffusion models are recursively trained with self-generated data. We conduct extensive experiments to reveal the key factor causing model collapse in text-to-image diffusion models. Our findings reveal that while model collapse is universally observed across various datasets and scenarios, it manifests in three distinct types of image degradation.

Theoretical studies on model collapse largely focused on diversity reduction, typically framed as either decreasing covariance in continuous domains~\cite{bertrand2023stability, alemohammad2023self, shumailov2024ai} or shrinking support in discrete domains~\cite{marchi2024heat, dohmatob2024tale}, which aligns with our observations on decreased recall. The finite number of generated samples has been identified as a key cause of model collapse~\cite{bertrand2023stability, fu2024towards, dohmatob2024tale, shumailov2024ai}, while others~\cite{marchi2024heat, alemohammad2023self, ferbach2024self} emphasize that sampling bias reduces the effective distribution. \ys{There also exists the study about recursive stability of generative models in the self-consuming training loops~\cite{fu2025theoretical}.} In contrast, we present a novel framework based on quantitative trait modeling, providing a fresh direction based on mean drift induced by the selection process. We further demonstrate that many existing theories of model collapse can be understood through extensions of classical statistical genetics.

Most existing works echo the importance of incorporating a large proportion of real data~\cite{bertrand2023stability, fu2024towards, alemohammad2023self, ferbach2024self, fu2025theoretical} or accumulating data over iterations~\cite{gerstgrasser2024model} to mitigate model collapse. \cite{gillman2024} instead suggests a self-correcting self-consuming loop using an expert model, a physics simulator for human motion generation, to correct synthetic outputs. Several concurrent works seek to address model collapse through their theoretical insights. \ys{\cite{dohmatob2024strong} studies the effects of synthetic data in scaling laws, showing how the scales of models affect model collapse behaviors.} Moreover, \cite{feng2024beyond} and \cite{amin2025escaping} emphasize the importance of data verification in preventing performance degradation, while \cite{zhu2024synthesize} introduces a novel token editing strategy for data curation in LLM tasks. On the other hand, \cite{zhu2024analyzing} and \cite{alemohammad2024self} propose model-specific solutions leveraging real data to counteract the negative effects of synthetic data in image-generative models. In our work, we propose an alternative solution through \emph{reusable image generation}, and show that it can generate high-quality images comparable to the optimal hyperparameter without extensive search.

\section{Model Collapse in Self-Consuming Chain of Diffusion Finetuning}
\label{section:observation}

\begin{figure*}[t!]
    \centering
    \begin{subfigure}[t]{0.51\textwidth}
        \centering
        \includegraphics[width=1.0\textwidth]{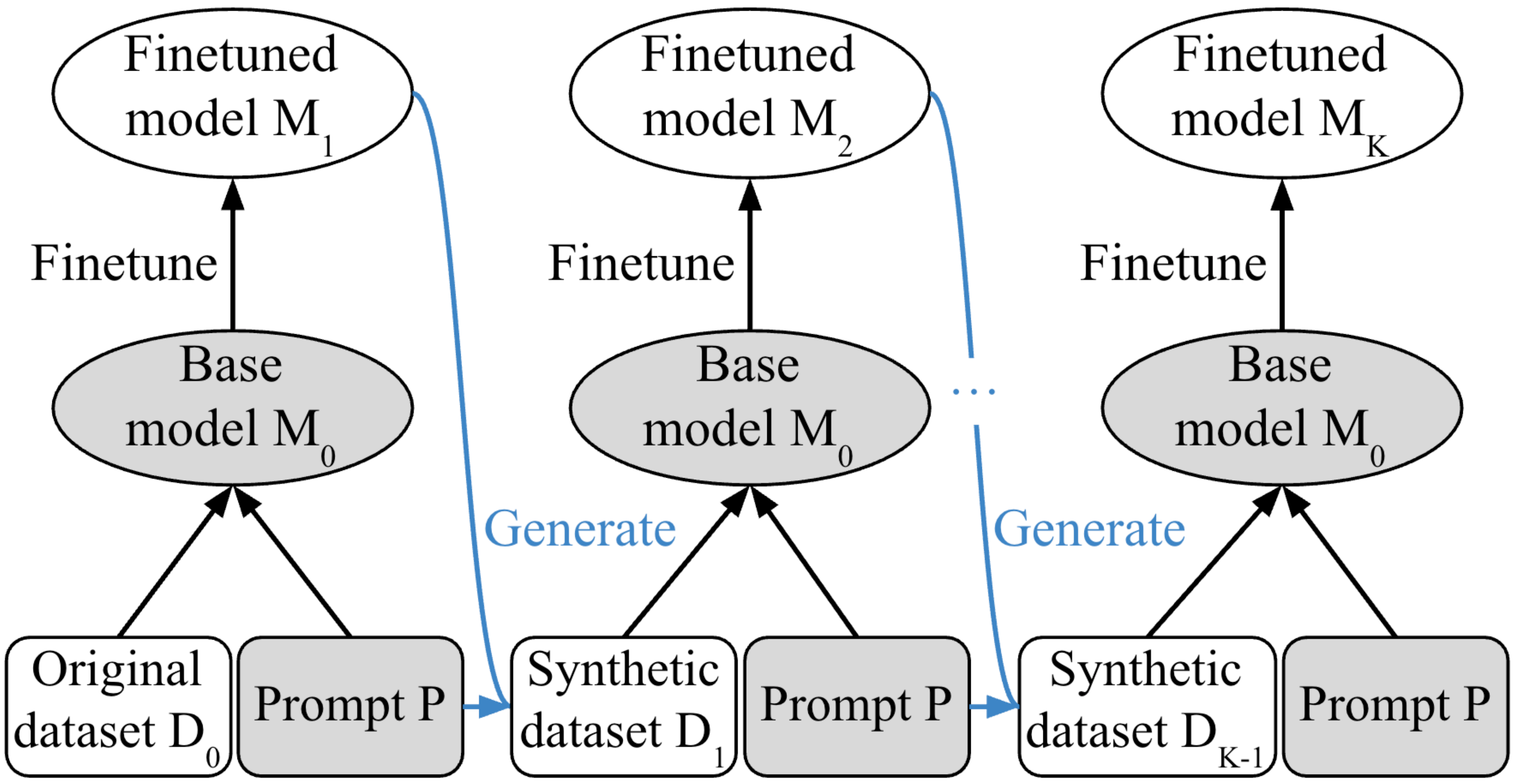}
        \subcaption{
        Self-Consuming Chain of Diffusion Finetuning.
        }
        \label{figure:diagram}
    \end{subfigure}
    \qquad
    \centering
    \begin{subfigure}[t]{0.43\textwidth}
        \centering
        \includegraphics[width=0.85\textwidth]{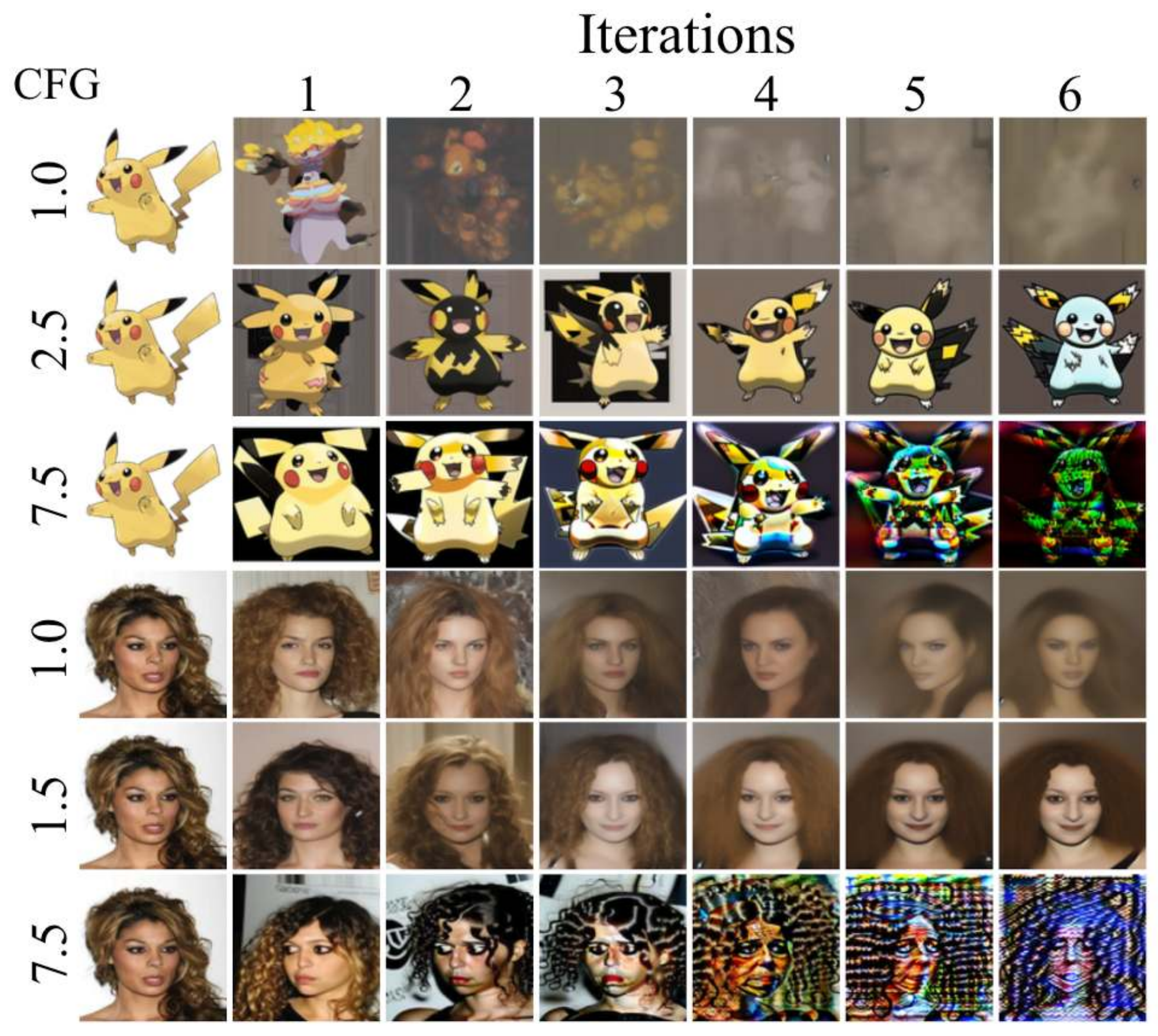}
        \subcaption{
        Image degradation in the Chain of Diffusion.
        }
        \label{figure:baseline_datasets}
    \end{subfigure}
    \vspace{-0.5em}
    \caption{
    (a) 
    \textbf{Overall pipeline of the Chain of Diffusion.}
    Given a pretrained text-to-image diffusion model $M_0$ and a prompt set $P$, a finetuned model $M_k$ is trained using $D_{k-1}$ generated at the previous iteration $k-1$.
    Then, $M_k$ generates $D_k$ using the same prompt set $P$, building a fully synthetic loop.
    Chain of diffusion begins with the original real dataset $D_0$.
    (b)
    \textbf{Image degradation universally occurs across multiple domains, and low and high CFG scales lead to blurry images and high-frequency degradation, respectively.} As the Chain of Diffusion progresses, the severity of image degradation intensifies, which holds consistently across four datasets and $11$ scenarios in Section~\ref{subsection:analysis}. CFG $2.5$ for Pokemon and $1.5$ for CelebA exhibit an ideal middle ground where both types of degradation slow down. More images in Appendix~\ref{appendix:baseline_experiments}.
    }
    \label{figure:introduction}
    \vspace{-1em}
\end{figure*}

\subsection{Problem setting \& Experimental Setup}

\textbf{Chain of Diffusion.} We begin with formally defining the self-consuming Chain of Diffusion finetuning. Given a pretrained generative model $M_0$, an original training image set $D_0 = \{ x_{0, i} | i \in [0, N-1] \}$, and a prompt set $P = \{ y_{i} | i \in [0, N-1] \}$, where $N$ is the number of total images in the dataset, each image $x_{0, i}$ is paired with a corresponding text prompt $y_i$. $M_{k+1}$ is a model finetuned from $M_0$ using the generated image set $D_k = \{ x_{k, i} | i \in [0, N-1] \}$ and the prompt set $P$, which simulates a fully synthetic loop~\cite{alemohammad2023self}. Then, $M_{k+1}$ generates a set of images $D_{k+1}$ for the next iteration using the prompt set $P$:
\begin{equation}
    M_{k+1} = \text{Finetune}(M_0, D_k, P),
\end{equation}
\begin{equation}
    D_{k+1} = \text{Generate}(M_{k+1}, P).
\end{equation}
During the Chain of Diffusion, $M_0$ and $P$ are fixed across iterations, and one image is generated per one prompt to maintain the dataset size for all iterations. The overall pipeline is shown in Figure~\ref{figure:diagram}. We note that this setting is consistent with \cite{shumailov2023curse, fu2024towards, guo_curious_2024, briesch2023large, alemohammad2023self}.

\textbf{Model and datasets.} We use Stable Diffusion v1.5~\cite{rombach2022high} as the pretrained model $M_0$ and apply LoRA \cite{hu2021lora} to finetune $M_0$ at each iteration. We build our implementation on \cite{kohya_ss} and experiment on four datasets: Pokemon~\cite{pokedex}, Kumapi~\cite{kaggle-illustrations-kumapi390}, Butterfly~\cite{butterflydataset}, and CelebA-1k~\cite{liu2015faceattributes} to investigate various domains including animation, handwriting, and real pictures. $M_0$ is finetuned for 100 epochs during each iteration. More details can be found in Appendix~\ref{appendix:experimental_setting:hyperparameters} and \ref{appendix:experimental_setting:datasets}.

\textbf{Evaluation metrics.} We use Frechet Inception Distance (FID)~\cite{heusel2017gans} to measure image quality. Following \cite{stein2024exposing}, we use DiNOv2~\cite{oquab2023dinov2} as a feature extractor since it is more consistent with our visual inspection than Inception-V3 network~\cite{szegedy2016rethinking}. With a slight abuse of terms, we will still refer to the Frechet distance with DiNOv2 as the FID score. Additional experiments with other metrics (CLIP score~\cite{radford2021learning}, Sample-wise Feature Distance (SFD), and recall) are presented in Appendix~\ref{appendix:method:quantitative}.

We propose a new metric to quantify the degradation of generated images. We define collapse rate as the performance degeneration per iteration in the Chain of Diffusion:
\begin{equation}\label{equation:reusability}
    \text{collapse rate} = \Delta {\rm FID} = \frac{{\rm FID}_K - {\rm FID}_1}{K-1},
\end{equation}
where ${\rm FID}_k$ stands for the FID between $k$-th iteration set and the original training set. We use FID as a performance metric here, but this can be any other performance metric of interest. Note that a low collapse rate indicates more reusable images since the model does not degrade much. We have $K = 6$ for the rest of the paper.


\begin{figure}[t!]
    \centering
    \begin{subfigure}[b]{0.7\textwidth}
        \centering
        \includegraphics[width=1.0\textwidth]{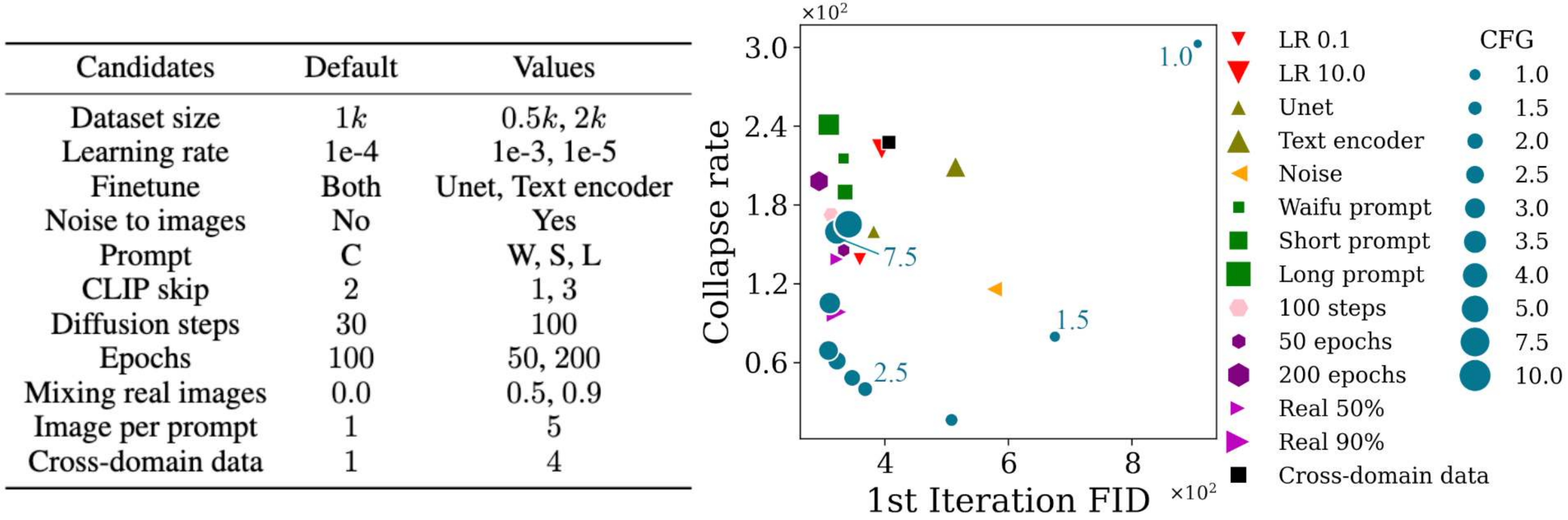}    
    \end{subfigure}
    \begin{subfigure}[b]{0.25\textwidth}
        \centering
        \includegraphics[width=1.0\textwidth]{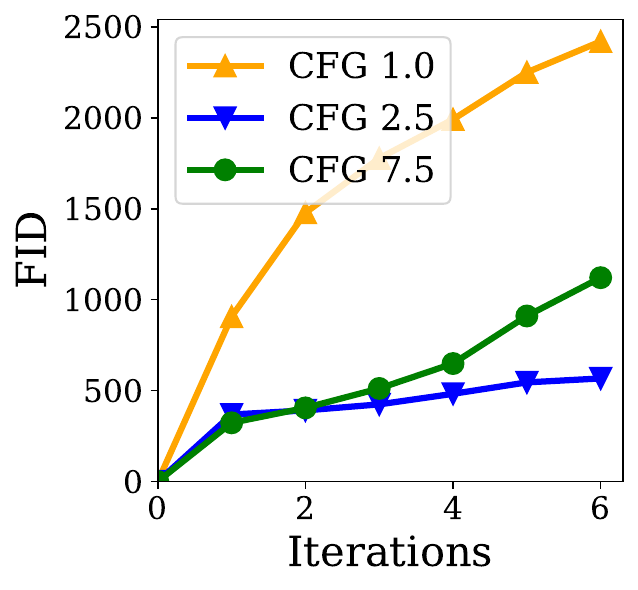}
    \end{subfigure}
    \vspace{-0.5em}
    \caption{\textbf{Left:} Description of $11$ potential factors (excluding CFG) that we examined as candidate sources for model collapse. All experiments were conducted on Pokemon except for the dataset size. For dataset size, we use CelebA since its original dataset is bigger, and we can subsample $500$, $1000$, and $2000$ images. For prompt, we concatenate prompts with different lengths. \textbf{Center:} All hyperparameter settings other than changing CFG show a high \emph{collapse rate} greater than 1.0, i.e., FID score increases by $\sim$2x in 6 iterations, indicating severe image degradation. The x-axis represents $\text{FID}_1$, quantifying the generation performance at the first iteration and the y-axis represents the \emph{collapse rate} defined in Eqn.~\ref{equation:reusability}. For both $\text{FID}_1$ and \emph{collapse rate}, lower is better. \textbf{Right:} Quantitative comparison for FID $\downarrow$ (Pokemon). CFG $2.5$ achieves the most robust performance. CFG $1.0$ degrades from the beginning of the chain while CFG $7.5$ begins to degrade in the third iteration, which aligns with the visual inspection in Figure~\ref{figure:baseline_datasets}.}
    \vspace{-1em}
    \label{figure:tradeoff_with_table}
\end{figure}

\subsection{Model Collapse in the Chain of Diffusion}
\label{subsection:analysis}

In this section, we make a series of observations regarding the model collapse behavior in the Chain of Diffusion. We conduct extensive investigations to reveal the most impactful factor in the model collapse and analyze how this factor contributes to the Chain of Diffusion.

\textbf{Observation 1: Model collapse is universal in the Chain of Diffusion.} We observe significant image degradation in all four datasets in the Chain of Diffusion (see Figure~\ref{figure:baseline_datasets} and Appendix~\ref{appendix:baseline_experiments}). The quality begins to deteriorate in the third iteration and drops rapidly once the degradation emerges, reaching an unrecognizable level in two or three additional iterations. Quantitative evaluation (FID, CLIP score, SFD, and recall) also indicates this degradation (Appendix~\ref{appendix:method:quantitative}). We observed significant drops in recall in all experiments, indicating diversity reduction discussed in several recent works.

We then investigated a variety of different scenarios (summarized in Figure~\ref{figure:tradeoff_with_table}) to see if this degradation is an anomaly of specific hyperparameters or a ubiquitous phenomenon. We tested various dataset sizes for $D_0$ and $D_k$, increasing the size of $D_k$ by generating more than one image per prompt, mixing real images of $D_0$ to $D_k$, changing the descriptiveness of prompts\footnote{C, W, S, and L for Combine, Waifu, Short, and Long, respectively. We concatenate BLIP and Waifu captions as default setting, referred to as Combine. More details can be found in Appendix~\ref{appendix:hyperparameter_experiments:prompt}.}, freezing U-Net or text encoder, and various other hyperparameters (\# sampling steps, \# epochs, learning rate, and CLIP skip). We also added a small Gaussian noise in each image in the original set $D_0$ to test if small random perturbations can improve reusability, or merged all four datasets into one to see if the extended domain delays model collapse (Cross-domain data). \emph{In all settings we tested, image degradation was universally present and very fast.} We plot the trade-off for Pokemon on Figure~\ref{figure:tradeoff_with_table} where y-axis is collapse rate and x-axis is the ${\rm FID}_1$ (better when closer to the origin). While adding noise to images and mixing $90\%$ real images to every iteration as proposed by \cite{gerstgrasser2024model, bertrand2023stability, fu2024towards, alemohammad2023self, ferbach2024self} show the lowest collapse rate, they still exhibit significant degradation (FID score has been doubled in $6$ iterations)\footnote{We do not display scenarios that change the training dataset size, such as Img/Prompt and Dataset Size, as varying sizes differ FID scales. We observe similar image degradation for them, presented in the Appendix~\ref{appendix:hyperparameter_experiments}.}.

Moreover, this degradation is unavoidable for Stable Diffusion XL (Appendix~\ref{appendix:hyperparameter_experiments:sdxl}). We further examined different settings from the literature where all previously generated images are accumulated (data accumulation~\cite{gerstgrasser2024model}, Appendix~\ref{appendix:hyperparameter_experiments:iteration_accumulation}) or the same model is iteratively finetuned (iterative retraining~\cite{bertrand2023stability, martinez2023combining}, Appendix~\ref{appendix:method:iterative_retraining}), and consistently observed model collapse.

\begin{figure*}[t!]
    \centering
    \begin{subfigure}[b]{0.23\textwidth} 
        \centering
        \includegraphics[height=0.94\textwidth]{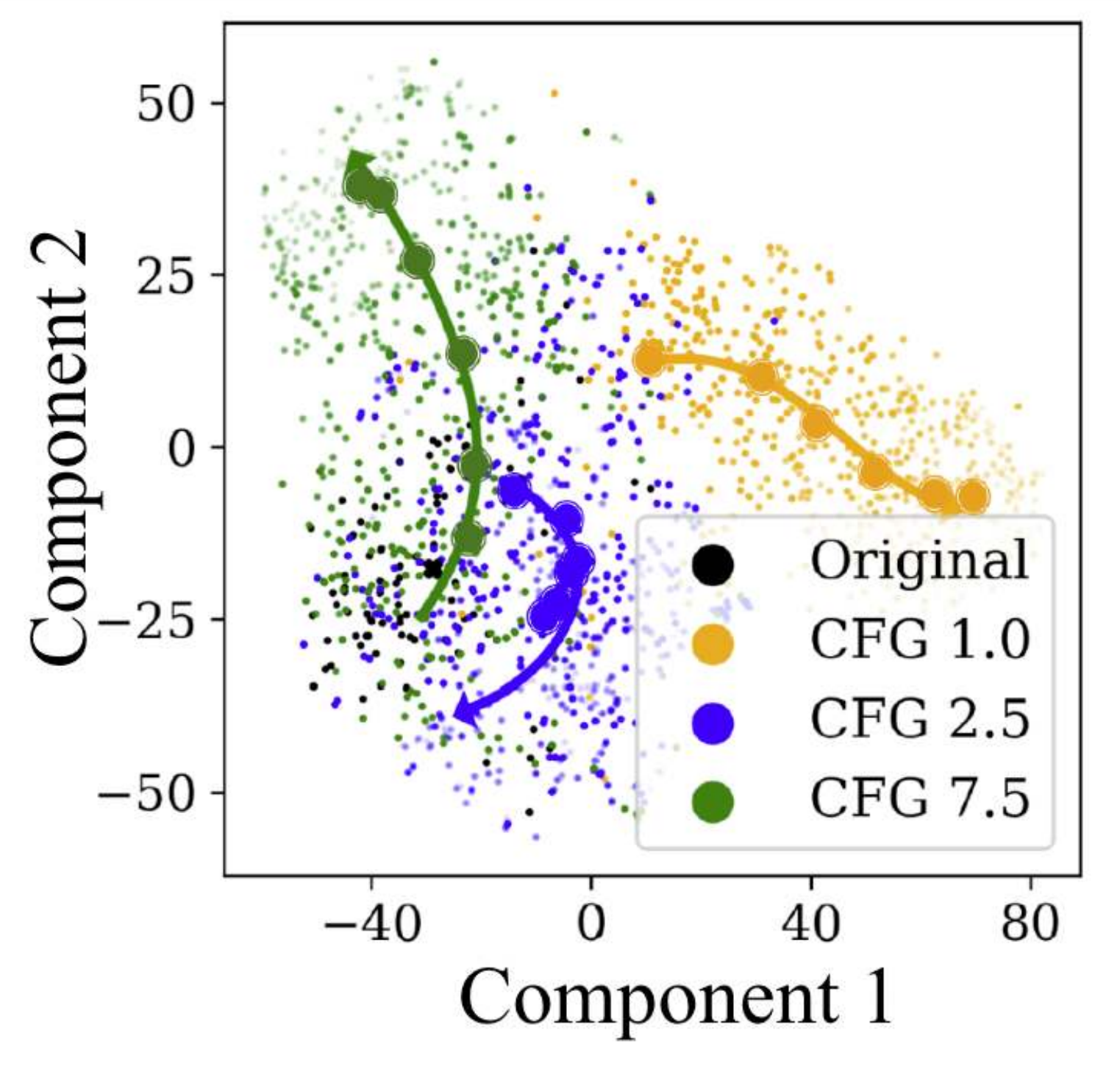}
        \subcaption{t-SNE of images.}
        \label{figure:tsne}
    \end{subfigure}
    \begin{subfigure}[b]{0.23\textwidth} 
        \centering
        \includegraphics[height=0.94\textwidth]{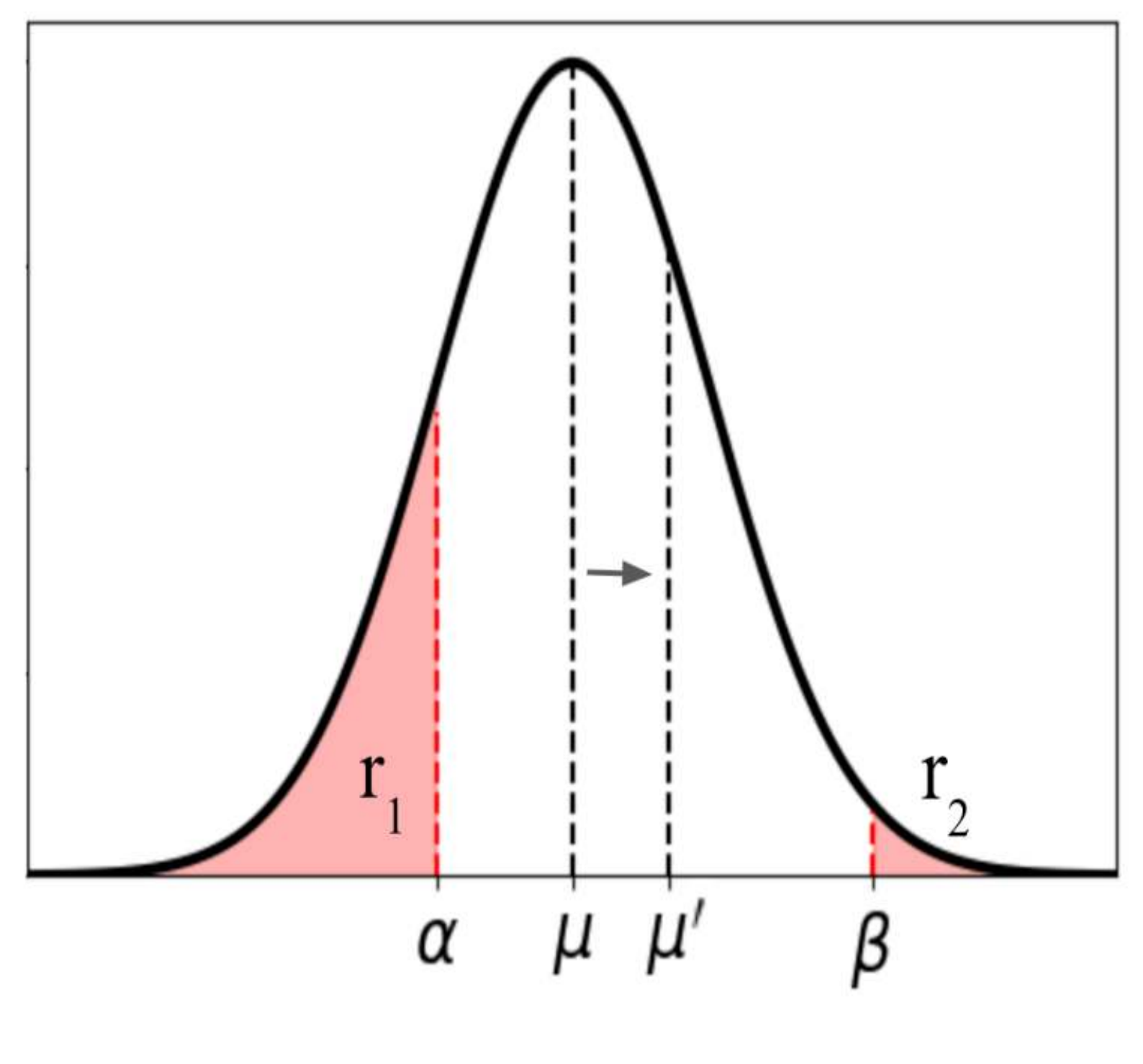}
        \subcaption{Selection mechanism.}
        \label{figure:simulation:truncation}
    \end{subfigure}
    \begin{subfigure}[b]{0.52\textwidth}
        \centering
        \includegraphics[width=0.94\textwidth]{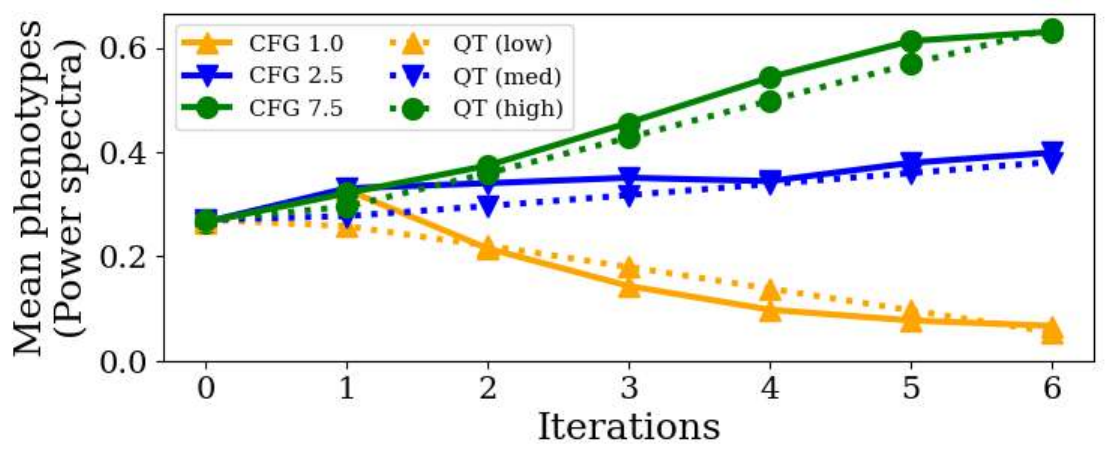}
        \subcaption{Power spectra of images vs. Mean of phenotypes.}
        \label{figure:simulation:mean}
    \end{subfigure}
    \vspace{-0.5em}
    \caption{
    (a) \textbf{t-SNE plot visualizes how generated images evolve from the original distribution (black) and shows distinct paths for three CFG scales (Pokemon).} Different CFG scales and iterations are differentiated with different colors and transparency. Arrows indicate how the distributions of generated images move for different CFG scales. While CFG 2.5 (blue) stays near the original images (black), high and low CFG scales (1.0 and 7.5) deviate fast, indicating image degradation.
    (b) \textbf{Selection mechanism with two-sided truncation.} $r_1$ and $r_2$ ratios of samples are truncated from the left and right tails, and the remaining $r = 1 - r_1 - r_2$ ratio of samples is selected. 
    (c) \textbf{Power spectra for different CFGs (Pokemon) and quantitative trait modeling results align well for different selection strategies.} 
    Directional selections with truncation can effectively explain our observations in Section~\ref{section:observation}: the behaviors of high- and low-frequency degradation and optimal CFG scale. Detailed settings are provided in Appendix~\ref{appendix:theory:setup}.
    }
    \label{figure:different_cfgs}
    \vspace{-1em}
\end{figure*}

\textbf{Observation 2: CFG is the most significant factor that impacts the model collapse.} Throughout all our experiments, classifier-free guidance (CFG) had the biggest impact on the speed of model collapse. CFG scale was first introduced in \cite{ho2022classifier} to modulate the balance between unconditional score and conditional score at each diffusion step as follows:
\begin{equation}
    \text{Total Score} = \text{Unconditional Score} \\ + \text{CFG} \cdot (\text{Conditional Score} - \text{Unconditional Score}).
\end{equation}
High CFG emphasizes the conditional score for the given prompt, which pushes the generation to align better with the prompt and often leads to higher-fidelity images. On the other hand, lower CFG places less weight on the conditional score and provides more diversity in generated images. For those familiar with temperature sampling~\cite{ackley1985learning}, CFG plays a similar role as temperature, which adjusts the trade-off between fidelity and diversity.

Figure~\ref{figure:tradeoff_with_table} shows that as we increase the CFG scale, the image quality in the first iteration improves (smaller ${\rm FID}_1$ on x-axis), which is expected from our understanding of CFG. Surprisingly, this comes at the cost of a worse collapse rate (an increase on the y-axis). When the CFG scale is as high as $7.5$ or $10.0$, the improvement in ${\rm FID}_1$ plateaus, and it worsens both ${\rm FID}_1$ and collapse rate. Similarly, when the CFG scale is too low---below $2.0$---the improvement in collapse rate plateaus, and both ${\rm FID}_1$ and collapse rate begin to increase. There is an optimal region of CFG values (near $2.5$, specific to Pokemon), where we achieve a low collapse rate while maintaining low ${\rm FID}_1$ as well. Figure~\ref{figure:tradeoff_with_table} presents FID for Pokemon to demonstrate how different CFG scales affect performance. CFG scale $2.5$ achieves the most robust performance for all metrics. Interestingly, optimal CFG scales differ for different styles: $2.5$ for animated or hand-writing datasets (Pokemon and Kumapi) and $1.5$ for photo datasets (CelebA and Butterfly) as shown in Appendix~\ref{appendix:baseline_experiments}.

\textbf{Observation 3: High CFG scales cause high-frequency degradation and low CFG scales cause low-frequency degradation.} CFG scale does not only affect the speed of model collapse, but also the pattern of model collapse. As shown in Figure~\ref{figure:baseline_datasets}, CFG $1.0$ makes the images progressively more blurry in the Chain of Diffusion, eventually collapsing to images without any structure, which we refer to low-frequency degradation. On the other hand, for CFG $7.5$ how images degrade looks completely different: some features start to be emphasized excessively, repetitive patterns begin to appear, and the overall color distribution becomes saturated. The t-SNE plot in Figure~\ref{figure:tsne} clearly demonstrates distinct paths for high, low, and medium CFG scales over iterations. They were consistent in all four datasets (Appendix~\ref{appendix:baseline_experiments}). In Section~\ref{section:theory}, we detail how different patterns of high-frequency power spectra from different CFG scales can be understood using the framework from genetic biology.

\noindent \textbf{Implications of our observations.} 
We showed that a high CFG of $7.5$, a common choice to generate visually appealing images, significantly increases collapse rate to achieve slightly better $\rm{FID}_1$. Sampling for maximizing the perceptual quality was coined as `sampling bias' in \cite{alemohammad2023self}. While they reported a monotonic increase in collapse rate as CFG increased from $1.0$ to $2.0$, we show that the holistic picture is not monotonic when we look CFG scales from $1.0$ to $10.0$. It shows an intriguing trade-off between perceptual quality and reusability, suggesting developers to consider reusability of images to achieve substantially improved future generations by carefully choosing CFG.

\section{Quantitative Trait Models for Fully-Synthetic Training Loops}
\label{section:theory}

This section introduces a novel perspective to understand model collapse in generative models by drawing parallels between genetic biology and the Chain of Diffusion. Distinguishable iterations in the Chain of Diffusion---where each iteration is separable with no duplicated individual, and the current iteration originates from the previous one---mirror genetic processes involving successive iterations of parents and offspring. We use the term ``iteration'' instead of ``generation'' for genetic generation to avoid confusion with image generation.

\ys{We begin by explaining how ``diversity collapse'' of language models in self-consuming training loops~\cite{marchi2024heat, dohmatob2024tale} could be explained by the classic Wright-Fisher model from statistical genetics. Then we introduce } Quantitative Trait modeling and its underlying mathematical assumptions. Based on them, we derive a theorem showing that the mean trait value exhibits linear drift, while the variance stabilizes over time. This model can successfully capture the three key behaviors observed in Section~\ref{section:observation}: high-frequency degradation, low-frequency degradation, and optimal CFG scale. It suggests that different CFG scales in the Chain of Diffusion can be viewed as varying selection strategies, with power spectra corresponding to quantitative traits.

\subsection{Connection to Statistical Genetics}

\ys{The Wright-Fisher (WF) model~\cite{wright1931evolution} describes how allele (genetic variants) frequencies evolve in a population over iterations, assuming no selection or mutation. In each iteration, the entire population is replaced, with the next iteration formed through random sampling with replacement---a process known as genetic drift. When the population size is finite, the WF model predicts that one allele will eventually reach fixation, while others are lost. We provide further explanations in the Appendix~\ref{appendix:theory:wf}.}

\ys{Language models (LMs) follow the same basic idea, with binary alleles replaced by a vocabulary of discrete words. LMs estimate word frequencies from a training corpus, and in self-consuming training loops, new texts are generated by the current LM and used to train the next. This acts analogously to genetic drift in large-scale, but finite, generations. Rare words may gradually disappear from the corpus---an effect that is irreversible. Recent works on diversity collapse in LMs~\cite{marchi2024heat, dohmatob2024tale} can be interpreted as fixation by drift, in line with the WF model. In contrast, image generation operates in a near-continuous space of pixel combinations (e.g., $(256^3)^{512 \times 512}$ for $512 \times 512$ RGB images), where no output is truly lost---unseen combinations can still be generated. In this setting, the process is typically modeled with Gaussian distributions~\cite{bertrand2023stability, shumailov2023curse, alemohammad2023self, shumailov2024ai}, and analogs of genetic drift and fixation emerge through the continuous variants of the WF model~\cite{tataru2017statistical}.}

\ys{Given its widespread use in genetics, the WF model has been extended to incorporate selection~\cite{he2017effects,kaj2024wright} or mutations~\cite{charlesworth2020long}, further enriching the analogy. These connections establish a strong conceptual correspondence between statistical genetics and model collapse---one that, to our knowledge, has been largely overlooked. Our work highlights this parallel and demonstrates how it can serve as a unifying theoretical perspective.}

\subsection{Quantitative Trait Modeling}

\ys{Prior works can be viewed through the lens of genetic drift, where the persistence or disappearance of certain word tokens is driven largely by chance. In contrast, we adopt a Quantitative Trait (QT) modeling, incorporating a form of natural selection that reflects human preferences, to explain the evolution of feature distributions and the onset of model collapse.} QT modeling in statistical genetics explores the evolution of quantitative phenotypes (e.g., height, weight, or color), which are influenced by many genetic and environmental factors. These traits are typically modeled as Gaussian-distributed variables and evolve across discrete iterations, where parent and offspring populations are distinguishable ($t$- and $t+1$-th iterations are clearly separable).

Let the distribution of phenotypes at the $t$-th iteration be denoted as $\mathcal{N} (\mu_t, \sigma_{P, t}^2)$ where $\mu_t$ and $\sigma_{P, t}^2$ are the mean and variance of quantitative phenotypes. The phenotypic variance $\sigma_{P, t}^2$ is the sum of the (additive) genetic variance $\sigma_{G, t}^2$ and the environmental variance $\sigma_E^2$, i.e., $\sigma_{P, t}^2 = \sigma_{G, t}^2 + \sigma_{E}^2$~\cite{falconer1996introduction}\footnote[5]{Genetic variance is composed of additive, dominance, and interaction variance. Here, we only consider additive variance, which is a common assumption in the field.}. When selection occurs, whether natural (e.g., faster animals surviving predators) or artificial (e.g., breeding livestock for higher milk production), it affects the distribution of the effective population that influences the next iteration. We consider directional selection with truncation as shown in Figure~\ref{figure:simulation:truncation}, where $r$ ratio of samples is selected by truncating $r_1$ from the left and $r_2$ from the right side of the distribution. Here, $r + r_1 + r_2 = 1$ and larger phenotypes are preferred when $r_1 > r_2$.

The (narrow-sense) heritability\footnote[6]{The heritability is narrow-sense when the genetic variance is restricted to additive variance.}, which is defined as the proportion of phenotypic variance attributable to additive genetic factors, can also be represented using the Breeder's Eqn.~\cite{lush2013animal} as:
\begin{equation}
    h_t^2 = \frac{\sigma_{G, t}^2}{\sigma_{P, t}^2} = \frac{\sigma_{G, t}^2}{\sigma_{G, t}^2 + \sigma_E^2} = \frac{\mu_{t+1} - \mu_{t}}{\mu_t' - \mu_t},
\end{equation}
where the mean phenotype of the next iteration $\mu_{t+1}$ is represented using the mean phenotype of selected individuals $\mu_t'$, the mean phenotype of the entire population  $\mu_t$, and heritability $h_t^2$. We assume the genetic variance for the next iteration is determined by the variance of selected individuals $\sigma_{G, t+1}^2 = \sigma_{P, t}'^2$. We prove the behaviors of mean and variance of phenotypes under this setting:
\begin{theorem}\label{theorem:convergence}
    Suppose the distributions of phenotypes follow Gaussian distribution and directional selection truncates individuals on both sides with ratios $r_1$ and $r_2$. 
    Then mean of phenotypes asymptotically increases (decreases) by $\frac{c_1 c_2}{\sqrt{1 - c_2}} \sigma_{E}$ per iteration
    and the variance converges to $\frac{1}{1 - c_2} \sigma_E^2$ when $r_1 > r_2$ ($r_2 < r_1$), where $c_1$ and $c_2$ are constants that depend on $r_1$ and $r_2$.
\end{theorem}
The proof of Theorem~\ref{theorem:convergence} is provided in Appendix~\ref{appendix:theory:proof}.

\subsection{Explaining the Chain of Diffusion with Quantitative Trait Modeling} 
\label{subsection:links}

Just as phenotypes evolve through hidden genotypes in QT modeling, image features similarly evolve in the Chain of Diffusion. We identify high-frequency power spectra as a crucial phenotype, influenced by CFG scale as a selection mechanism. Using 2D Fourier transforms, we analyze high-frequency components above a certain threshold. Different CFG scales correspond to different selection strategies: a high CFG selects individuals in the right tail of the distribution, favoring more detailed features with reduced diversity, while a low CFG selects from the left tail.

Figure~\ref{figure:simulation:mean} compares the evolution of power spectra in the Chain of Diffusion (solid line) with the phenotype mean modeled by Theorem~\ref{theorem:convergence} (dotted lines). The simulation parameters, including initial mean $\mu_0$ and genetic deviation $\sigma_{G, 0}$, are set to match the original image set at iteration $0$. Three ratio configurations, $r_1$ and $r_2$, effectively model the power spectra distribution: CFG $7.5$ is modeled as selecting the top $5$\%, CFG  $2.5$ selects $50$\% of the samples slightly favoring higher frequency, and CFG $1.0$ selected the lower $30$\%. This modeling accurately captures power spectra evolution over six iterations of the Chain of Diffusion. More details can be found in Appendix~\ref{appendix:theory:setup:parameters}.

\subsection{Further Discussions}

\ys{
Observable features of generated images are influenced by a complex interplay of factors like random noise, text signals, and various hyperparameters. The influence of CFG is thus multifaceted---analogous to how a single genotype gives rise to multiple phenotypes (e.g., red hair and freckles). Among many possible phenotypes affected by CFG, we identified one with a clear and consistent trend over iterations. It shows how the theoretical framework on mean-drift from QT modeling can interpret our empirical results, revealing that key ideas from statistical genetics naturally align with model collapse. Beyond the scope of our current findings, we believe that statistical genetics offers a unified perspective about model collapse by reducing complex dynamics of the self-consuming training loops to a few parameters, from which valuable insights can be drawn.
}

\section{Reusable Image Generation with ReDiFine}
\label{section:experiment}

In the previous sections, we have discovered a significant role of CFG in model collapse and that a good choice of CFG can mitigate the collapse while preserving the first iteration FID. However, the optimal CFG value differs for each dataset (e.g., 1.5 for CelebA, 2.5 for Pokemon), making iterative finetuning for evaluating the collapse rate on each configuration inevitable to find an optimal CFG. In practice, it is unlikely that non-expert end users will go through such a process to prevent a potential model collapse. This raises the question: \emph{How can we design a user-friendly finetuning and generation strategy that can slow down model collapse without CFG tuning?}

\begin{figure*}[t!]
    \centering
    \includegraphics[width=0.99\textwidth]{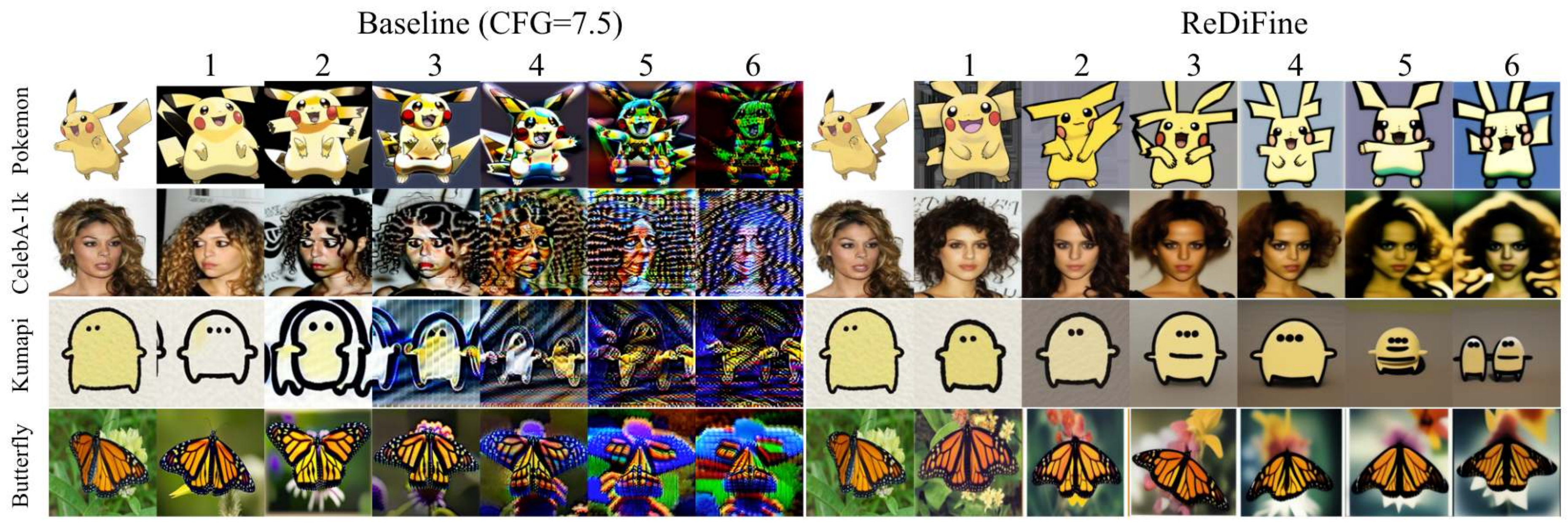}
    \vspace{-0.5em}
    \caption{
    \textbf{ReDiFine significantly improves the baseline without extensive hyperparameter tuning in the Chain of Diffusion.}
    It performs robustly across multiple datasets, resolving high-frequency degradation observed in the baseline with the default configuration.
    }
    \label{figure:samples_method}
\end{figure*}

To address this question, we again draw inspiration from the evolution process in nature where mutations naturally counteract genetic drift and preserve diversity. Furthermore, selection in nature is often a soft process rather than a hard truncation illustrated in Figure~\ref{figure:simulation:truncation}. This soft selection allows for the inclusion of outliers, thereby maintaining the overall genetic diversity. We connect these biological inspirations with two strategies: \emph{condition drop finetuning} to include more randomness and \emph{CFG scheduling} during generation to transform hard selection to a softer one\footnote[7]{We show a modified quantitative trait modeling result with these two modifications in Appendix~\ref{appendix:theory:setup} and show that it captures the ReDiFine experimental results effectively.}. We propose \textbf{Re}usable \textbf{Di}ffusion \textbf{Fine}tuning (ReDiFine) which integrates these two ideas and achieves a collapse rate similar to the optimal CFG, producing reusable images with minimal extra effort.

\noindent \textbf{Condition drop finetuning.}
We introduce condition drop finetuning, which randomly drops the text condition during finetuning to update both the conditional and unconditional scores. While suggested in the original CFG paper~\cite{ho2022classifier}, it was not a common practice during finetuning since good images are generated without it in the first iteration where model collapse is yet to happen (see Figure~\ref{figure:baseline_datasets}). However, the small $\text{Diff}$ ($= \text{Cond Score} - \text{Uncond Score}$) accumulates over iterations, leading to a significant gap as the Chain of Diffusion progresses. Condition drop finetuning with drop probability $0.2$ can preserve the norm of Diff over iterations (see Figure~\ref{figure:analysis:diffs}).

\begin{figure*}[t!]
    \centering
    \begin{subfigure}[b]{0.234\textwidth}
        \centering
        \includegraphics[width=1.0\textwidth]{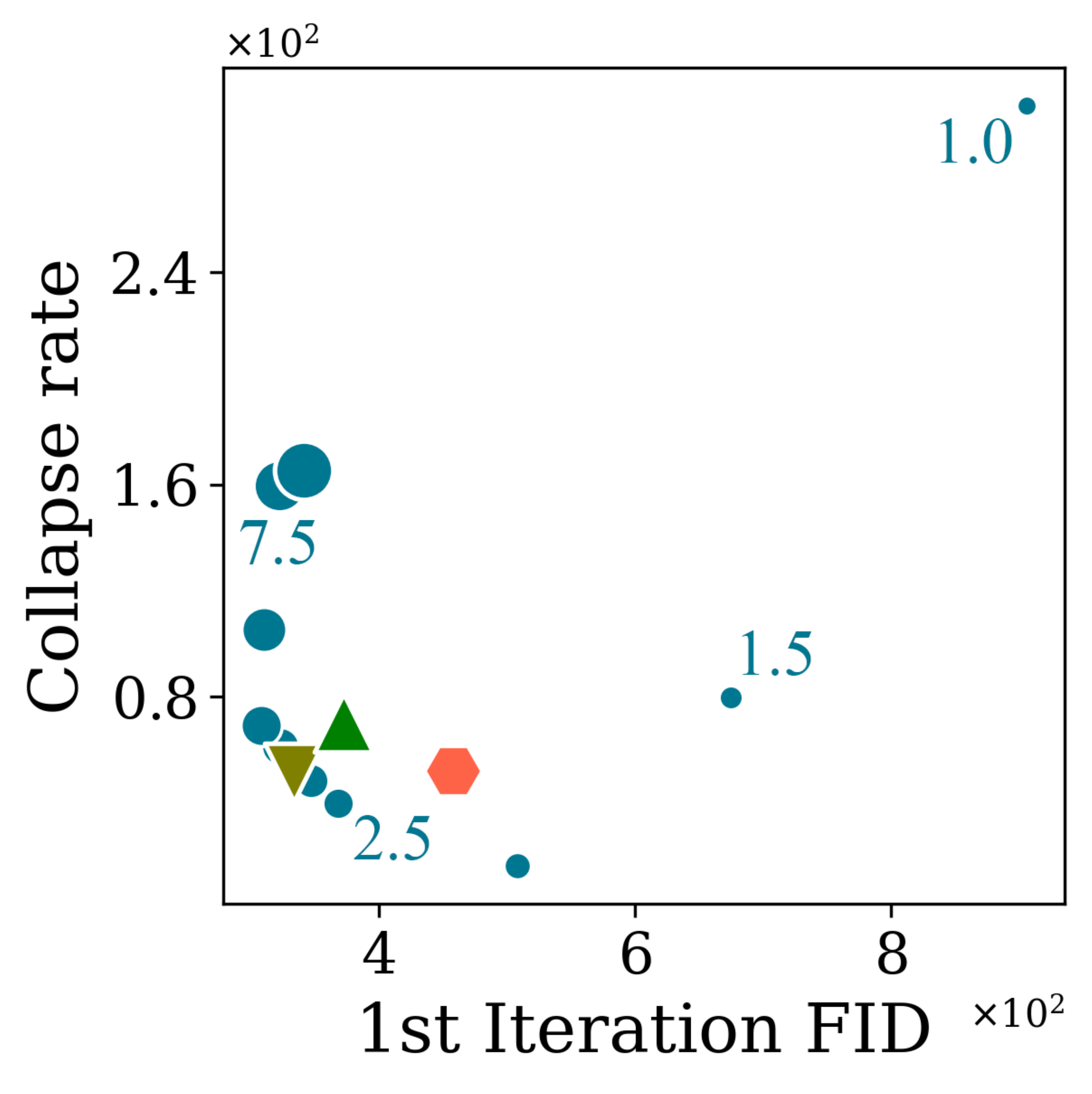}
        \subcaption{Pokemon.}
        \label{figure:metric:pokemon:tradeoff}
    \end{subfigure}
    \hfill
    \begin{subfigure}[b]{0.218\textwidth}
        \centering
        \includegraphics[width=1.0\textwidth]{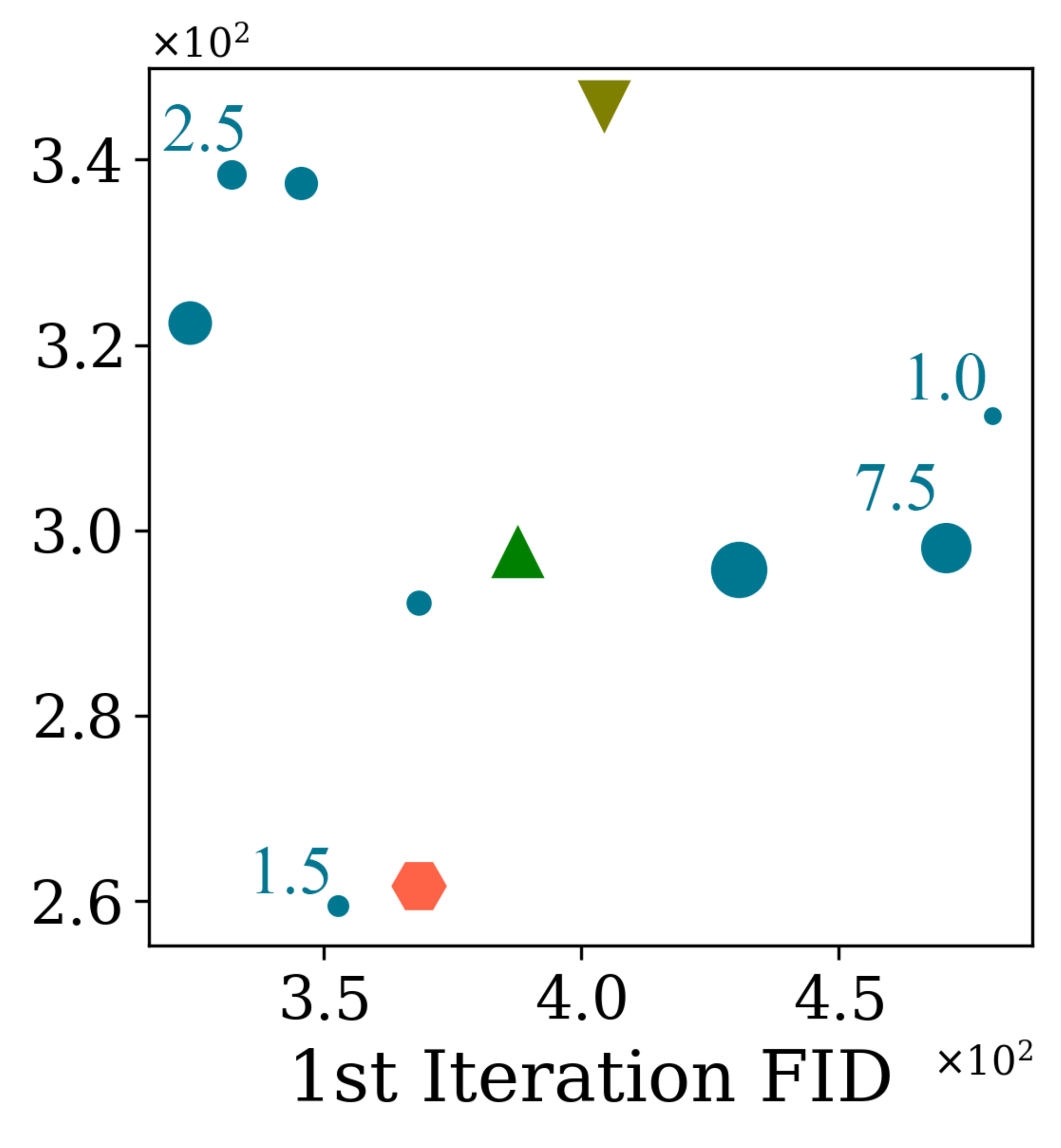}
        \subcaption{CelebA-1k.}
        \label{figure:metric:celeba:tradeoff}
    \end{subfigure}
    \hfill
    \begin{subfigure}[b]{0.218\textwidth}
        \centering
        \includegraphics[width=1.0\textwidth]{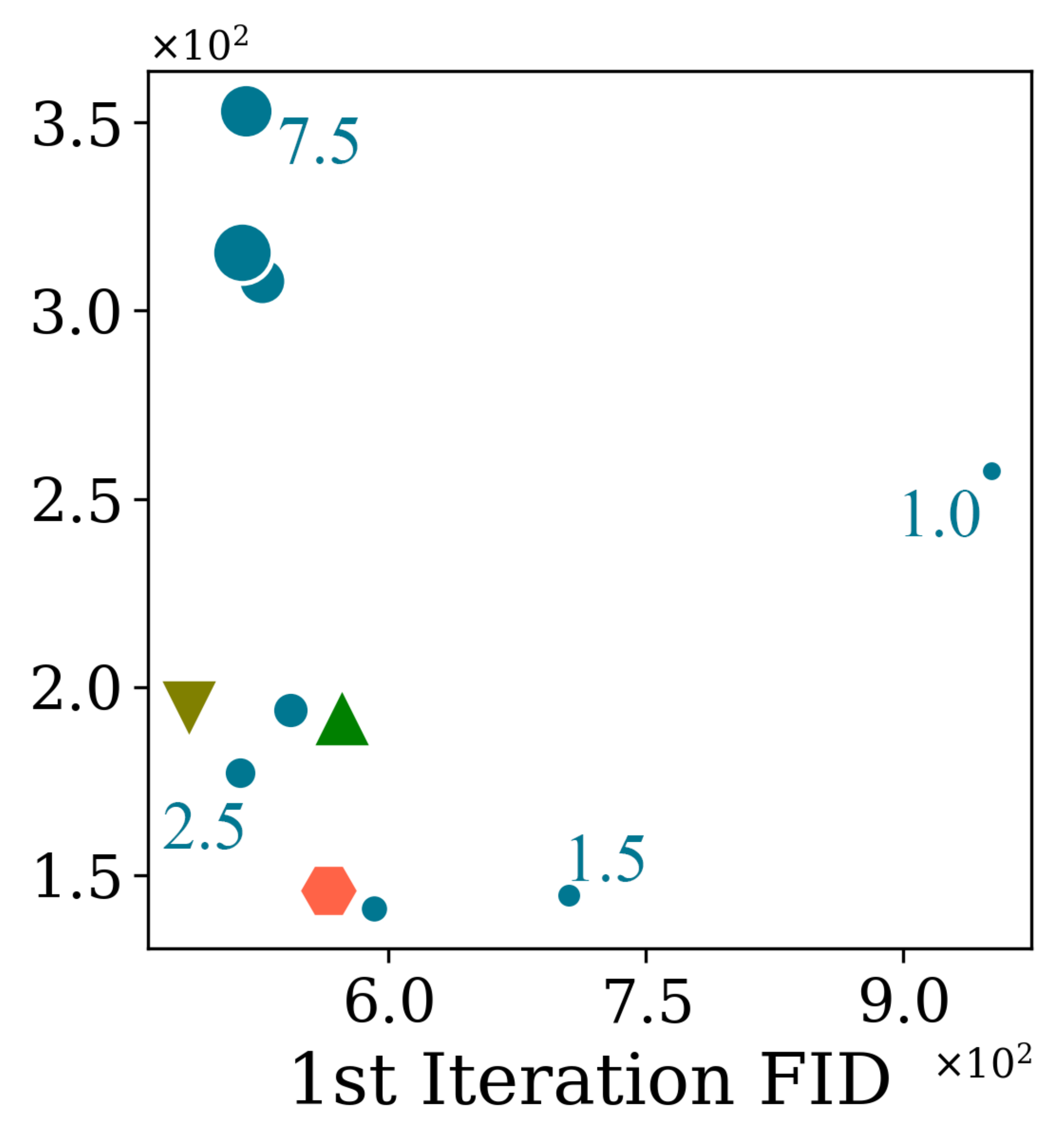}
        \subcaption{Kumapi.}
        \label{figure:metric:kumapi:tradeoff}
    \end{subfigure}
    \hfill
    \begin{subfigure}[b]{0.305\textwidth}
        \centering
        \includegraphics[width=1.0\textwidth]{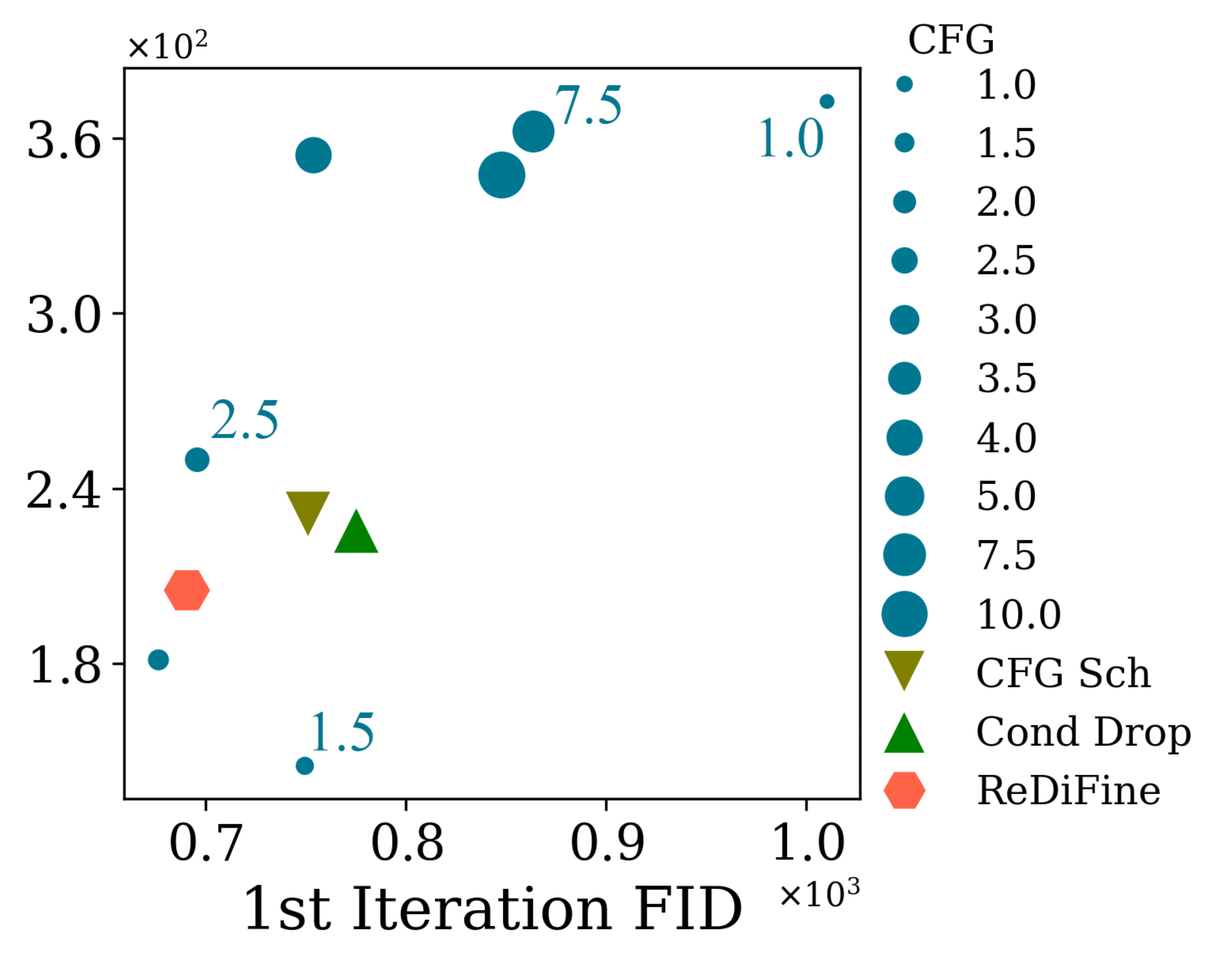}
        \subcaption{Butterfly.}
        \label{figure:metric:butterfly:tradeoff}
    \end{subfigure}
    \vspace{-0.5em}
    \caption{
    \textbf{ReDiFine performs comparably to the optimal CFG scale, outperforming the baseline (CFG=7.5) on collapse-FID trade-off across all four datasets}.
    While the optimal CFG scale varies for different datasets, ReDiFine consistently achieves low collapse rate and FID at the same time (lower is better). Note that the differences in FID are relatively smaller than those in collapse rate, supporting the necessity to evaluate collapse rate in the Chain of Diffusion.
    }
    \label{figure:method:tradeoffs}
    \vspace{-1em}
\end{figure*}

\noindent \textbf{CFG Scheduling.}
We propose gradually reducing the CFG scale during diffusion steps to mitigate the impacts of overemphasizing the conditional score in later stages, which can lead to high-frequency degradation. Specifically, we exponentially decrease the CFG scale $s$ during $T$ diffusion steps as $s = s_0 \times e^{ - \alpha \times t / T}$, where $s_0$ is the initial CFG scale and $\alpha$ is the rate of exponential decay. This scheduling approach is consistent with findings by \cite{balaji2022ediff}, which suggest that different diffusion steps contribute uniquely to the generation process. We note here that ReDiFine adopts exponential decrease of CFG scale for simplicity, but any advanced CFG scheduling methods~\cite{sadat2023cads, karras2024guiding, kynkaanniemi2024applying} can be explored in future research.

\noindent \textbf{ReDeFine Results.} We present the generated images and quantitative metrics for ReDiFine, showing its robust performance without hyperparameter tuning in all four datasets. We use the default initial CFG scale $s_0=7.5$, decay rate $\alpha = 2$, and condition drop probability $0.2$, except in the robustness comparison and ablation study. Figure~\ref{figure:samples_method} shows that ReDiFine generates significantly better images compared to the baselines (CFG=7.5), and performs comparably to the optimal CFG scales. In addition to the visual comparison, Figure~\ref{figure:method:tradeoffs} quantitatively shows the collapse-FID trade-off of ReDiFine. In all four datasets, ReDiFine shows substantially lower collapse rate (y-axis), compared to the baseline case (CFG=7.5). Furthermore, the performance of ReDiFine is close to the optimal Pareto curve spanned by different CFG scales, achieving similar performance as the optimal CFG values, demonstrating its effectiveness as a universal and user-friendly solution. In contrast, using a fixed CFG scale that works well for one dataset often fails on others: CFG $2.5$ is optimal for Pokemon but performs poorly for CelebA-1k and Butterfly, and CFG $1.5$ is optimal for CelebA-1k but performs poorly for Pokemon and Kumapi. Further experimental results of ReDiFine for six additional iterations, cross-domain data, and iterative retraining can be found in Appendix~\ref{appendix:method}.

\noindent \textbf{Ablation study \& Further analyses.} We conducted an ablation study to understand the contributions of condition drop finetuning and CFG scheduling to ReDiFine. As shown in triangles on Figure~\ref{figure:method:tradeoffs}, using only one of those strategies outperforms ReDiFine in Pokemon, but in all other datasets, using only one strategy shows higher collapse rate than ReDiFine. Especially in CelebA-1k, using either one of them showed significantly worse performance than ReDiFine. It suggests that combining condition drop and CFG scheduling builds robustness to the method, making ReDiFine effective across all tested datasets. We conducted further analyses on ReDiFine, examining the distribution of latent features, the evolution of the norm of Diff, the power spectra using 2D Fourier transforms, and forensic fingerprints based on prior studies~\cite{corvi2023intriguing, corvi2023detection}. Our analysis shows that ReDiFine effectively preserves the latent distribution and the norms of Diff over six iterations, with forensic fingerprints closely resembling those of the optimal CFG case. Detailed results are provided in Appendix~\ref{appendix:analysis}.

\section{Conclusion} 
\label{section:conclusion}

The influx of AI-generated data into the world is inevitable and training sets that consist of synthetic data will be part of the AI development pipeline. In this paper, we empirically and theoretically studied the scenario of finetuning a generative model with its own generated data, where a gradual degradation called ``model collapse'' happens. We (1) identify the most impactful factor through comprehensive empirical investigations, (2) develop a novel theoretical perspective inspired by statistical genetics to explain model collapse, and (3) propose ReDiFine strategy for diffusion finetuning to slow down image quality degradation in model collapse.

We started this paper with a question: can current AI models learn from their own output and improve themselves? Our paper shows a glimpse that widely-used text-to-image models are not ready to improve from their own creation quite yet. While we presented one solution focused on generating \emph{reusable data}, many open directions remain, such as developing algorithms that can distinguish between real and synthetic data and apply different learning techniques accordingly. We further discuss the scopes and limitations of this work in Appendix~\ref{appendix:limitations}.

\bibliographystyle{unsrt}
\bibliography{model_collapse}


\newpage
\appendix

\section{Scope of Work, Limitations and Broader Impacts}
\label{appendix:limitations}

\subsection{Scope and limitations}
\label{appendix:limitations:scope}

In this paper, we provide three main contributions: significant observations from extensive experiments, a novel theoretical framework inspired by statistical genetics, and an effective mitigation strategy for model collapse. We detail the boundary and scope of our work in the literature to better understand our contributions.

First, our empirical observations are based on self-consuming training loops and finetuning of Stable Diffusion v1.5. Model collapse is a universal phenomenon observed in recent machine learning training, as high-quality generated contents are now available across multiple domains. We target diffusion finetuning with LoRA as it is one of the most practical interests in the field, where tons of finetuned models and images are being released to the Internet every day. While we already included the effects of mixing real images, iterative retraining, and domain accumulation to extend the setting from fully self-consuming training loops, we remain open to further investigations about full training, different finetuning methods, and architectures for future work.

Next, our theoretical framework focuses on introducing an interesting analogy between two fields to open a new area for discussion. We showed that QT modeling from statistical genetics provides a powerful tool to understand model collapse based on the structural analogy between self-consuming training loops and genetic processes. Further extensions based on advanced methods from statistical genetics would be an important future direction in investigating the model collapse of generative models. Specifically, finding connections between these advanced theoretical methods with self-consuming training loops augmented with real data or accumulated data would be an exciting future direction to extend this analogy.

Finally, the proposed ReDiFine is intended to be an efficient solution that can avoid multiple evaluations of the Chain of Diffusion while achieving performance comparable to that of the optimal CFG scale. It could perform comparably to the optimal CFG on four different datasets, which is a promising achievement as the first solution specific to diffusion finetuning. Investigating a more powerful method specific for diffusion finetuning would be a promising direction for future research, but we aim to introduce a simple method without additional real images or hyperparameter tuning that demonstrates the concept and importance of generating reusable images.

\subsection{Broader Impacts}
\label{appendix:limitations:impacts}

Generating reusable images for further training or finetuning holds the potential to significantly enhance the quality, stability, and robustness of future generative models. By promoting continuity across iterations, this approach can improve long-term performance and reduce degradation in self-consuming loops. However, as generative models become more powerful and self-reinforcing, there is an increasing risk that such advancements could be misused to replicate, mimic, or infringe upon copyrighted assets, licensed material, or the creative work of artists—especially when the boundary between generated and human-made content becomes less clear.

We strongly emphasize that any use of generated content for further training must be carried out with a deep understanding of the rights and intent associated with the original data. Respecting intellectual property, license agreements, and the creative labor of individuals is essential to maintaining ethical standards in machine learning research. While our work aims to improve the technical reliability of generative models, we advocate for responsible usage and encourage the community to develop frameworks and norms that protect creators while fostering innovation.

\section{Experimental setup}
\label{appendix:experimental_setting}

\subsection{Hyperparameters}
\label{appendix:experimental_setting:hyperparameters}

We finetune Stable Diffusion v1.5~\cite{rombach2022high} using LoRA~\cite{hu2021lora} at each iteration, with ft-MSE~\cite{huggingface-sd-vae} as a fixed VAE to project images into latent space. Horizontal flip is the only image augmentation applied. Our implementation follows \cite{kohya_ss} and is built on PyTorch v2.2.2~\cite{Ansel_PyTorch_2_Faster_2024}, with torchvision 0.17.2, running on CUDA 12.4 using NVIDIA A-100 and L40S GPUs. Each finetuning and generation takes about 2 hours for ~1k images. All default hyperparameters are listed in Table~\ref{table:experimental_setting:hyperparameters}.

\begin{table}[h!]
    \centering
    \caption{
    Default hyperparameters used for the Chain of Diffusion.
    }
    \label{table:experimental_setting:hyperparameters}
    \begin{tabular}{@{}ll@{}}
        \toprule
        \textbf{Hyperparameter}      & \textbf{Value}              \\ 
        \midrule
        Optimizer                    & AdamW \\
        Learning Rate - Unet         & $0.0001$               \\
        Learning Rate - CLIP         & $0.00005$              \\
        LoRA Weight Scaling          & $8$                    \\ 
        LoRA Rank                    & $32$                   \\
        Batch Size                   & $6$                    \\
        Max Epochs             & $100$                  \\
        CLIP Skip                    & $2$                    \\
        Noise Offset                 & $0.0$                  \\
        Mixed Precision              & fp16                 \\
        Loss Function                & MSE                   \\
        Min SNR gamma                & $5.0$                  \\
        Max Gradient Norm Clipping   & $1.0$                  \\
        Caption Dropout Rate         & $0.0$                  \\
        Sampler                      & Euler A              \\
        Classifier-Free Guidance Scale & $7.5$                \\
        Number of Diffusion Steps               & $30$                   \\ 
        Number of Images per Prompt & $1$ \\
        \bottomrule
    \end{tabular}
\end{table}

\subsection{Datasets}
\label{appendix:experimental_setting:datasets}

We use four image datasets to demonstrate the universal nature of degradation: Pokemon~\cite{pokedex}, Kumapi~\cite{kaggle-illustrations-kumapi390}, Butterfly~\cite{butterflydataset}, and CelebA-1k~\cite{liu2015faceattributes}, covering animation, handwriting, and real images. All images are resized to $512 \times 512$ pixels. Text prompts are generated using BLIP captioner~\cite{https://doi.org/10.48550/arxiv.2201.12086} and Waifu Diffusion v1.4 tagger~\cite{hakurei2024waifu}. Sample images and prompts can listed in Table~\ref{table:samples:all}.

\paragraph{Pokemon.}The Pokemon dataset~\cite{pokedex} contains $1008$ images indexed by number, with prompts combining BLIP captions (length $50$-$75$ words) and Waifu Diffusion tagger.

\newcolumntype{C}[1]{>{\centering\arraybackslash}m{#1}}
\newcommand{\includeimage}[1]{\includegraphics[width=\linewidth]{#1}}

\begin{table}[h!]
    \centering
    \caption{Sample images and prompts for Pokemon, CelebA-1k, Kumapi, and Butterfly datasets.}
    \label{table:samples:all}
    \scalebox{1.0}{
        \begin{tabularx}{\textwidth}{C{0.8\textwidth} C{0.2\textwidth}}
            \midrule
            \multicolumn{2}{c}{\textbf{Pokemon}} \\
            \midrule
            a green pokemon with red eyes and a leaf on the back of its head and tail, an image of the pokemon character with a red eye and big green tail, all set up to look like it is holding a lea, ultra-detailed, high-definition, high quality, masterpiece, sugimori ken (style), solo, smile, open mouth, simple background, red eyes, white background, standing, full body, pokemon (creature), no humans, fangs, transparent background, claws, Bulbasaur
            & \includegraphics[width=2cm]{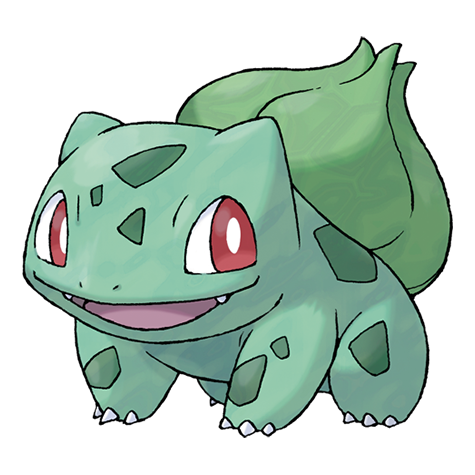} \\
            a very cute looking pokemon with a big leaf on its back and a big leaf on its head, a very cute little pokemon character with leaves in the back ground around his chest and hea, white background, ultra-detailed, high-definition, high quality, masterpiece, sugimori ken (style), solo, red eyes, closed mouth, standing, full body, pokemon (creature), no humans, fangs, transparent background, claws, outline, white outline, animal focus, fangs out
            & \includegraphics[width=2cm]{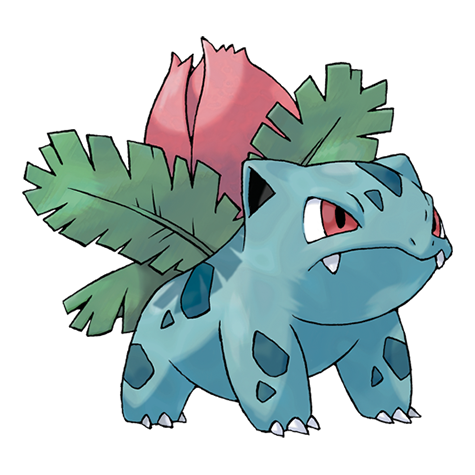} \\
            \midrule
            \multicolumn{2}{c}{\textbf{CelebA-1k}} \\
            \midrule
            a woman with brown hair smiling and posing for a picture in front of a mirror and gold and white stripes
            & \includegraphics[width=2cm]{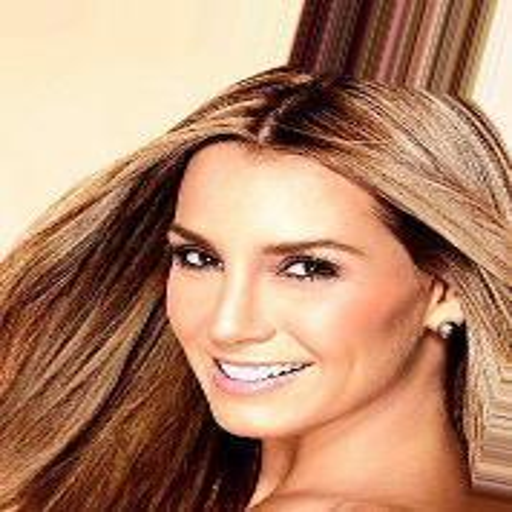} \\
            a woman with a very long red hair smiles and laughs on a city street and other people in the background
            & \includegraphics[width=2cm]{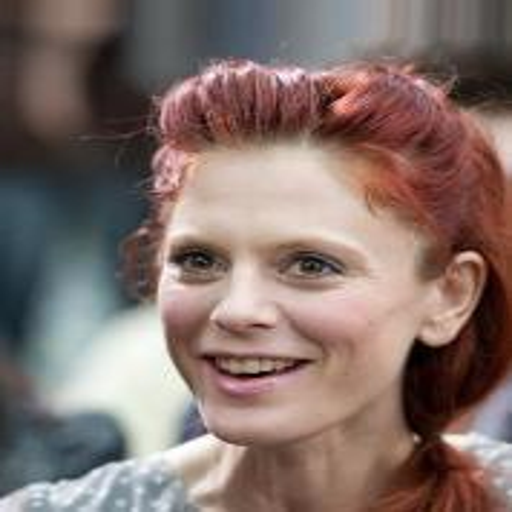} \\
            \midrule
            \multicolumn{2}{c}{\textbf{Kumapi}} \\
            \midrule
            solo, simple background, food, donut, grey background, no humans, food focus, still life, Kumapi style
            & \includegraphics[width=2cm]{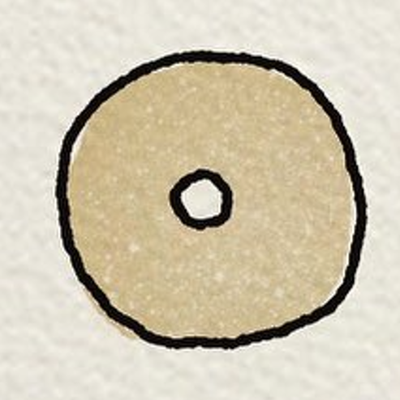} \\
            solo, looking at viewer, cute yellow figure, two tiny hands and feet, simple background, black dot eyes, white background, grey background, no humans, Kumapi style
            & \includegraphics[width=2cm]{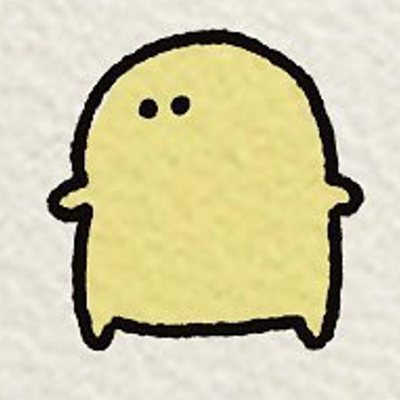} \\
            \midrule
            \multicolumn{2}{c}{\textbf{Butterfly}} \\
            \midrule
            a crimson patched longwing butterfly with a red and black stripe on its wings, wings are long, narrow, rounded, black, crossed on fore wing by broad crimson patch, and on hind wing by narrow yellow line
            & \includegraphics[width=2cm]{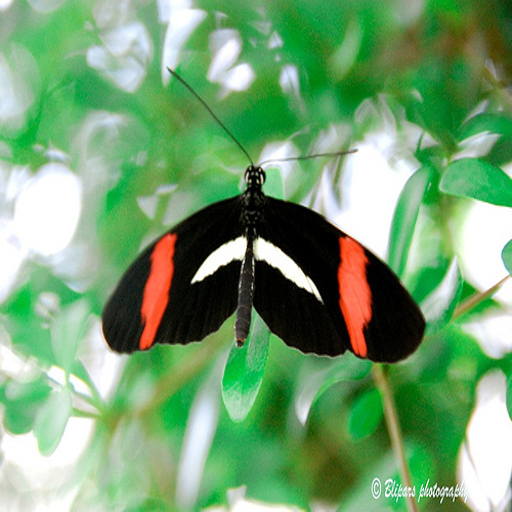} \\
            a Common Buckeye butterfly is sitting on a flower in the sun, wings scalloped and rounded except at drawn-out fore wing tip, on hind wing, 1 large eyespot near upper margin and 1 small eyespot below it. Eyespots are black, yellow-rimmed, with iridescent blue and lilac irises, on fore wing, 1 very small near tip and 1 large eyespot in white fore wing bar.
            & \includegraphics[width=2cm]{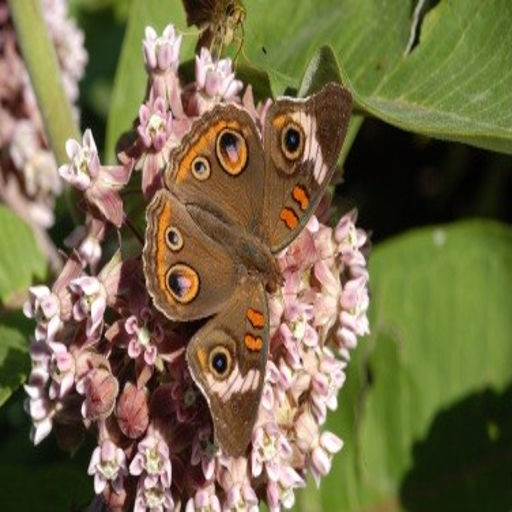} \\
            \bottomrule
        \end{tabularx}
    }
\end{table}

\paragraph{CelebA-1k.}CelebA-1k is a subsample of $1000$ images from CelebA~\cite{liu2015faceattributes}, with BLIP captions ($25$-$50$ words).

\paragraph{Kumapi.}Kumapi~\cite{kaggle-illustrations-kumapi390} includes $391$ handwriting-style images, with Waifu-generated prompts and manual adjustments.

\paragraph{Butterfly.}Butterfly~\cite{butterflydataset} includes $832$ real picture images across $8$ species, using BLIP captions (length $50$-$75$ words) and species descriptions.

\subsection{Evaluation metrics}
\label{appendix:experimental_setting:metrics}

\paragraph{FID} The Fréchet Inception Distance (FID)~\cite{heusel2017gans} is a widely used metric for evaluating the quality of generative models. It measures the similarity between the distributions of generated images and real images by comparing their feature representations extracted from a pretrained Inception network. Following \cite{stein2024exposing}, we use DiNOv2 feature extractor~\cite{oquab2023dinov2} instead of Inception, as it aligns better with human evaluation for high-quality images.

\paragraph{CLIP score} The CLIP score is a metric used to evaluate the alignment between an image and a text description, leveraging the pretrained CLIP (Contrastive Language-Image Pre-training) model~\cite{radford2021learning}. CLIP is designed to encode both images and text into a shared embedding space, where semantically related image-text pairs have high cosine similarity. Based on this embedding space, the CLIP score measures the text-alignment capability of text-to-image generative models by computing cosine similarities of generated images and text descriptions used for generations.

\paragraph{SFD} We suggest and compute the average sample-wise feature distance (SFD) between a pair of images corresponding to the same text prompt as the fidelity metric applicable for text-to-image generation. It computes text-prompt-wise similarities of the real image and generated image, more directly comparing two pairs of images while other metrics compares the overall distribution of datasets. We calculate 
\begin{equation}
    {\rm SFD}_k = {\frac{1}{N}} \sum d(f(x_{k, i}), f(x_{0, i}))
\end{equation}
to evaluate each iteration $k$. SFD overcomes the problem of FID being sensitive to the number of images to compare.

\paragraph{Recall} As a key metric for evaluating the diversity of generative models, recall measures the fraction of the real data distribution that is captured by the generated data distribution. Following the method proposed by \cite{kynkaanniemi2019improved}, we utilize the DiNOv2 feature extractor and set the number of neighbors to 5 for our evaluations.

\section{Different CFG Scales}
\label{appendix:baseline_experiments}

\begin{figure}[h!]
    \centering
    \includegraphics[width=0.6\textwidth]{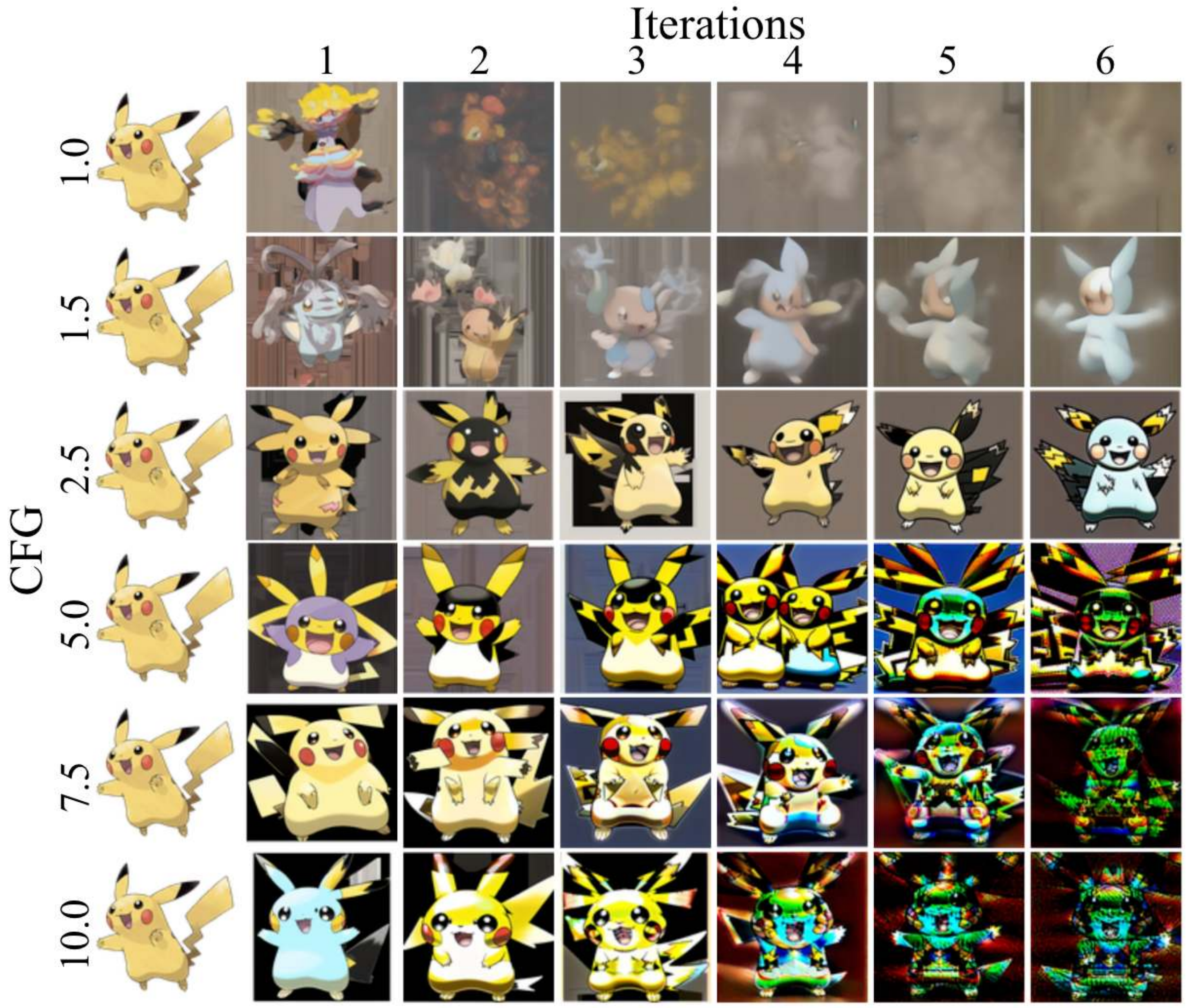}
    \caption{
    Chain of Diffusion for Pokemon at various CFG scales. The optimal CFG scale is $2.5$.
    }
    \label{figure:baseline:pokemon}
\end{figure}

\begin{figure}[h!]
    \centering
    \includegraphics[width=0.6\textwidth]{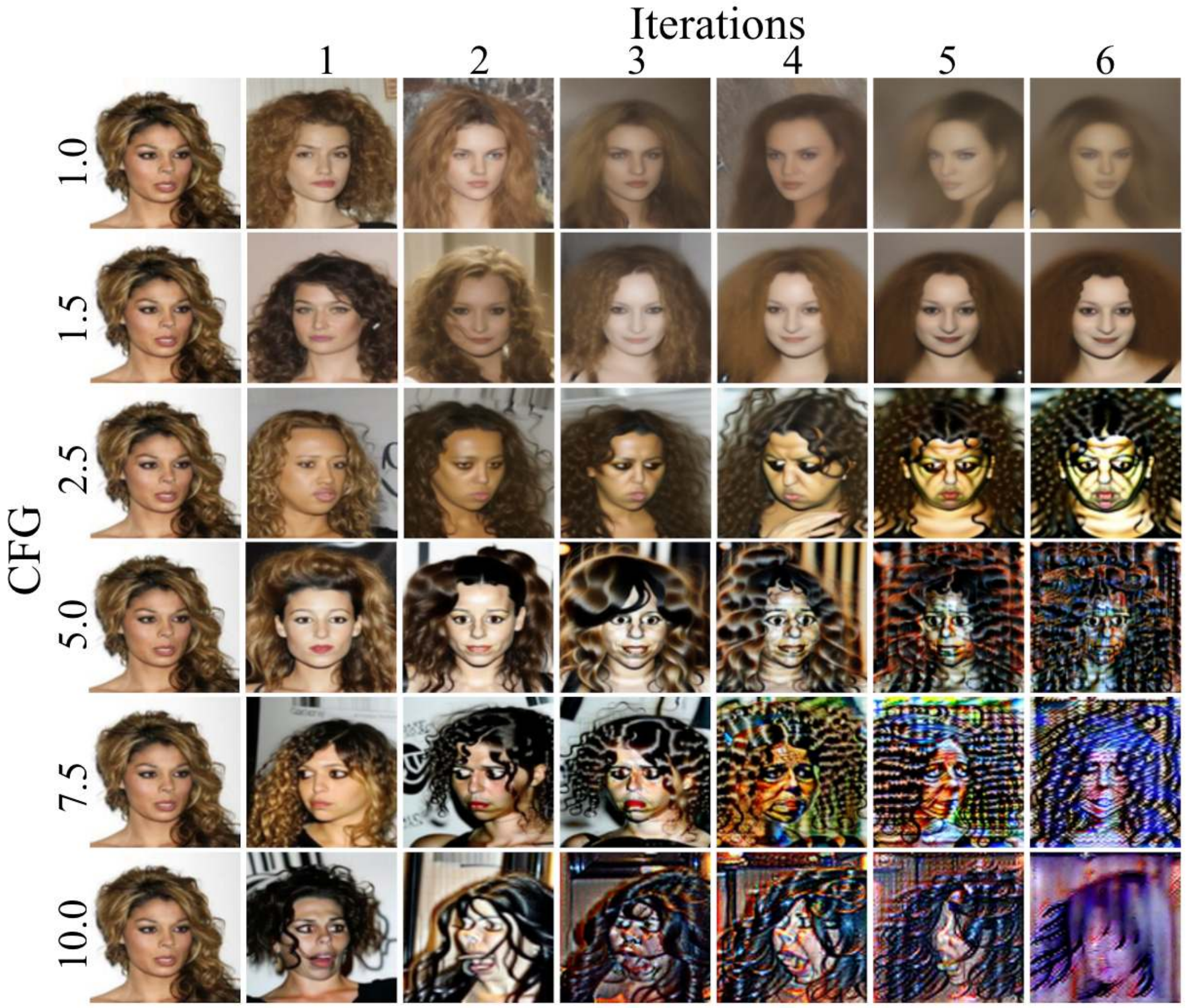}
    \caption{
    Chain of Diffusion for CelebA-1k at various CFG scales. The optimal CFG scale is $1.5$.
    }
    \label{figure:baseline:celeba}
\end{figure}

\begin{figure}[h!]
    \centering
    \includegraphics[width=0.6\textwidth]{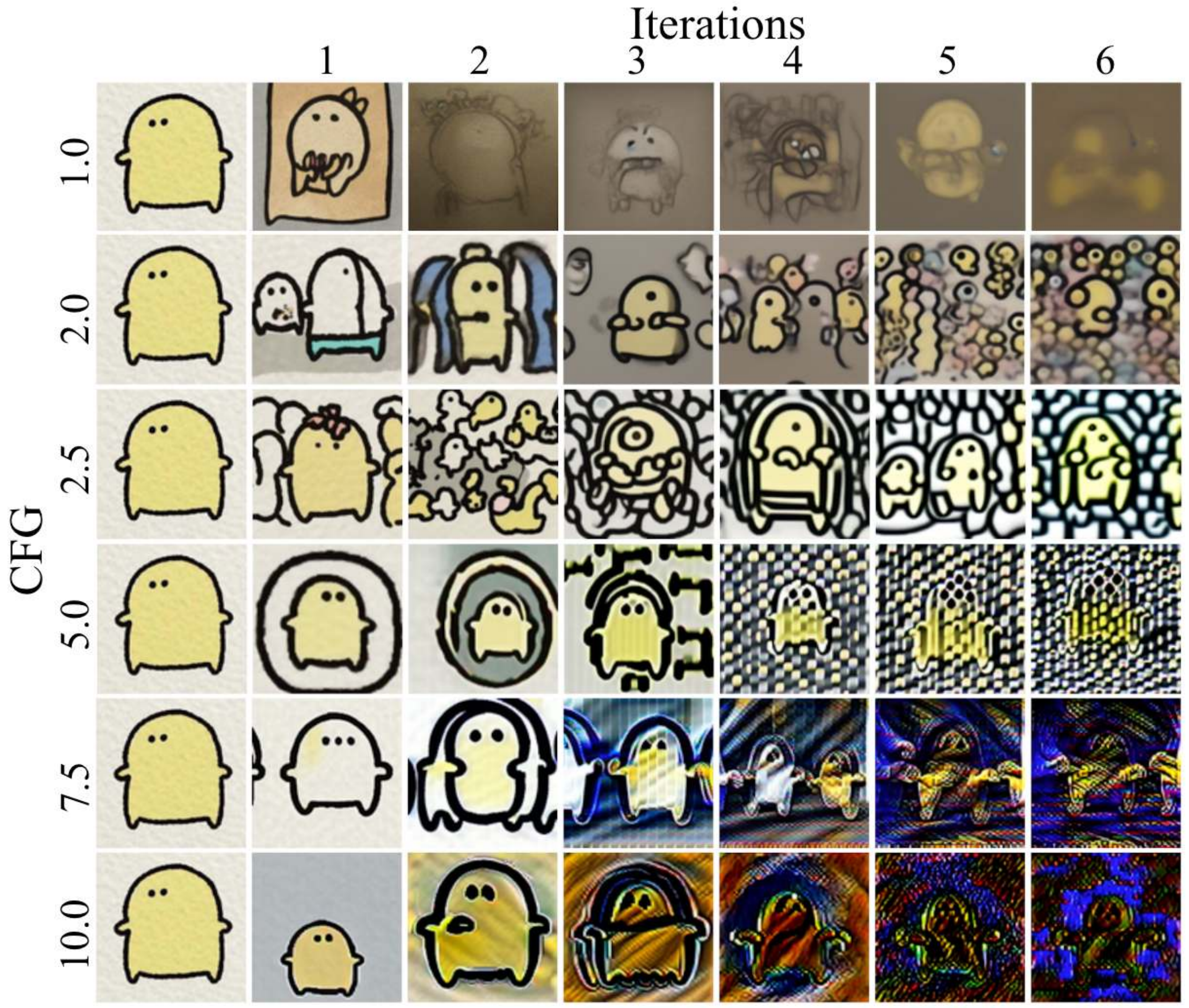}
    \caption{
    Chain of Diffusion for Kumapi at various CFG scales. The optimal CFG scale is $2.5$.
    }
    \label{figure:baseline:kumapi}
\end{figure}

\begin{figure}[h!]
    \centering
    \includegraphics[width=0.6\textwidth]{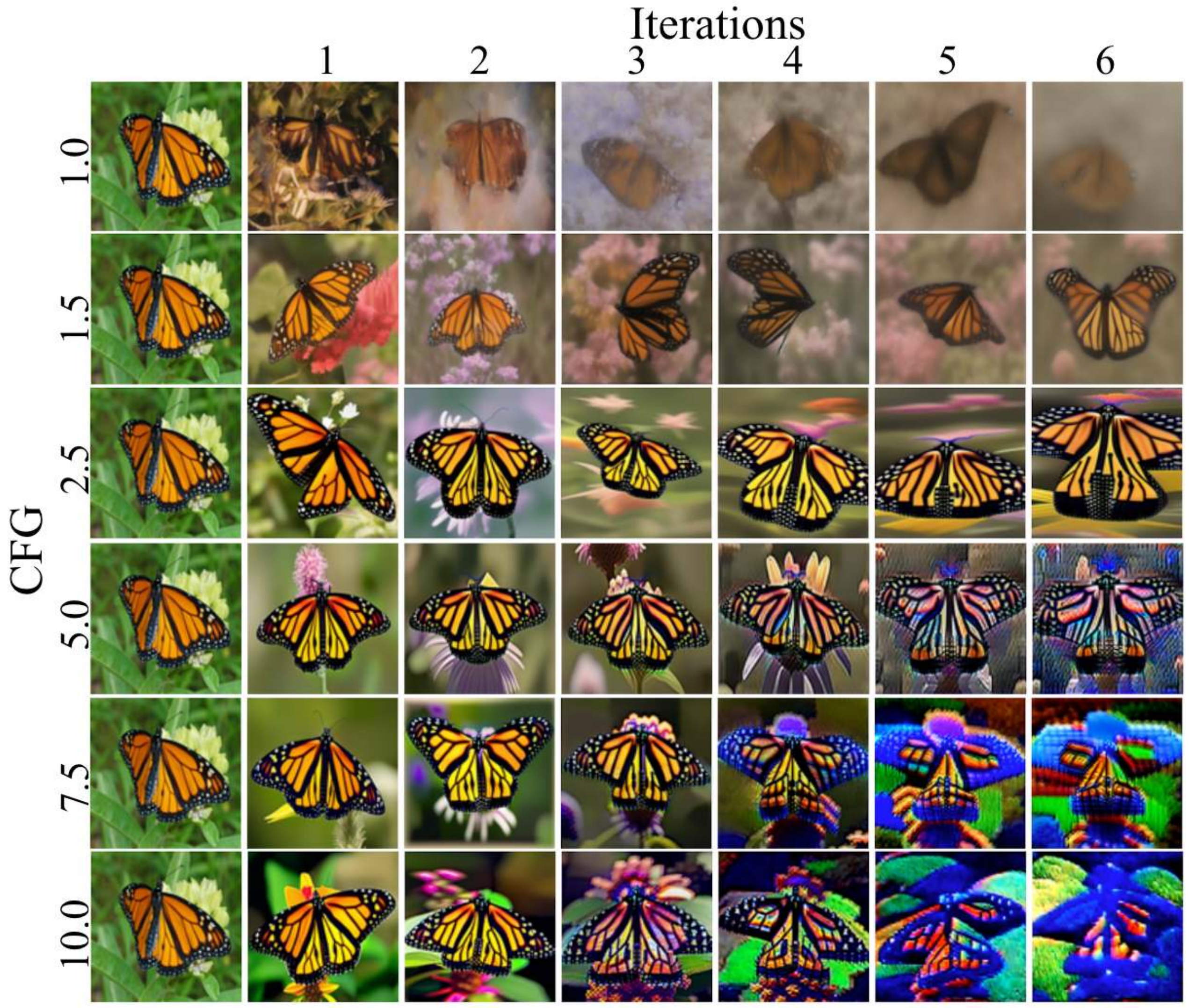}
    \caption{
    Chain of Diffusion for Butterfly at various CFG scales. The optimal CFG scale is $1.5$.
    }
    \label{figure:baseline:butterfly}
\end{figure}

This section presents images from Chain of Diffusion at various CFG scales for the four datasets. Figure~\ref{figure:baseline:pokemon}, \ref{figure:baseline:celeba}, \ref{figure:baseline:kumapi}, and \ref{figure:baseline:butterfly} corresponds to Pokemon, CelebA-1k, Kumapi, and Butterfly, respectively. Each dataset shows results for five CFG scales, covering high, medium, and low values. The optimal medium CFG scale is $2.5$ for Pokemon and Kumapi, and $1.5$ for CelebA-1k and Butterfly. Lower-than-optimal CFG scales lead to low-frequency degradations, while higher-than-optimal scales result in high-frequency degradations with saturated colors and repetitive patterns. Notably, the optimal CFG for Pokemon and Kumapi causes severe degradation in CelebA-1k and Butterfly.

\section{Hyperparameter investigations to unveil the most significant factor of degradation}
\label{appendix:hyperparameter_experiments}

This section presents experimental results identifying the most significant factors contributing to degradation in the Chain of Diffusion. We systematically vary each hyperparameter from Figure~\ref{figure:tradeoff_with_table}, using the default settings from Table~\ref{table:experimental_setting:hyperparameters}, to assess their impact on degradation. The CFG scale is fixed at $7.5$ for all cases.

\subsection{Training set size}
\label{appendix:hyperparameter_experiments:size}

\begin{figure}[h!]
    \centering
    \includegraphics[width=0.6\textwidth]{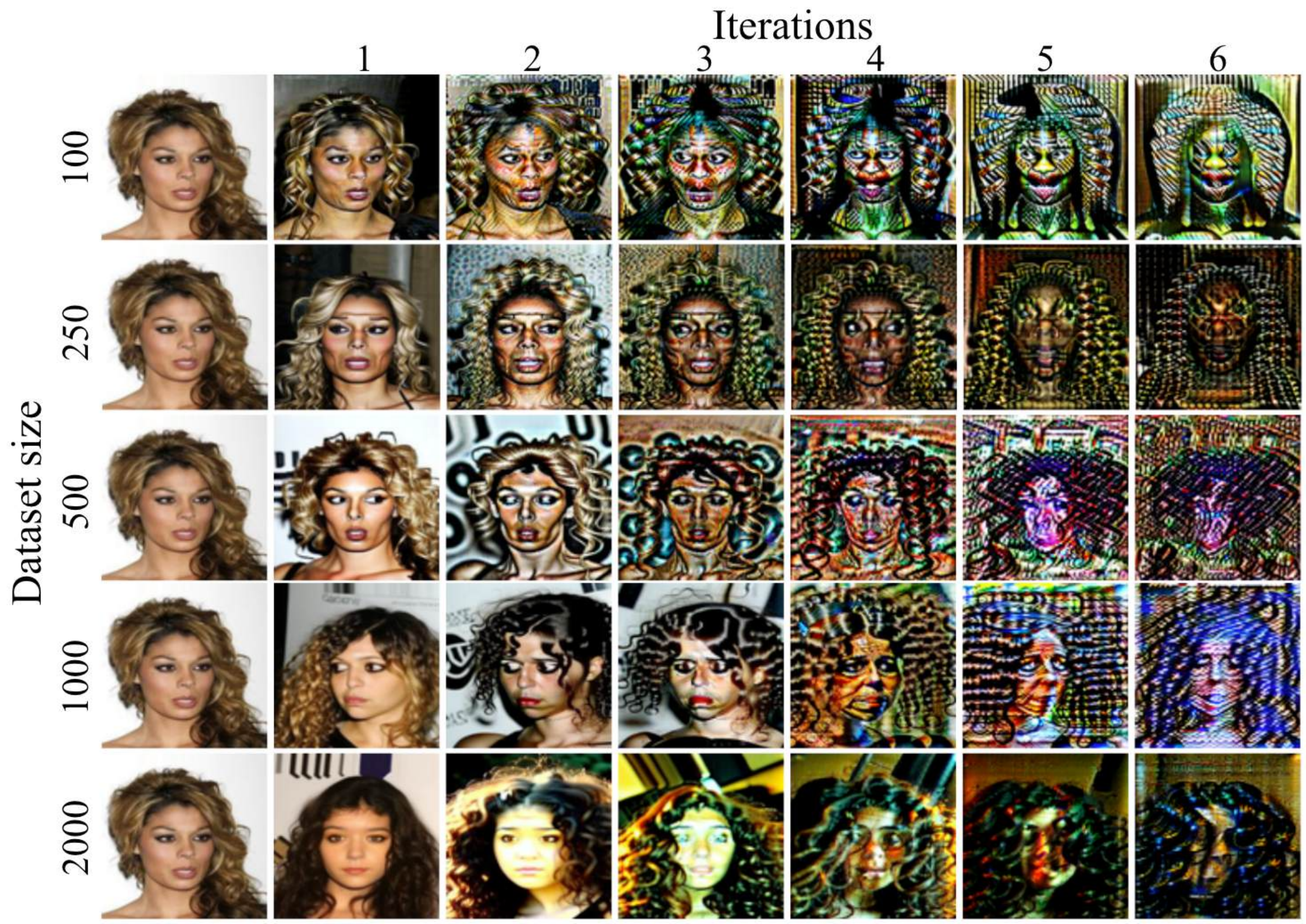}
    \caption{
    Chain of Diffusion on CelebA dataset with varying training set sizes. Degradation occurs faster with smaller sets, but all result in severe degradation by the $6$th iteration.
    }
    \label{figure:hyperparameters_experiments:size}
\end{figure}

We used CelebA dataset~\citep{liu2015faceattributes} to examine how training set size (both $D_0$ and $D_k$) impacts degradation in the Chain of Diffusion. By subsampling, we adjusted the training set to $100$, $250$, $500$, and $2000$ images. Degradation occurs regardless of dataset size, as shown in Figure~\ref{figure:hyperparameters_experiments:size}, but appears earlier with smaller sets. By the $6$th iteration, images degrade severely for all cases. The number of parameter updates was kept constant across all dataset sizes.

\subsection{Number of images per prompt}
\label{appendix:hyperparameter_experiments:more_than_one}

\begin{figure}[h!]
    \centering
    \includegraphics[width=0.6\textwidth]{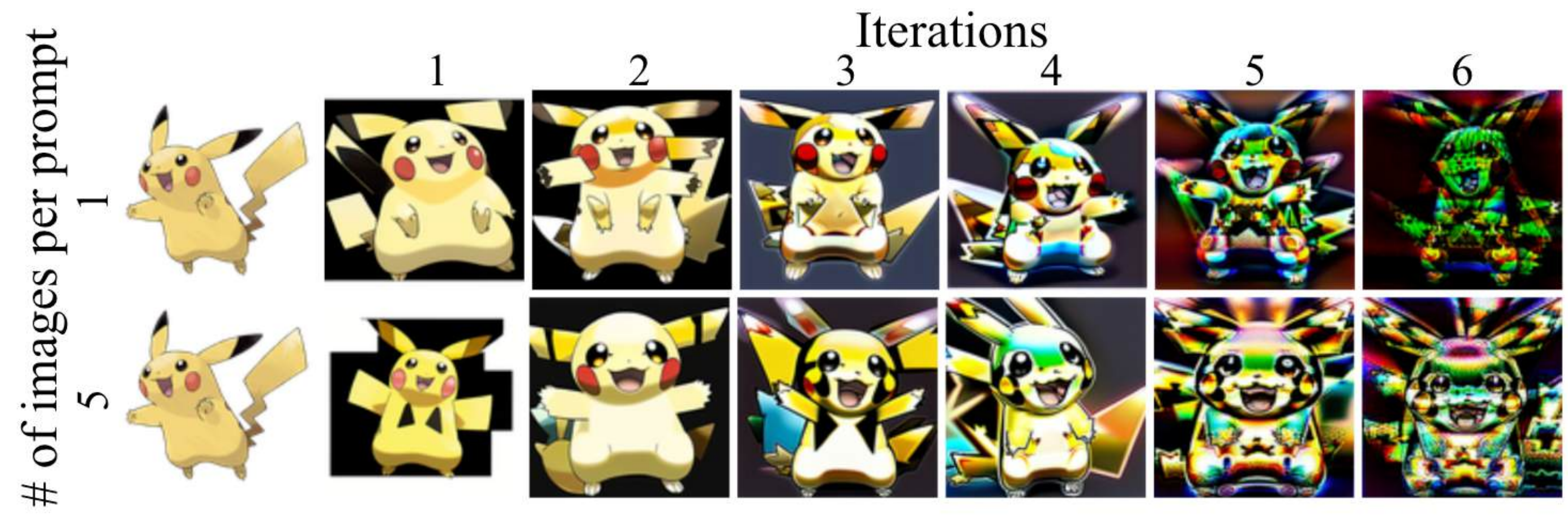}
    \caption{
    Chain of Diffusion on Pokemon dataset with multiple images generated per prompt. Increasing the training set size in this way does not mitigate degradation.
    }
    \label{figure:hyperparameters_experiments:more_than_one}
\end{figure}

Generating multiple images per prompt is a simple way to increase training set diversity and can be considered as a solution to mitigate degradation in the Chain of Diffusion. We tested this by generating $5$ times more images. As shown in Figure~\ref{figure:hyperparameters_experiments:more_than_one}, while degradation is slightly delayed (by one iteration), it remains unmitigated, and the high computational cost makes this approach impractical.

\subsection{Mixing real images to synthetic sets}
\label{appendix:hyperparameter_experiments:mixing}

\begin{figure}[h!]
    \centering
    \includegraphics[width=0.6\textwidth]{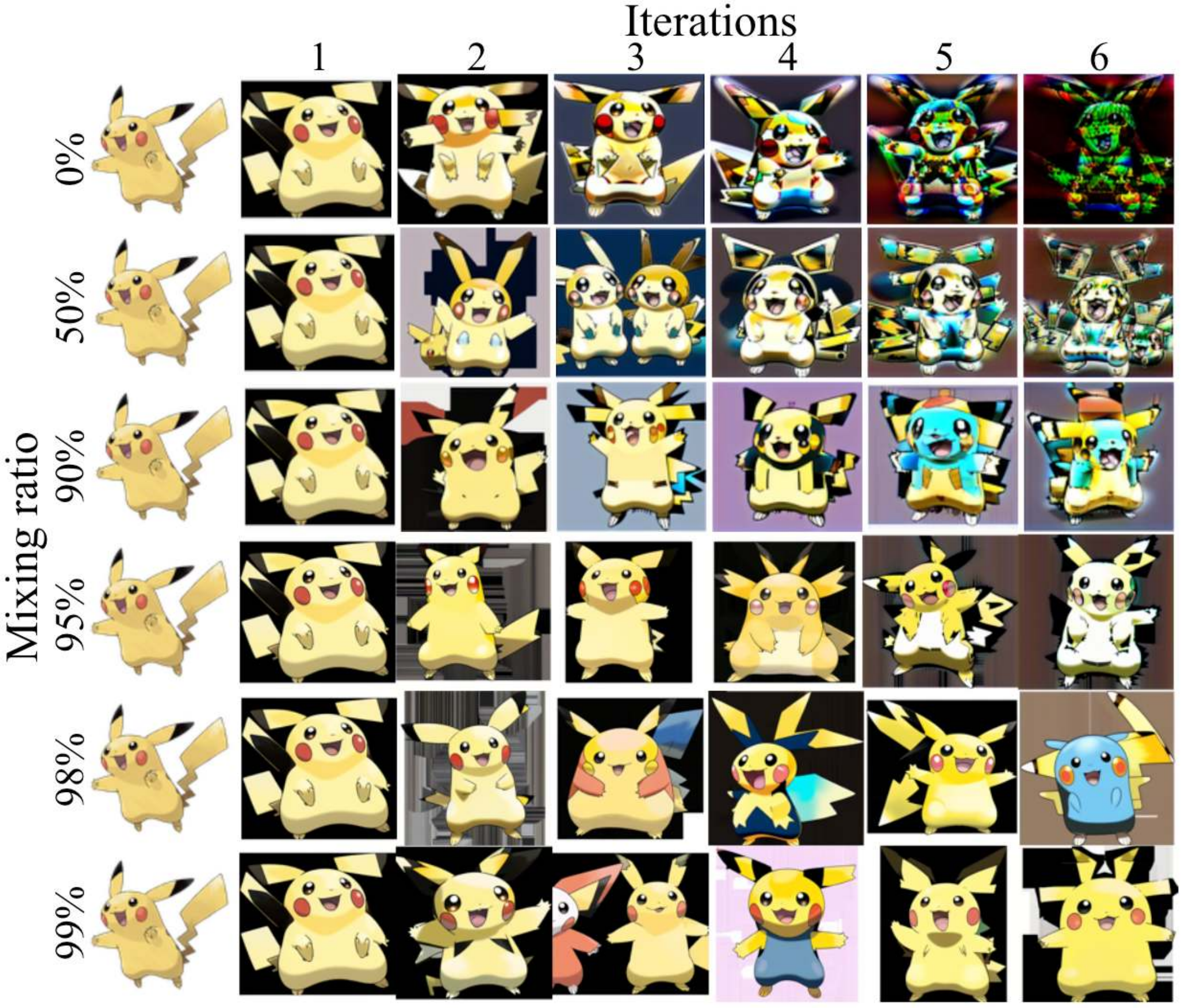}
    \caption{
    Chain of Diffusion on Pokemon dataset with real images randomly replacing synthetic images at each iteration. A $50$\% replacement rarely slows degradation, while $10$\% synthetic images are sufficient to initiate it.
    }
    \label{figure:hyperparameters_experiments:mixing:pokemon}
\end{figure}

\begin{figure}[h!]
    \centering
    \includegraphics[width=0.6\textwidth]{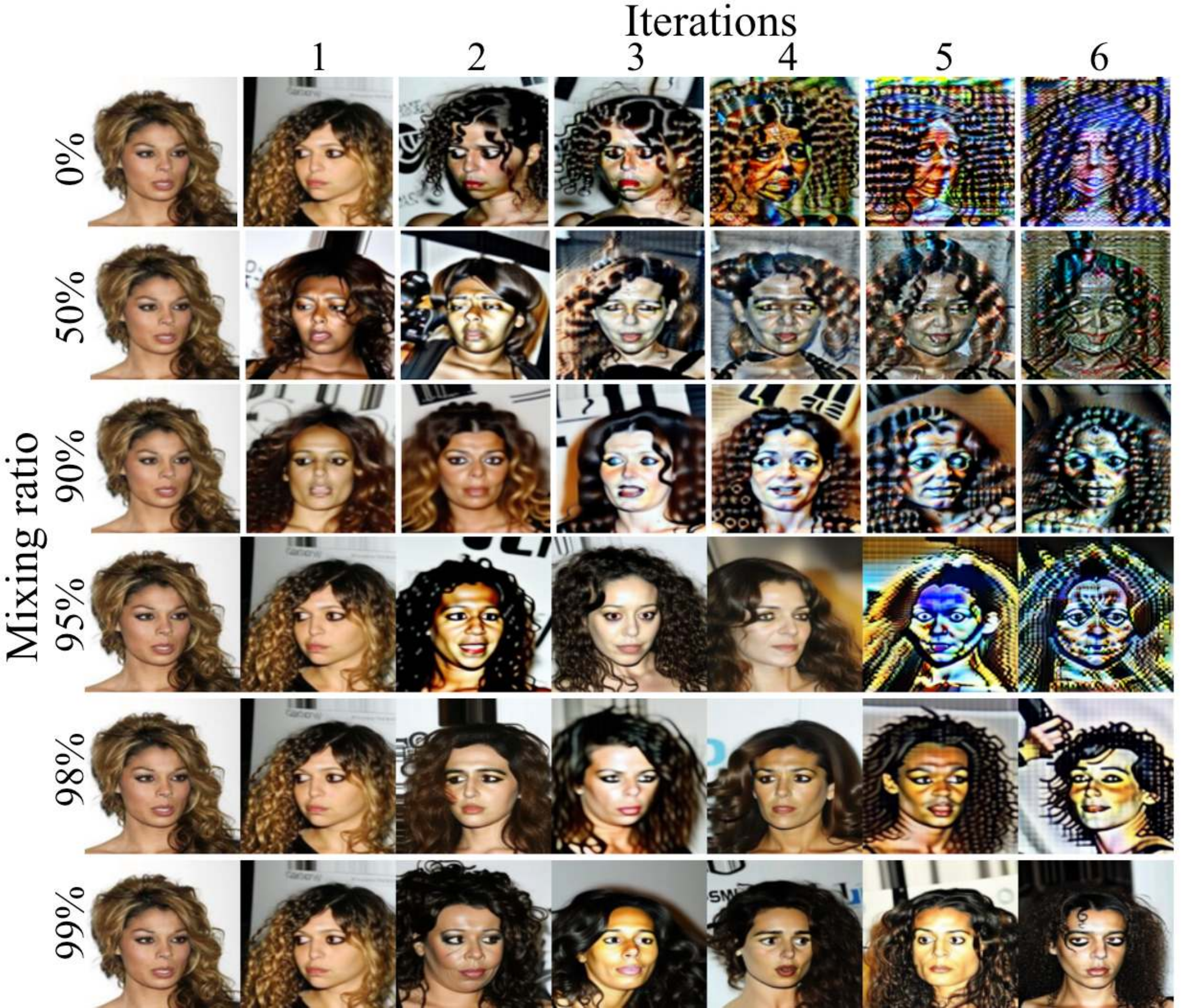}
    \caption{
    Chain of Diffusion on CelebA-1k dataset with real images randomly replacing synthetic images at each iteration. A $50$\% replacement rarely slows degradation, while $5$\% synthetic images are sufficient to initiate it. It suffers from more severe degradation than Pokemon dataset as compared with Figure~\ref{figure:hyperparameters_experiments:mixing:pokemon}.
    }
    \label{figure:hyperparameters_experiments:mixing:celeba}
\end{figure}

Many previous works suggest augmenting the training set with real images to mitigate degradation during iterative training. We investigated whether mixing images from the original training set into the synthetic set at each iteration could alleviate this issue. At each iteration, images are randomly replaced with corresponding real images. Figure~\ref{figure:hyperparameters_experiments:mixing:pokemon} and \ref{figure:hyperparameters_experiments:mixing:celeba} show how degradation in the Chain of Diffusion varies when $50$\%, $90$\%, $95$\%, $98$\% and $99$\% of images are replaced for Pokemon and CelebA-1k datasets, respectively. Notably, even $5$\% synthetic images are sufficient to induce degradation, and $50$\% replacement rarely slows it down. CelebA-1k dataset appears to be significantly more susceptible to degradation. Importantly, using only $5\%$ and $2\%$ synthetic images is sufficient to trigger model collapse in Pokemon and CelebA-1k datasets. This demonstrates the catastrophic impact of synthetic data on diffusion finetuning and raises the question of whether synthetic data detection can be a safe solution.

\subsection{Prompt Set}
\label{appendix:hyperparameter_experiments:prompt}

\begin{figure}[h!]
    \centering
    \includegraphics[width=0.6\textwidth]{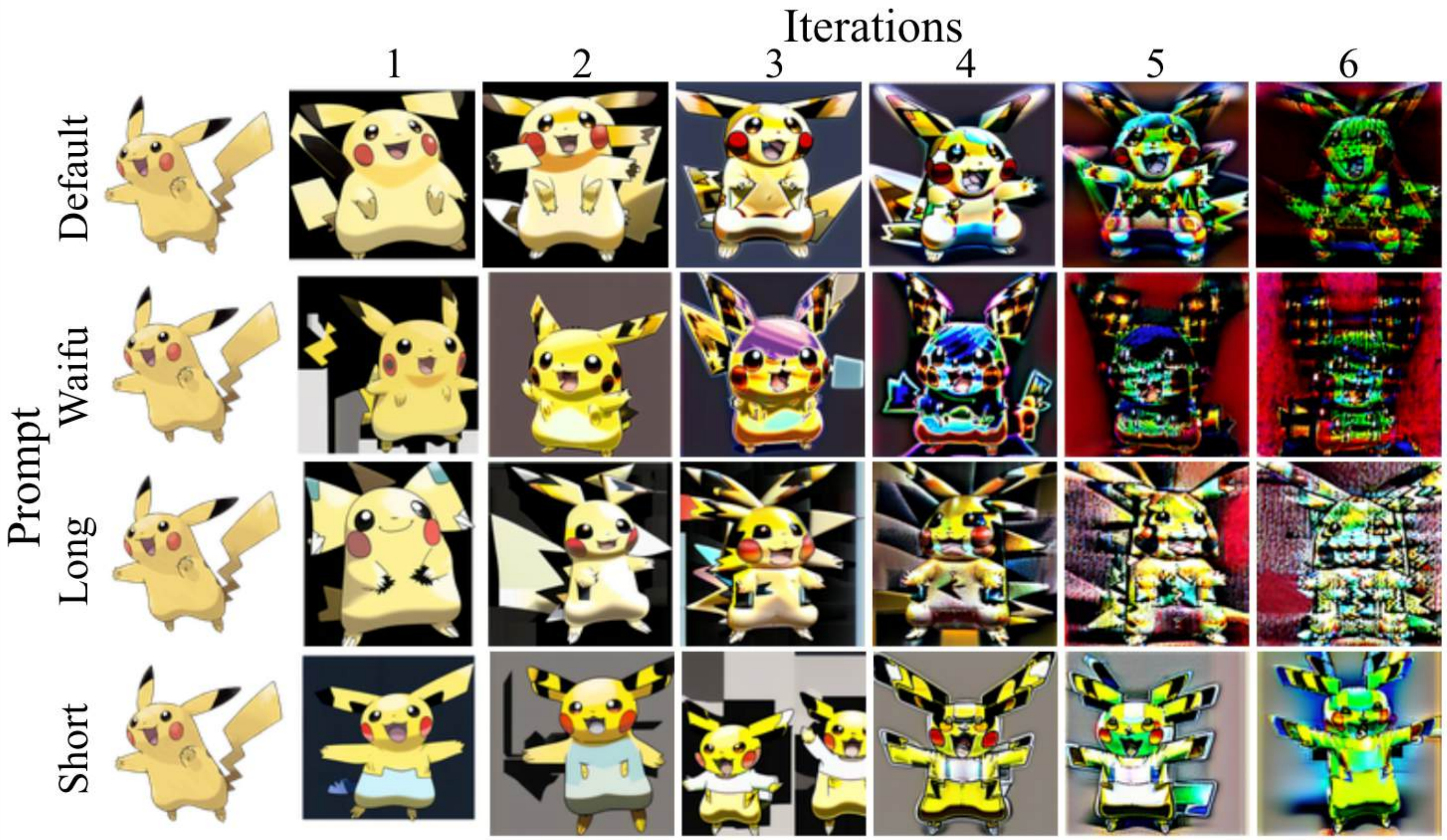}
    \caption{
    Chain of Diffusion on Pokemon dataset with different prompts. The default prompts are concatenations of BLIP captions ($50$-$75$ words) and Waifu captions. We compare the Chain of Diffusion using default captions, Waifu captions, short (less than $25$ words) and long ($50$-$75$ words) BLIP captions.
    }
    \label{figure:hyperparameters_experiments:prompts:pokemon}
\end{figure}

\begin{figure}[h!]
    \centering
    \includegraphics[width=0.6\textwidth]{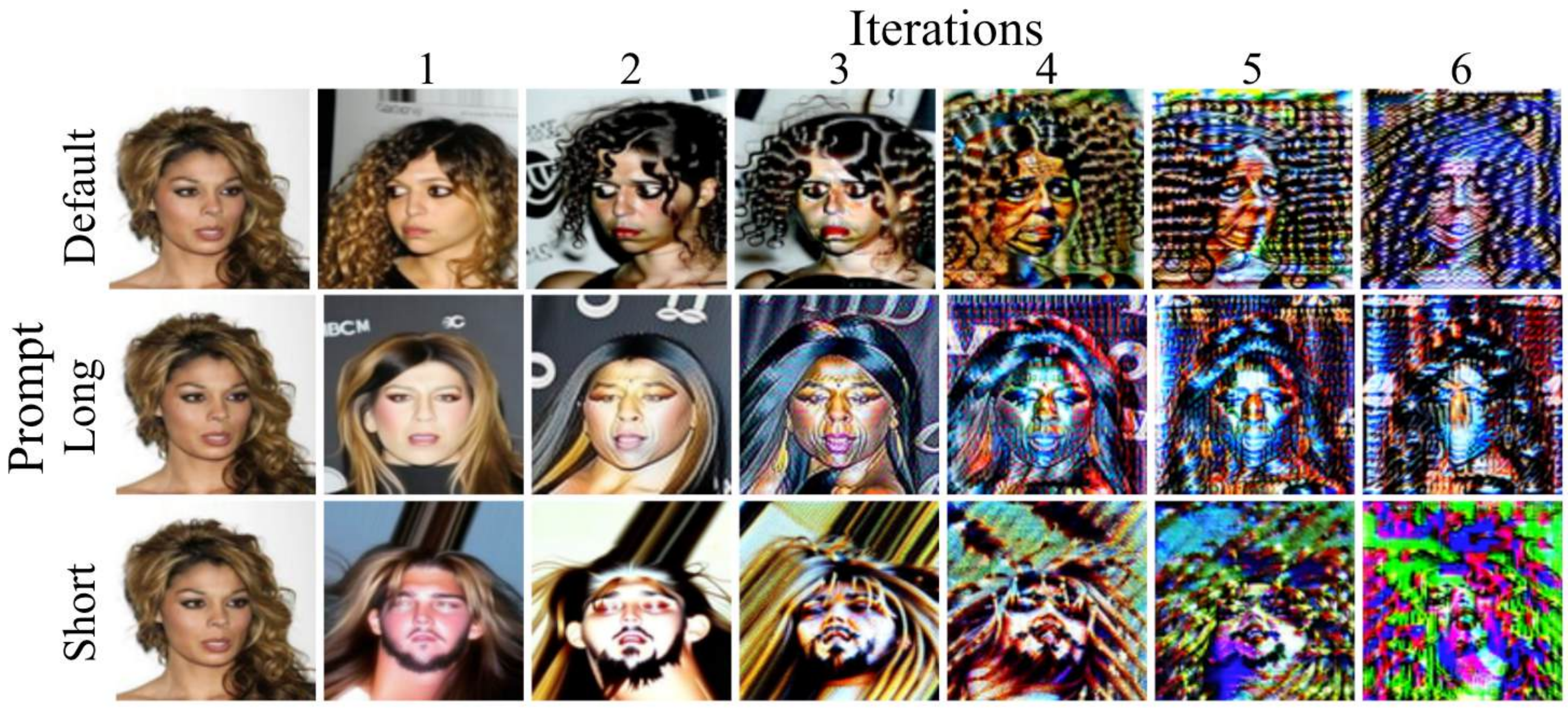}
    \caption{
    Chain of Diffusion on CelebA-1k dataset with different prompts. The default prompts range from $25$ to $50$ words. We compare the Chain of Diffusion using longer prompts (over $50$ words) and shorter prompts (under $25$ words).
    }
    \label{figure:hyperparameters_experiments:prompts:celeba}
\end{figure}

We hypothesized that the descriptiveness of prompts influences degradation in the Chain of Diffusion. We tested various prompt sets for Pokemon and CelebA-1k datasets. For Pokemon dataset, the default prompt set consists of concatenated Waifu and BLIP captions, with BLIP captions ranging from $50$ to $75$ words. Figure~\ref{figure:hyperparameters_experiments:prompts:pokemon} shows how using only Waifu prompts and varying BLIP caption lengths affect degradation. Notably, different styles of high-frequency degradation were observed; shorter prompts reduced repetitive patterns but decreased diversity. In CelebA-1k dataset, varying BLIP caption lengths resulted in similar degradation levels, but prompts that were either insufficiently or excessively descriptive caused the images to deviate from the originals, as shown in Figure~\ref{figure:hyperparameters_experiments:prompts:celeba}.

\subsection{U-Net and Text-Encoder}
\label{appendix:hyperparameter_experiments:unet_text_encoder}

\begin{figure}[h!]
    \centering
    \includegraphics[width=0.6\textwidth]{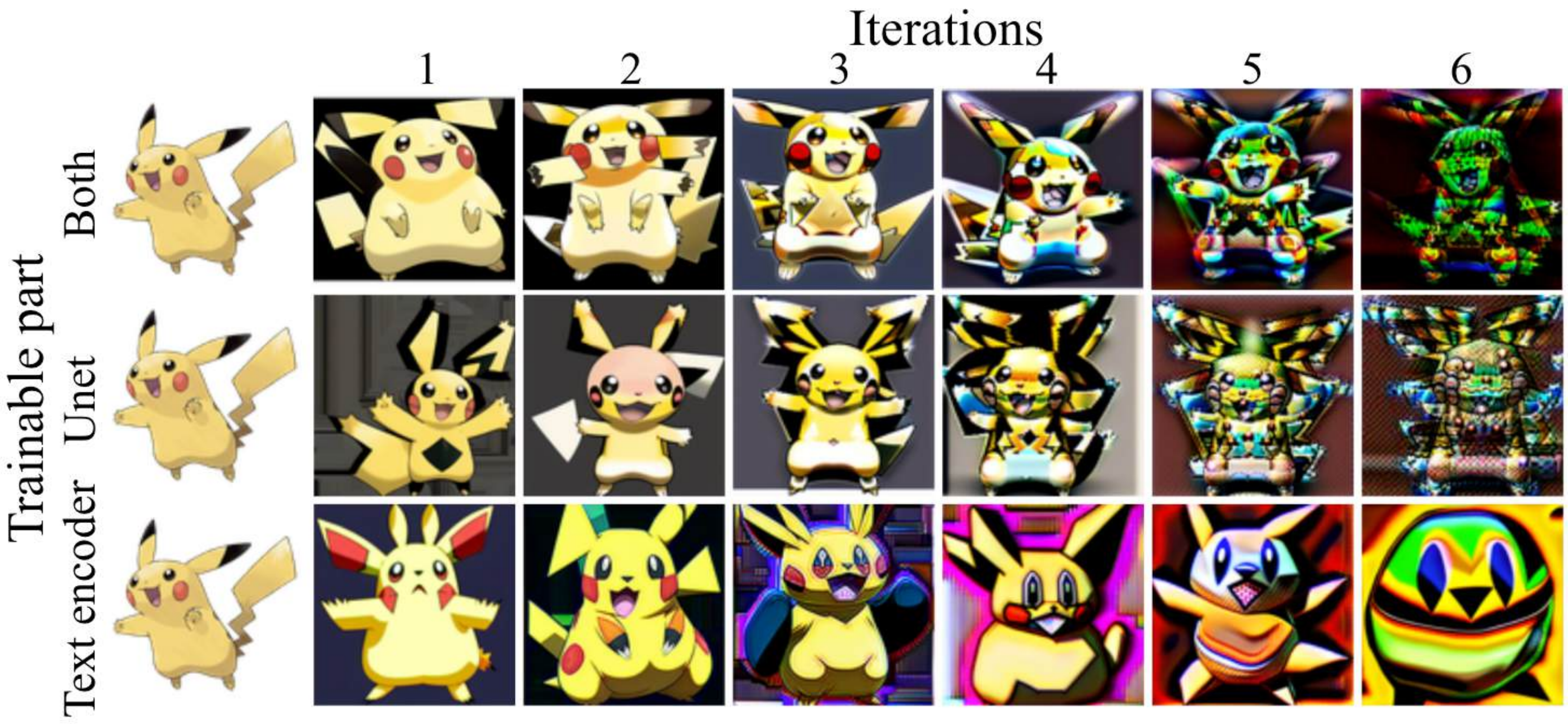}
    \caption{
    Chain of Diffusion on Pokemon dataset with either the U-Net or text encoder finetuned.
    }
    \label{figure:hyperparameters_experiments:unet_text_encoder}
\end{figure}

Figure~\ref{figure:hyperparameters_experiments:unet_text_encoder} illustrates the Chain of Diffusion with either the U-Net or text encoder finetuned. When the text encoder is not updated (second row), similar degradation occurs. However, the degradation pattern changes when the U-Net is not updated, as the model's ability to generate images remains unchanged. In contrast, updating the text encoder results in a loss of image content preservation.

\subsection{Number of diffusion steps}
\label{appendix:hyperparameter_experiments:steps}

We investigated whether an insufficient number of diffusion steps during generation contributes to degradation. Figure~\ref{figure:hyperparameters_experiments:steps_epochs} shows that increasing the number of diffusion steps does not enhance the Chain of Diffusion.

\subsection{Number of Epochs}
\label{appendix:hyperparameter_experiments:epochs}

\begin{figure}[h!]
    \centering
    \includegraphics[width=0.6\textwidth]{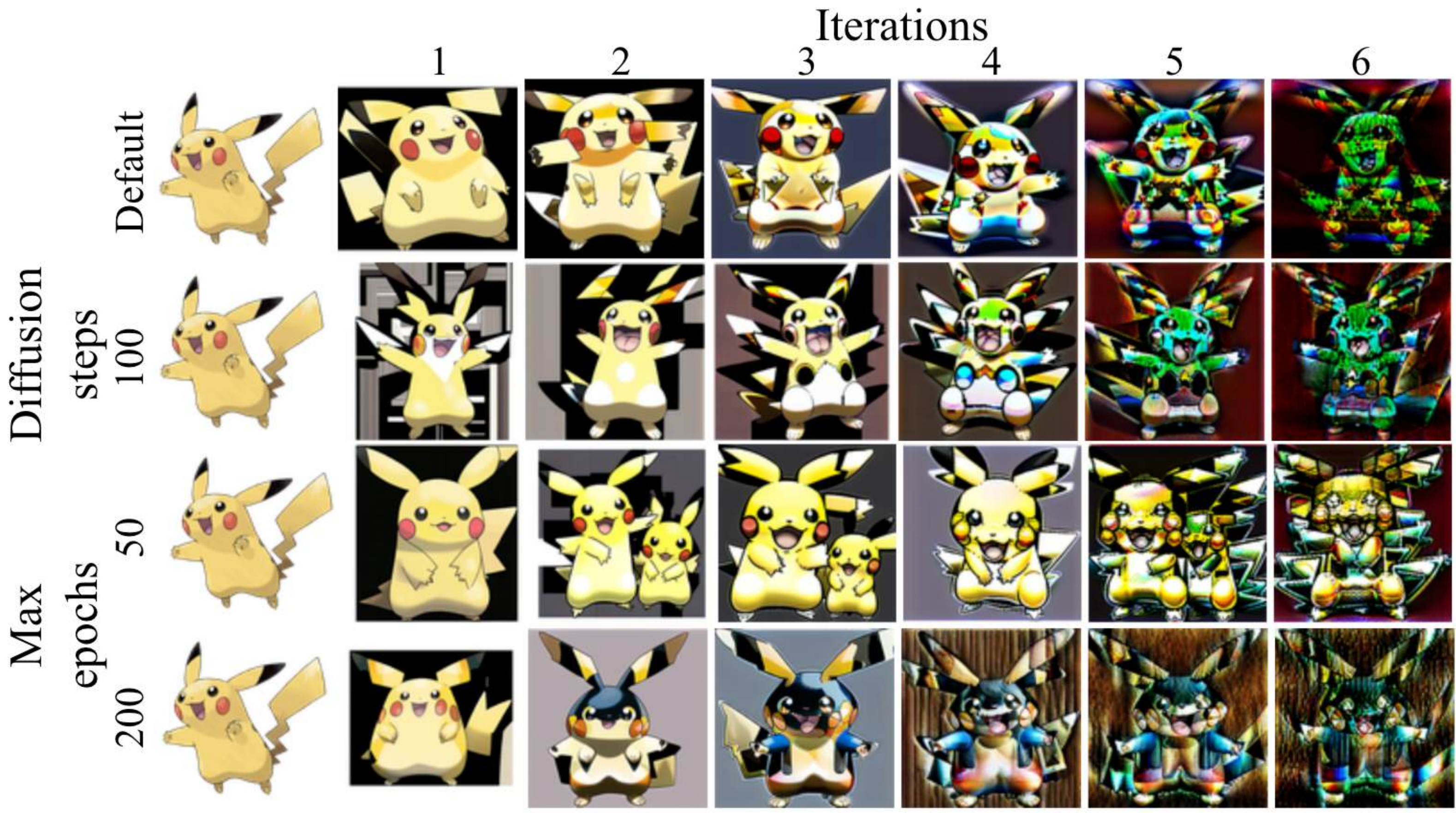}
    \caption{
    Chain of Diffusion on Pokemon dataset with varying diffusion steps and training epochs. Both increased diffusion steps and differing training epochs fail to mitigate degradation, resulting in similar patterns.
    }
    \label{figure:hyperparameters_experiments:steps_epochs}
\end{figure}

We also examined whether insufficient or excessive training affects our default setting. As shown in Figure~\ref{figure:hyperparameters_experiments:steps_epochs}, images from the initial iterations exhibit similar quality, resulting in comparable degradations. We set the default finetuning to $100$ epochs since loss values continue to decrease after $50$ epochs. 

\subsection{Learning Rate}
\label{appendix:hyperparameter_experiments:lr}

\begin{figure}[h!]
    \centering
    \includegraphics[width=0.6\textwidth]{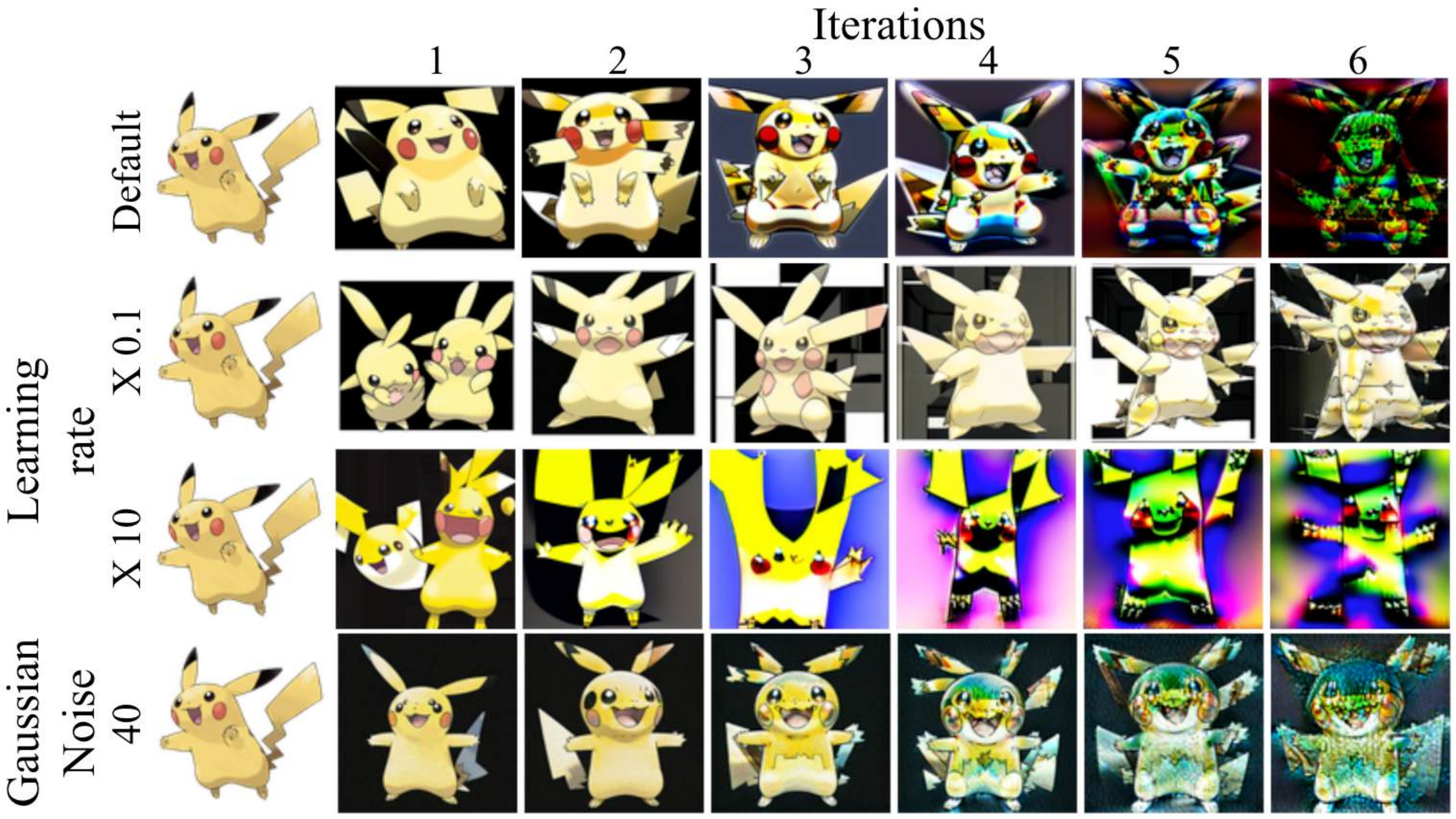}
    \caption{
    Chain of Diffusion on Pokemon dataset with varying learning rates and added Gaussian noise to the original training set.
    }
    \label{figure:hyperparameters_experiments:lr_noise}
\end{figure}

Similarly, we assessed finetuning adequacy in Figure~\ref{figure:hyperparameters_experiments:lr_noise} by adjusting the learning rates for the U-Net and text encoder by x$10$ and x$0.1$. Images from the initial iterations show that the default values are suitable for finetuning. Although the styles are different, degradation consistently occurs across different learning rates.

\subsection{CLIP Skip}
\label{appendix:hyperparameter_experiments:clip_skip}

\begin{figure}[h!]
    \centering
    \includegraphics[width=0.6\textwidth]{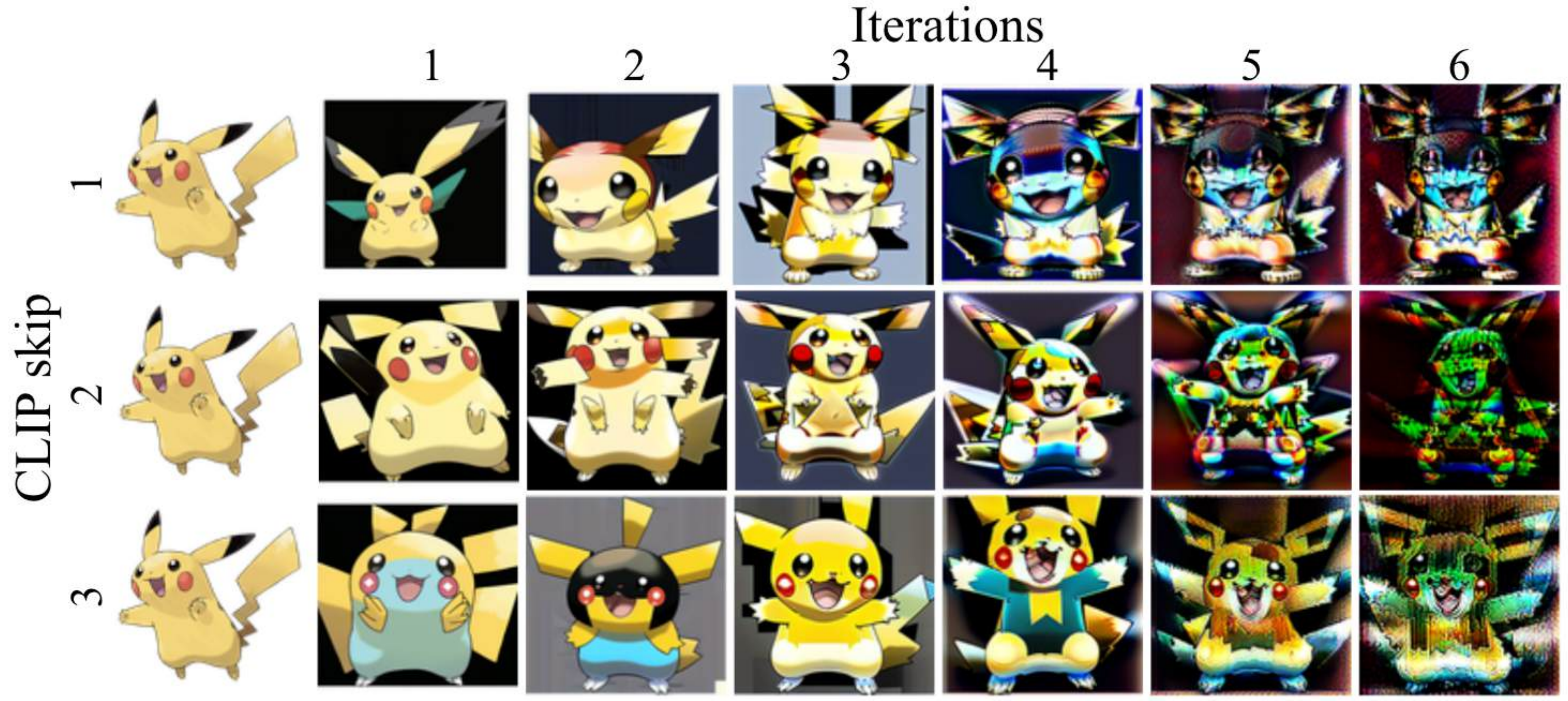}
    \caption{
    Chain of Diffusion on Pokemon dataset using different CLIP skip hyperparameters. The CLIP skip hyperparameter shows a negligible effect on degradation.
    }
    \label{figure:hyperparameters_experiments:clip_skip}
\end{figure}

We investigated the impact of the CLIP skip hyperparameter on degradation. The CLIP skip determines which intermediate feature from the CLIP text encoder is used as the text embedding for conditional generation, with smaller values selecting features closer to the output and larger values selecting those nearer to the input text. As shown in Figure~\ref{figure:hyperparameters_experiments:clip_skip}, this hyperparameter has minimal effect on degradation patterns.

\subsection{Adding Gaussian noise to the original training set}
\label{appendix:hyperparameter_experiments:input_noise}

We examined whether differences between real and synthetic images contribute to degradation by adding random Gaussian noise to the original training set $D_0$. Figure~\ref{figure:hyperparameters_experiments:lr_noise} illustrates that while the characteristics of the original training set have some effect, degradation still occurs. This supports our findings that degradations are universal across real, animation, and handwritten images.

\subsection{Stable Diffusion XL}
\label{appendix:hyperparameter_experiments:sdxl}

\begin{figure}[h!]
    \centering
    \includegraphics[width=0.6\textwidth]{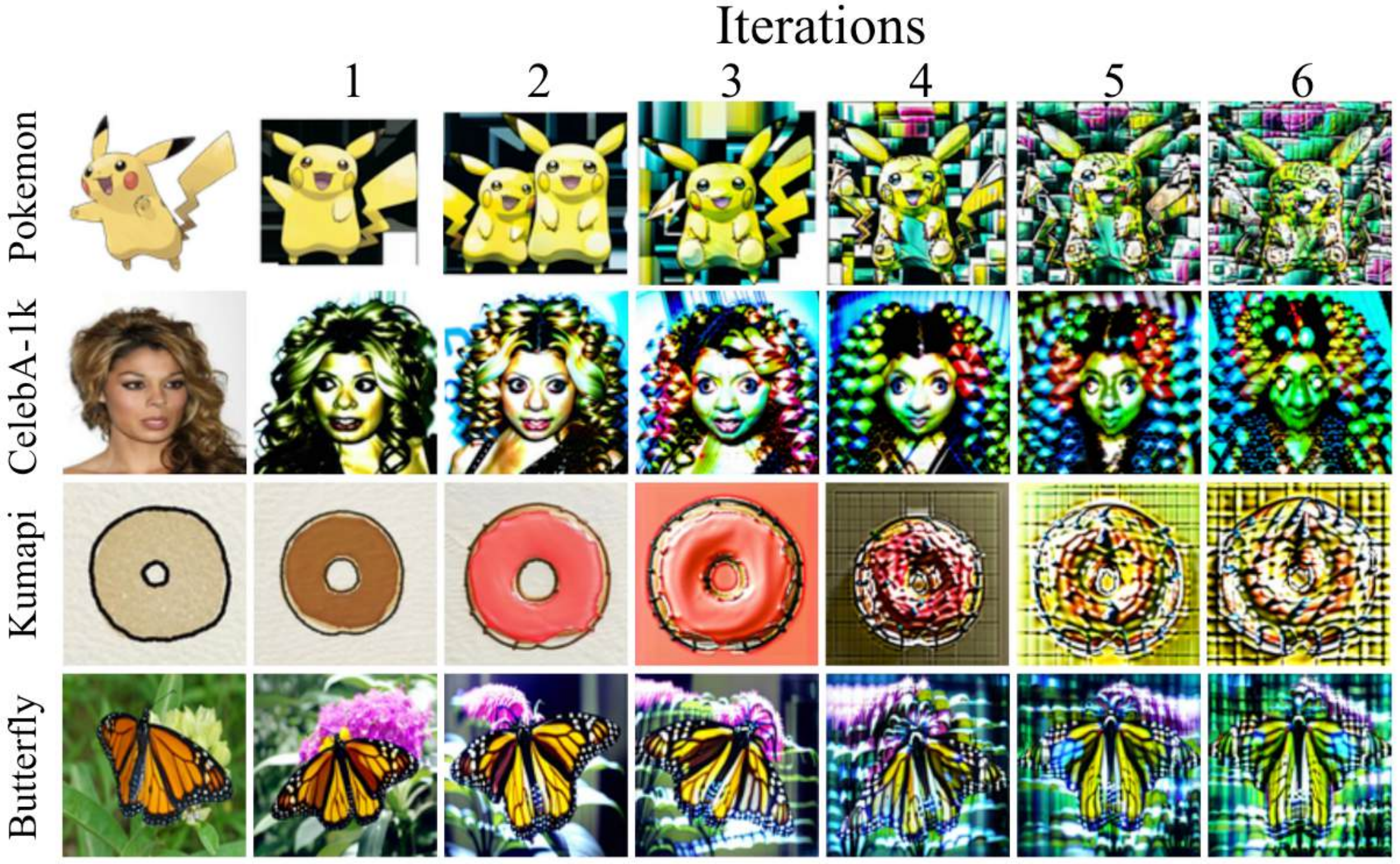}
    \caption{
    Chain of Diffusion of SDXL on four datasets. Due to insufficient investigation into optimal hyperparameters for finetuning SDXL, our experiments largely rely on those from Stable Diffusion v1.5.
    }
    \label{figure:hyperparameters_experiments:sdxl}
\end{figure}

We investigated image degradation for a different Stable Diffusion model, noting that the optimal hyperparameters for finetuning SDXL using LoRA are not well established. Consequently, we applied the same hyperparameters used for Stable Diffusion v1.5, which may be suboptimal. To manage space complexity, we reduced the batch size to $2$ and maintained a resolution of $512 \times 512$, as the first iteration images exhibit impressive quality. Results are presented in Figure~\ref{figure:hyperparameters_experiments:sdxl}.

\subsection{Data Accumulation}
\label{appendix:hyperparameter_experiments:iteration_accumulation}

Data accumulation experiments aim to investigate whether concept overfitting and disappearing are major reasons for model collapse. The training set for iteration $t$ is the combination of all previously generated sets, including the original training set. The number of training epochs is controlled accordingly to maintain the total number of updates.

\begin{figure}[h!]
    \centering
    \includegraphics[width=0.6\textwidth]{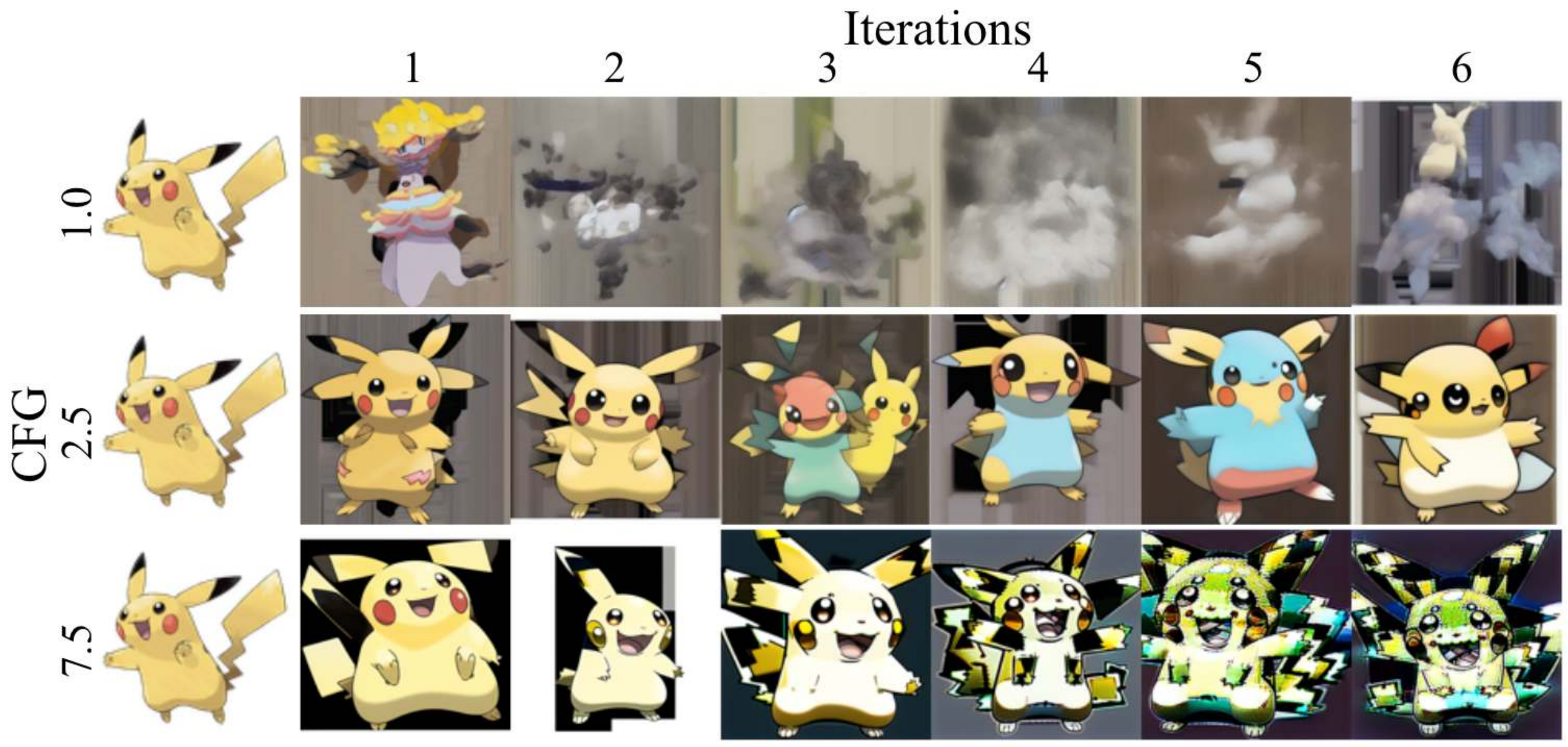}
    \caption{
    Chain of Diffusion on Pokemon dataset when training set is accumulated from previous iterations. All concepts from previous iterations are preserved for finetuning.
    }
    \label{figure:hyperparameters_experiments:iteration_accumulation:pokemon}
\end{figure}

\begin{figure}[h!]
    \centering
    \includegraphics[width=0.6\textwidth]{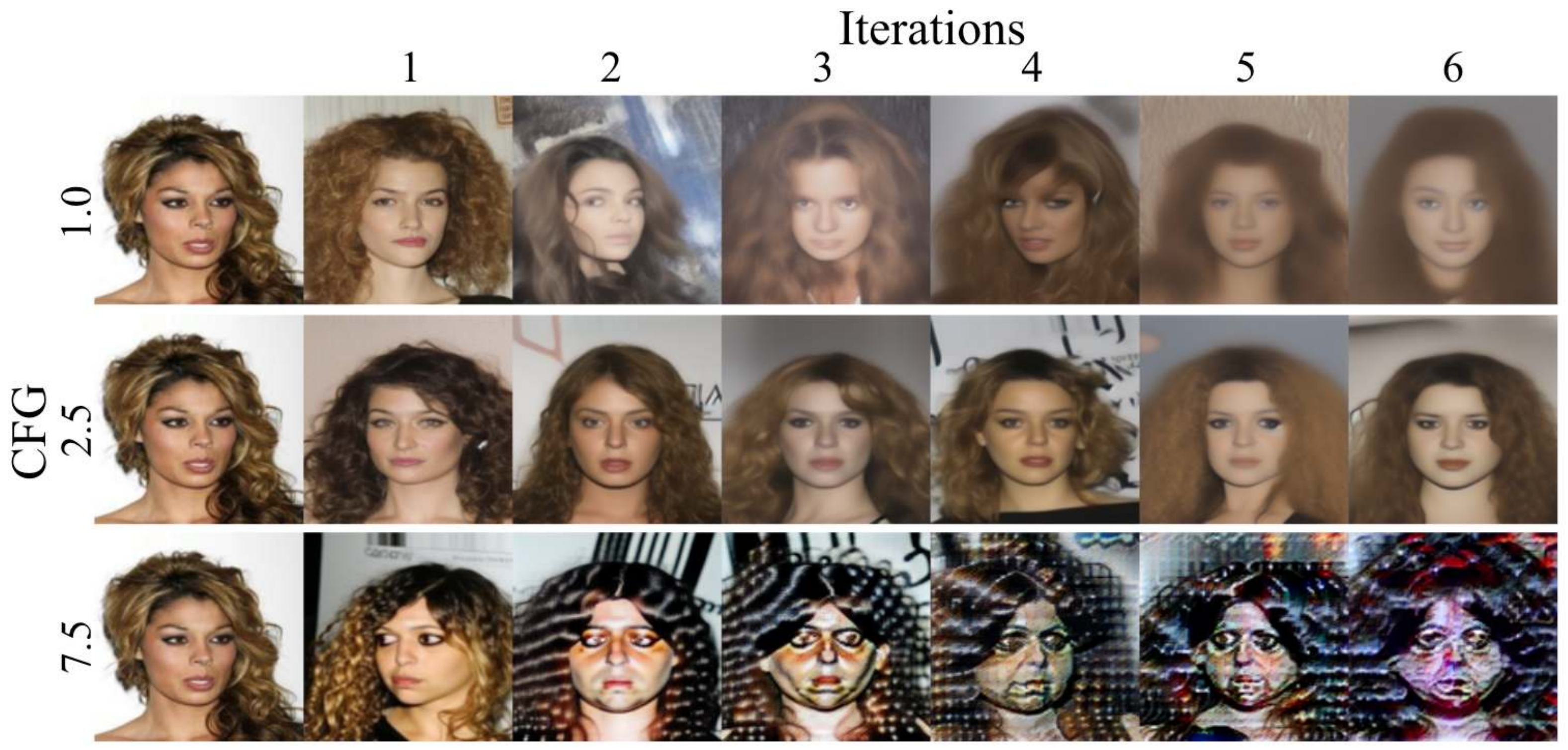}
    \caption{
    Chain of Diffusion on CelebA-1k dataset when training set is accumulated from previous iterations. All concepts from previous iterations are preserved for finetuning.
    }
    \label{figure:hyperparameters_experiments:iteration_accumulation:celeba}
\end{figure}

\section{Quantitative Trait Modeling}
\label{appendix:theory}

\subsection{Wright-Fisher model - Genetic drift and fixation}
\label{appendix:theory:wf}

\ys{The Wright-Fisher model describes the evolution of discrete allele frequencies in a finite population. We assume distinguishable iterations (no overlapping individuals) and no selection or mutation. In each iteration, all individuals from the previous iteration (the parents) are replaced by new individuals (the offspring). The offspring are sampled from the parental population via random sampling with replacement; for simplicity, we assume a haploid model. }

\ys{Consider a population of constant size $N$ and two alleles, $A$ and $a$. If allele $A$ is present in $i$ individuals in the parent interaction, the probability that it appears in $j$ individuals in the offspring iteration is given by the binomial distribution:
\begin{equation}
    P_{ij} = \binom{N}{j}({i\over N})^j(1 - {i\over N})^{N-j}, 0 \leq i, j \leq N.
\end{equation}
This process defines a Markov chain, where the state in each iteration depends only on the state in the previous one. The Wright-Fisher model thus characterizes the stochastic dynamics of allele frequencies across iterations.}

\ys{Certain states in a Markov chain are absorbing---that is, once entered, they cannot be exited. In the Wright-Fisher model, the fixation or loss of an allele corresponds to such absorbing states. Once the population becomes entirely composed of a single allele, the other allele cannot reappear under the assumptions of no mutation or migration. Importantly, in a finite population, the process is guaranteed to eventually reach one of these two absorbing states.The probability that allele $A$ becomes fixed depends solely on its initial frequency in the population.}

\subsection{Proof of Theorem~\ref{theorem:convergence}}
\label{appendix:theory:proof}

{\it{Proof.}} 
The difference between the successive means in quantitative trait modeling is given by:
\begin{equation}
    \Delta \mu_t = \mu_{t+1} - \mu_t = h_t^2 (\mu_t' - \mu_t).
\end{equation}
When Gaussian distribution with mean $\mu$ and variance $\sigma^2$ is truncated on both sides with $r_1$ and $r_2$ ratios, the mean and variance of truncated Gaussian distribution are expressed as:
\begin{equation}\label{equation:truncation_mean}
    \mu_{trun} = \mu - \frac{\varphi(\beta) - \varphi(\alpha)}{\Phi(\beta) - \Phi(\alpha)} \sigma, 
\end{equation}
\begin{equation}\label{equation:truncation_variance}
    \sigma_{trun}^2 = (1 - \frac{\beta \varphi(\beta) - \alpha \varphi(\alpha)}{\Phi(\beta) - \Phi(\alpha)} - (\frac{\varphi(\beta) - \varphi(\alpha)}{\Phi(\beta) - \Phi(\alpha)})^2 ) \sigma^2, 
\end{equation}
where $\varphi$ and $\Phi$ are the probability density function (PDF) and cumulative distribution function (CDF) of the standard normal distribution, and $\alpha = \Phi^{-1}(r_1)$ and $\beta = \Phi^{-1}(1 - r_2)$. Accordingly, given $\mu = \mu_t$ and $\sigma^2 = \sigma_{P, t}^2$ with $\mu_t' = \mu_{trun}$ and $\sigma_{G, t+1}^2 = \sigma_{P, t}'^2 = \sigma_{trun}^2$, we have
\begin{equation}\label{equation:delta_mu}
    \Delta \mu_t = h_t^2 (\mu_t' - \mu_t) = \frac{\sigma_{G, t}^2}{\sigma_{P, t}^2} c_1 \sigma_{P, t} = c_1 \frac{\sigma_{P, t-1}'^2}{\sigma_{P, t}^2} \sigma_{P, t} = c_1 c_2 \frac{\sigma_{P, t-1}^2}{\sigma_{P, t}^2} \sigma_{P, t},
\end{equation}
where $c_1 = | \frac{\varphi(\beta) - \varphi(\alpha)}{\Phi(\beta) - \Phi(\alpha)} |$ and $c_2 = 1 - \frac{\beta \varphi(\beta) - \alpha \varphi(\alpha)}{\Phi(\beta) - \Phi(\alpha)} - (\frac{\varphi(\beta) - \varphi(\alpha)}{\Phi(\beta) - \Phi(\alpha)})^2$ when $r_1 > r_2$. On the other hand, the mean phenotypes decreases when $r_1 < r_2$ as:
\begin{equation}
    \Delta \mu_t = - c_1 c_2 \frac{\sigma_{P, t-1}^2}{\sigma_{P, t}^2} \sigma_{P, t}.
\end{equation}
Furthermore, the phenotype variance converges over time as:
\begin{equation}
    \sigma_{P, t}^2 = \sigma_{G, t}^2 + \sigma_E^2 = \sigma_{P, t-1}'^2 + \sigma_E^2 = c_2 \sigma_{P, t-1}^2 + \sigma_E^2.
\end{equation}
\begin{equation}\label{equation:sigma_converge}
    \sigma_{P, t}^2 - \frac{\sigma_E^2}{1 - c_2} = c_2 (\sigma_{P, t-1}^2 - \frac{\sigma_E^2}{1 - c_2}) = \cdots = c_2^{t} (\sigma_{P, 0}^2 - \frac{\sigma_E^2}{1 - c_2}).
\end{equation}
As a result, the phenotype variance converges to $\frac{\sigma_E^2}{1 - c_2}$ for $0 < c_2 < 1$ (the variance of truncated distribution is smaller than the variance of the original distribution), and the mean asymptotically increases (decreases) by $\frac{c_1 c_2}{\sqrt{1 - c_2}} \sigma_E$ per iteration when $r_1 > r_2$ ($r_2 > r_1$). \hfill $\square$

\subsection{Simulation setup}
\label{appendix:theory:setup}

Our simulation aims to demonstrate that our experimental results can be modeled using quantitative trait modeling. We show that the radial sum of power spectra of images is one of the phenotypes explained by our theoretical analysis.

\subsubsection{Computing the radial sum of power spectra.} 
\label{appendix:theory:setup:spectrum}

The power spectra of images are computed as the square of the magnitude of the 2D Fourier transform. Here, we demonstrate how to compute the radial sum of power spectra in order to use the sum of radial power spectra above a certain threshold. Given a set of images $\{ I_i \}$ where $i \in \{ 1, 2, \cdots, N \}$, the 2D Discrete Fourier Transform (DFT) of an image $I_i$ of size $M \times N$ is computed as:
\begin{equation}
    F_i(u, v) = \sum^{M-1}_{x=0} \sum^{N-1}_{y=0} I_i(x, y) e^{-2 \pi j (\frac{ux}{M} + \frac{vy}{N})}
\end{equation}
where $F_i(u, v)$ represents the frequency component at coordinates $(u, v)$. The power spectrum of an image $I_i$ at $(u, v)$ is the square of the magnitude of its Fourier transform $P_i(u, v) = |F_i(u, v)|^2$. We compute the radial sum of the power spectra using a norm of each frequency component $(u, v)$ as $r(u, v) = \sqrt{(\frac{u}{M})^2 + (\frac{v}{N})^2}$. Given the threshold frequency $\tau$, we compute the radial sum of the high-frequency power spectra of an image $I_i$ as
\begin{equation}
    S_i = \sum^{M-1}_{u=0} \sum^{N-1}_{v=0} P_i(u, v) \cdot {\rm I} (r(u, v) > \tau)
\end{equation}
where ${\rm I}( \cdot )$ is an indicator function. We compute the sum of high-frequency components because low-frequency components tend to be noisy and use the threshold frequency of $0.02$. Then, the total sum of power spectra for all images can be written as:
\begin{equation}
    S = \sum^{N}_{i=0} S_i = \sum^{N}_{i=0} \sum^{M-1}_{u=0} \sum^{N-1}_{v=0} P_i(u, v) \cdot {\rm I} (r(u, v) > \tau).
\end{equation}
Practically, we shift the Fourier transform maps so that the frequency norm larger than $\frac{1}{\sqrt{2}}$ is considered to be in the opposite direction. Code for detailed implementation comes from the official code of \cite{corvi2023intriguing}.

\begin{figure}[t!]
    \centering
    \includegraphics[width=0.6\textwidth]{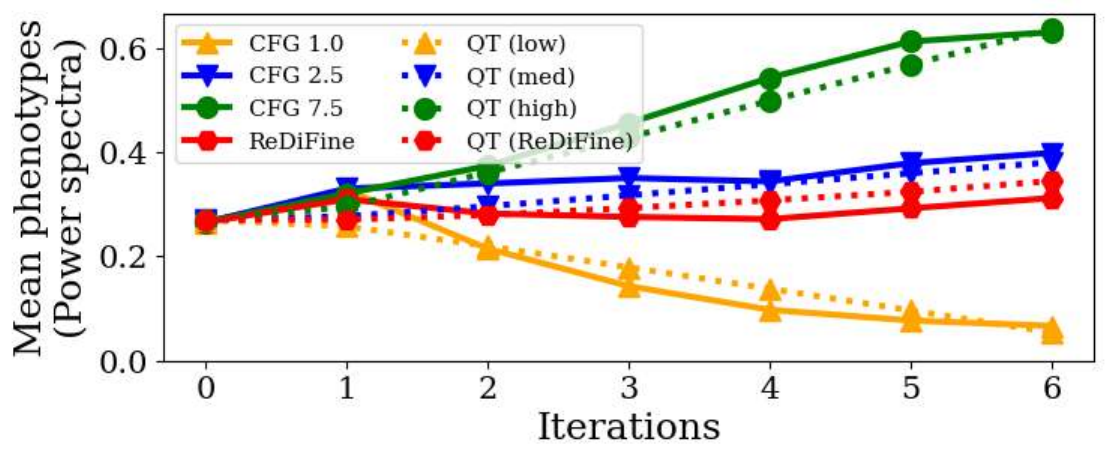}
    \caption{
    Results comparing the simulation for mutations and power spectra of images generated by ReDiFine (with CFG scale $7.5$). Power spectra and simulation are plotted in solid and dotted lines, respectively. Our modifications to heritability and selection process successfully demonstrate changes occur to images by ReDiFine.
    }
    \label{figure:simulation_mean_mutation}
\end{figure}

\subsubsection{Simulation parameters matching different CFG scales.} 
\label{appendix:theory:setup:parameters}

We simulate different selection strategies using truncation ratios $r_1$ and $r_2$. To model high, medium, and low CFG scales from our experiments, we apply $(0.025, 0.675)$, $(0.5, 0.09)$, and $(0.95, 0.0002)$ for $(r_1, r_2)$, respectively, which correspond to CFG scales of $7.5$, $2.5$, and $1.0$. For initial values in the simulation, we compute the mean ($0.027$) and standard deviation ($0.056$) from the original training set (iteration $0$), using them as the initial values for mean and genetic standard deviation for our simulation. We set the environmental standard deviation to $0.25$.

\subsubsection{ReDiFine and mutations.} 
\label{appendix:theory:setup:mutations}

ReDiFine combines condition drop finetuning and CFG scheduling, drawing inspiration from genetic mutation mechanisms that counter distributional shifts and help maintain genetic diversity in population genetics. \ys{In the Chain of Diffusion, training and generation can be viewed analogously to reproduction and natural survival. ReDiFine modifies both stages to preserve image quality across iterations.}

\ys{The two core components—condition drop finetuning and CFG scheduling—are rooted in biological analogies. Condition drop finetuning, akin to mutation, introduces controlled variability to ensure that the offspring distribution (the next iteration) remains aligned with the parent distribution (the previous iteration). This mitigates over-specialization (generated images becoming too uniform or isolated) and prevents degeneration under strong selection pressures. CFG scheduling, inspired by natural selection, dynamically adjusts the CFG scale during generation, promoting diversity by allowing broader phenotypic expression while retaining directed evolution toward desired traits. By integrating these biologically inspired strategies, ReDiFine improves the stability of the Chain of Diffusion and prevents catastrophic degradation across iterations.}

We apply two modifications to our theoretical analysis of Section~\ref{section:theory}---adding mutation variance to heritability and smoothing truncations---to simulate the effects of ReDiFine in the Chain of Diffusion. Specifically, we add the mutation standard deviation $\sigma_{M}$ of $0.1$ to heritability as:
\begin{equation}
    h_t^2 = \frac{\sigma_{G, t}^2}{\sigma_{P, t}^2 + \sigma_{M}^2} = \frac{\sigma_{G, t}^2}{\sigma_{G, t}^2 + \sigma_E^2 + \sigma_M^2},
\end{equation}
representing the randomness added to each iteration due to mutations. This influences the effects of previous iteration to current iteration, which reflect finetuning. Moreover, we apply exponential tails to truncations instead of cut-off thresholds where samples outside the truncation area can be randomly selected with exponential distribution ($e^{- \alpha d(x)}$ where $d(x)$ is a distance to truncation zone and we use $\alpha = 0.1$). This modified selection simulates the effect of CFG scheduling during image generation.

Figure~\ref{figure:simulation_mean_mutation} shows the effects of two modifications to our simulation results, and they closely align with the power spectra of images generated by ReDiFine. This demonstrates that the effects of ReDiFine can be understood as interference similar to mutations in population genetics. This suggests further research on model collapse motivated from other fields like biology.

\section{ReDiFine} 
\label{appendix:method}

\subsection{Visual Inspections}
\label{appendix:method:samples}

\begin{figure}[h!]
    \centering
    \includegraphics[width=0.6\textwidth]{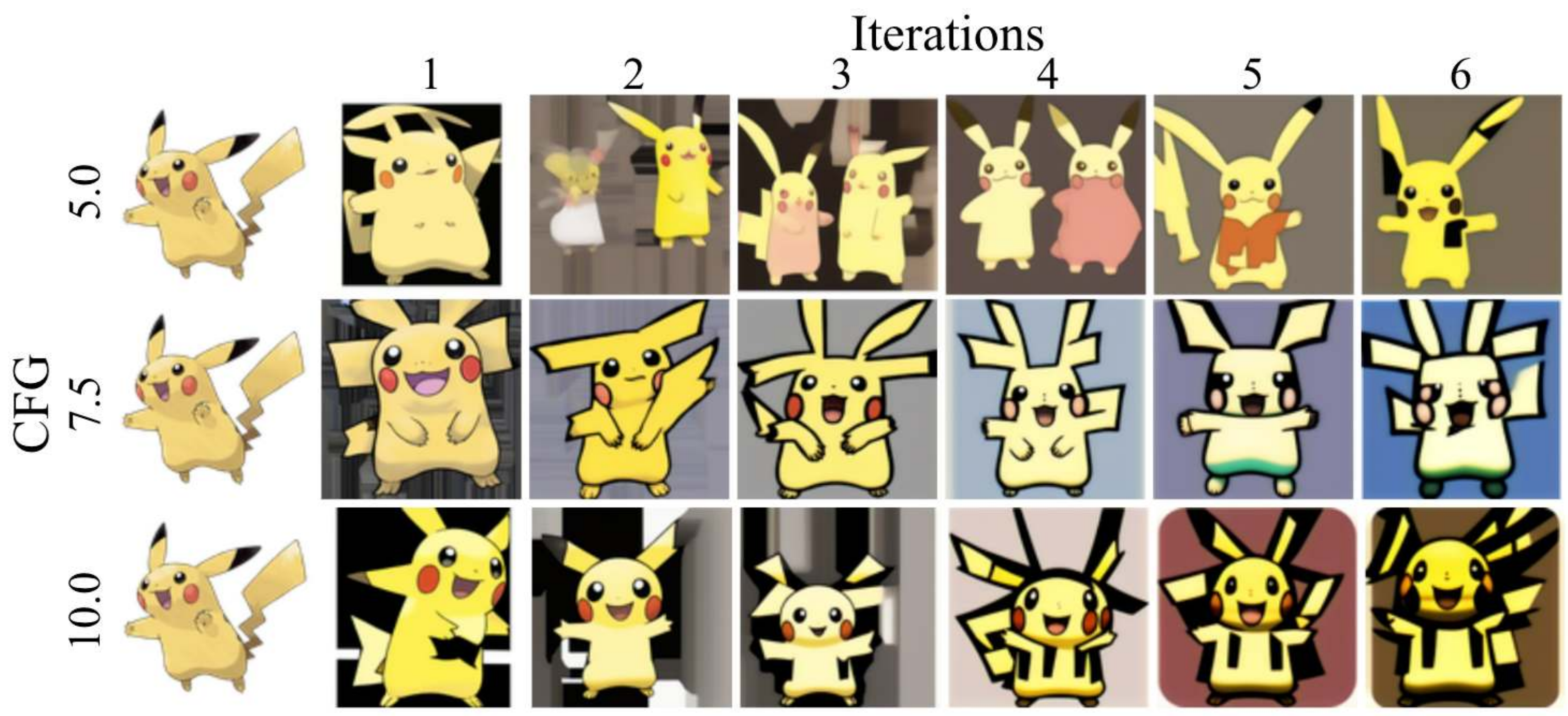}
    \caption{
    Chain of Diffusion of ReDiFine with different CFG scales on Pokemon dataset.
    ReDiFine successfully achieves robust image qualities for varying CFG scales.
    }
    \label{figure:method_experiments:pokemon}
\end{figure}

\begin{figure}[h!]
    \centering
    \includegraphics[width=0.6\textwidth]{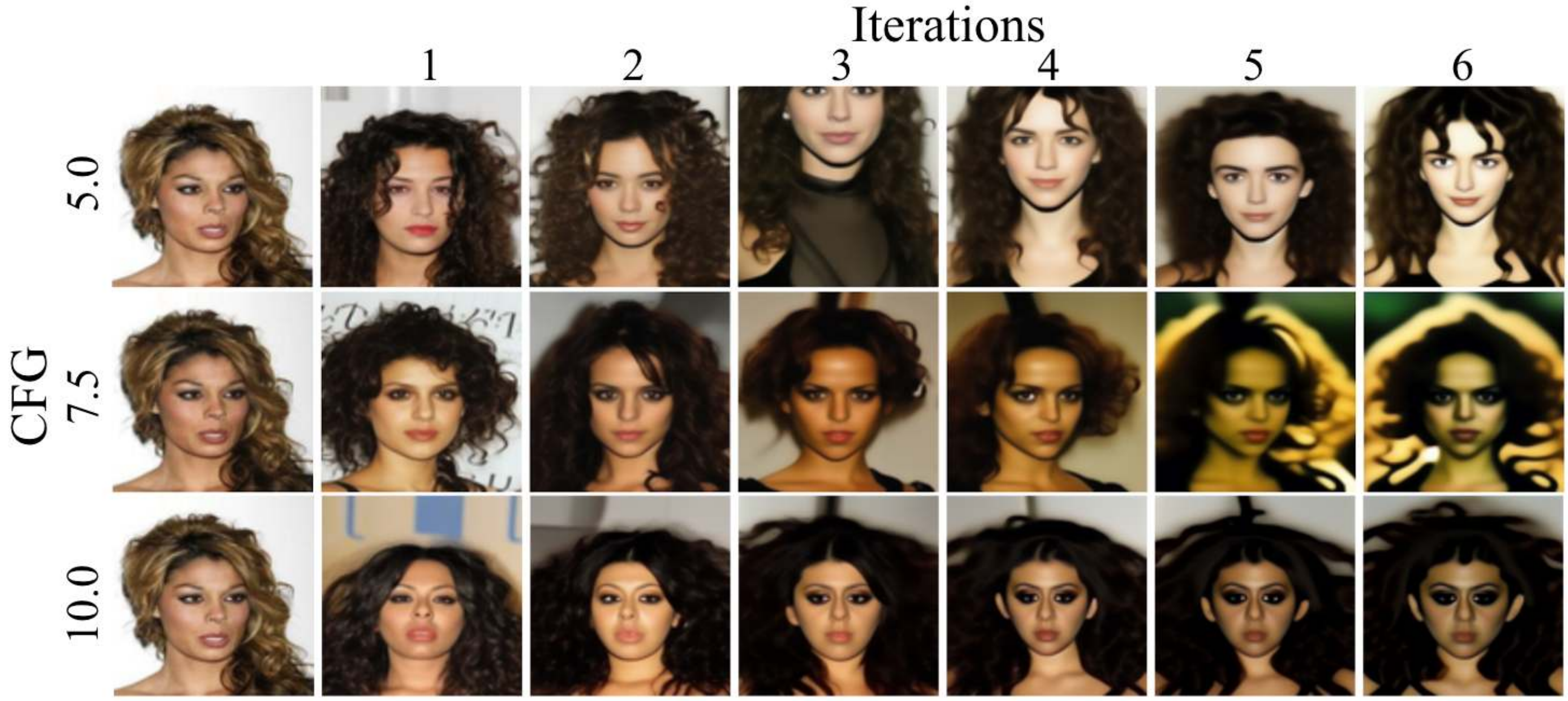}
    \caption{
    Chain of Diffusion of ReDiFine with different CFG scales on CelebA-1k dataset.
    ReDiFine successfully achieves robust image qualities for varying CFG scales.
    }
    \label{figure:method_experiments:celeba}
\end{figure}

\begin{figure}[h!]
    \centering
    \includegraphics[width=0.6\textwidth]{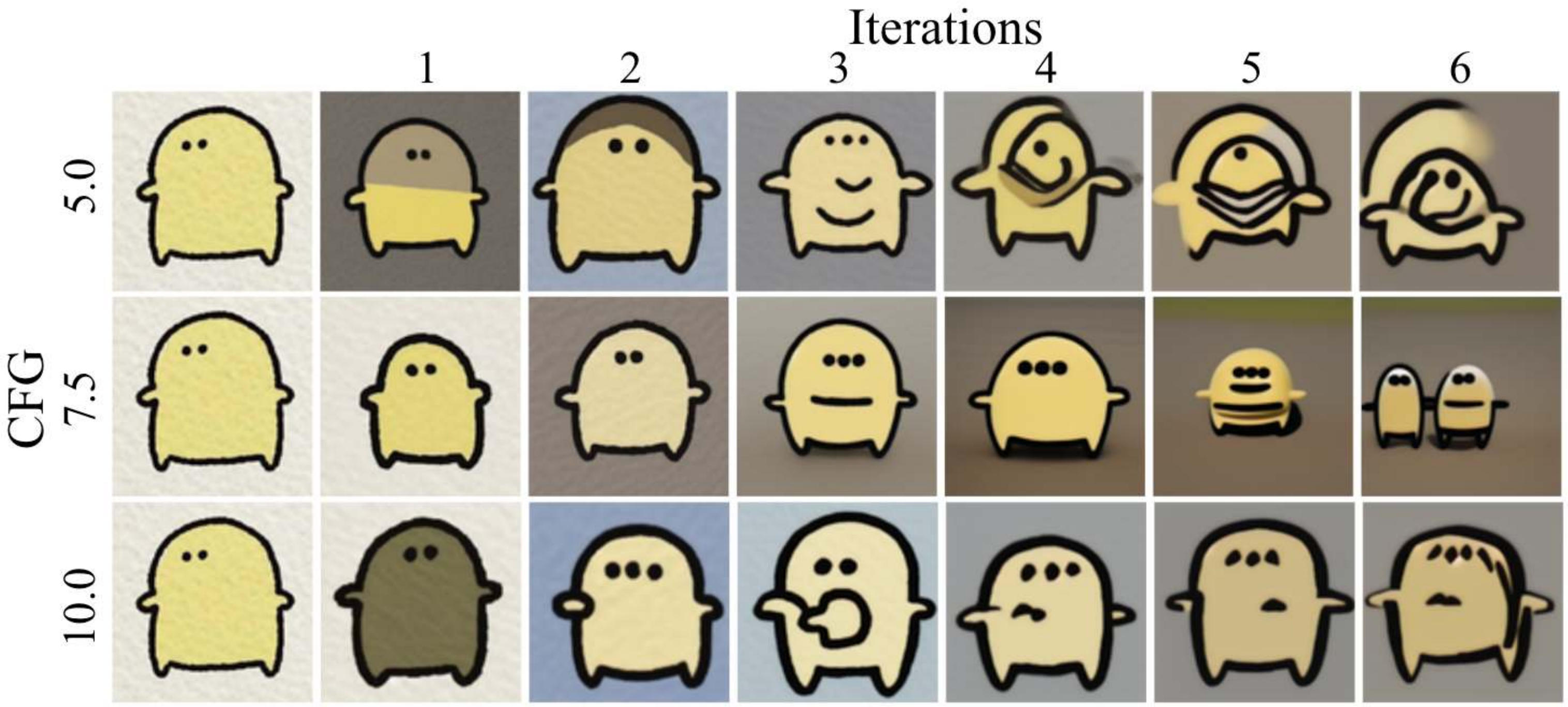}
    \caption{
    Chain of Diffusion of ReDiFine with different CFG scales on Kumapi dataset.
    ReDiFine successfully achieves robust image qualities for varying CFG scales.
    }
    \label{figure:method_experiments:kumapi}
\end{figure}

\begin{figure}[h!]
    \centering
    \includegraphics[width=0.6\textwidth]{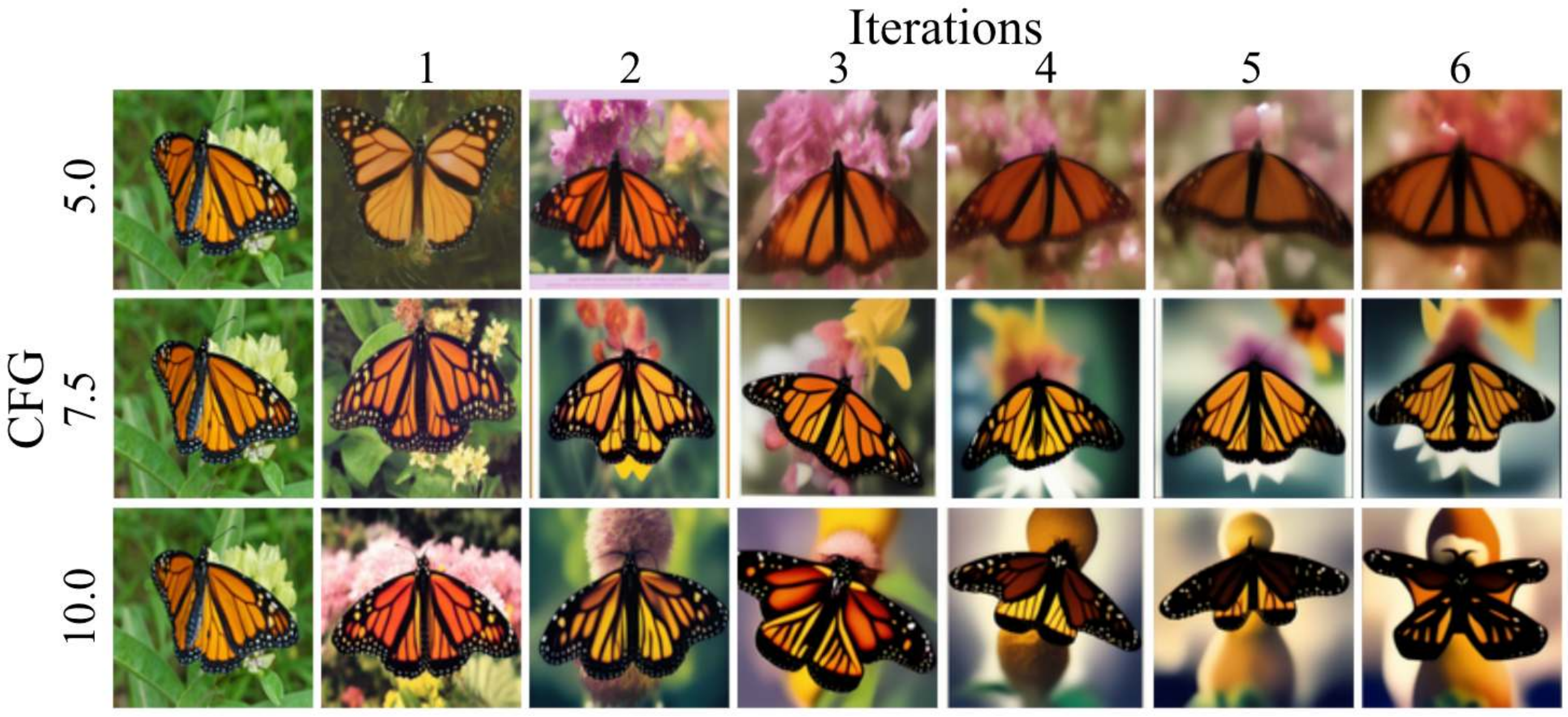}
    \caption{
    Chain of Diffusion of ReDiFine with different CFG scales on Butterfly dataset.
    ReDiFine successfully achieves robust image qualities for varying CFG scales.
    }
    \label{figure:method_experiments:butterfly}
\end{figure}

Figure~\ref{figure:method_experiments:pokemon}, \ref{figure:method_experiments:celeba}, \ref{figure:method_experiments:kumapi}, and \ref{figure:method_experiments:butterfly} show how robust ReDiFine is to different CFG scales. It successfully mitigates the high-frequency degradations for a wide range of CFG scales.

\subsection{Further iterations}
\label{appendix:method:more_iterations}

\begin{figure}[h!]
    \centering
    \includegraphics[width=0.99\textwidth]{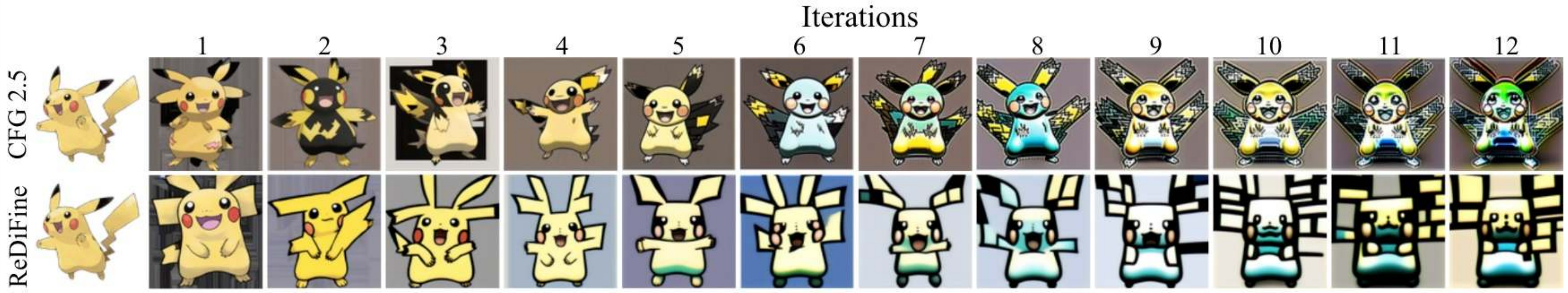}
    \caption{
    Chain of Diffusion of baseline with the optimal CFG scale and ReDiFine for more iterations. The generation quality of ReDiFine is preserved for additional iterations while optimally found CFG scale $2.5$ fails to maintain the image qualities. Similar high-frequency degradation is observed.
    }
    \label{figure:method_experiments:more_iterations}
\end{figure}

We conduct additional experiments to compare the baseline with the optimal CFG scale and ReDiFine over extended iterations. As shown in Figure~\ref{figure:method_experiments:more_iterations}, ReDiFine consistently generates images of similar quality up to 12 iterations, whereas the optimally tuned CFG scale fails to sustain image quality. This decline suggests that repeated hyperparameter searches are necessary to identify suitable CFG scales for subsequent iterations. Such an approach becomes increasingly impractical as the number of iterations grows, highlighting the limitations of relying on the optimal CFG scale to mitigate model collapse.

\subsection{Cross-domain Data}
\label{appendix:method:dataset_accumulation}

\begin{figure}[h!]
    \centering
    \includegraphics[width=0.99\textwidth]{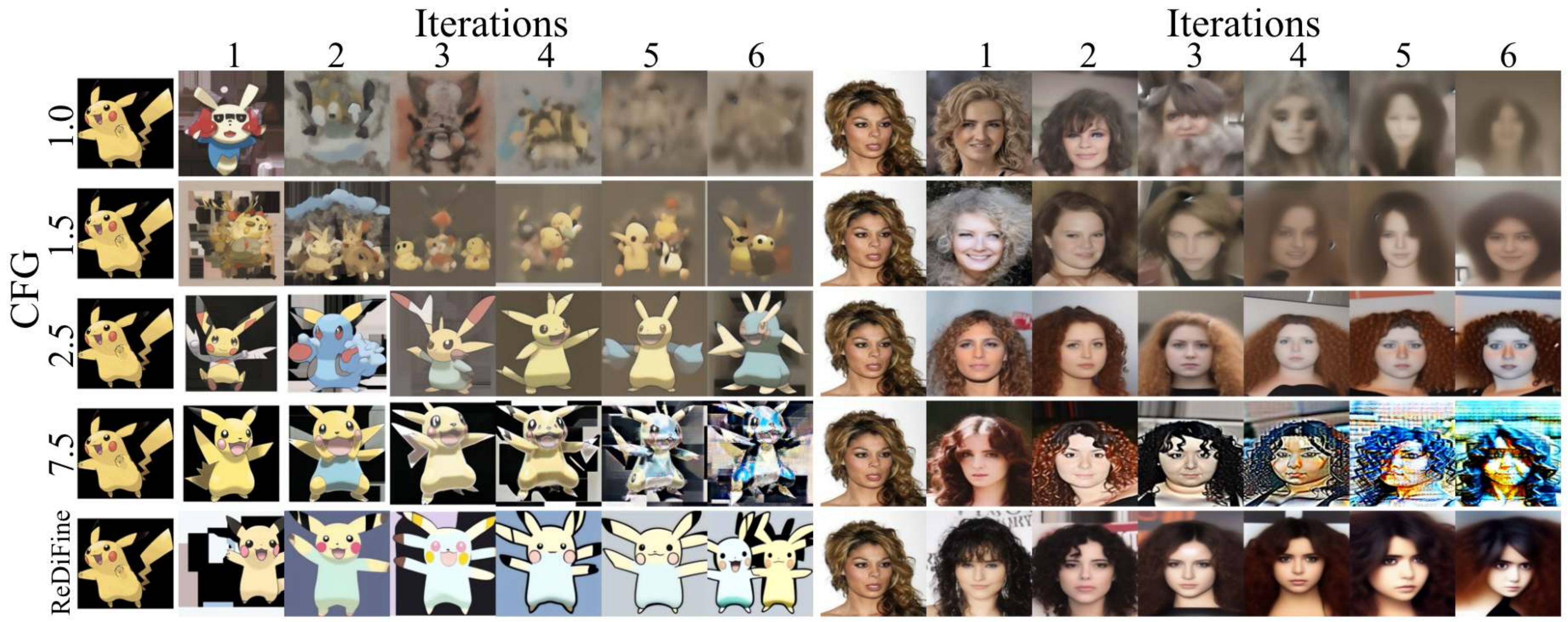}
    \caption{
    Chain of Diffusion with the multi-domain set of datasets (Pokemon, CelebA-1k, Kumapi, and Butterfly) used for finetuning. Model collapse remains evident at low and high CFG scales. A single CFG scale (e.g., $1.5$ or $2.5$) fails to achieve optimal performance across both the Pokemon and CelebA-1k datasets. In contrast, ReDiFine successfully generates high-quality, reusable images simultaneously. While images for Kumapi and Butterfly datasets are not displayed due to space constraints, they are included in the finetuning process along with the other datasets.
    }
    \label{figure:method_experiments:dataset_accumulation}
\end{figure}

We additionally conduct experiments when the training dataset is the cross-domain set of four datasets (Pokemon, CelebA-1k, Kumapi, and Butterfly) to investigate whether having a broader range of concepts impact model collapse. The merged dataset serves as the original training set, while the combined captions are used for image generation at each iteration. The results, presented in Figure~\ref{figure:method_experiments:dataset_accumulation}, show that despite the increased number of images and the inclusion of diverse concepts and domains, model collapse persists at both low and high CFG scales. Moreover, no single CFG scale (e.g., $1.5$ or $2.5$) can consistently produce high-quality, reusable images across all datasets, highlighting the limitations of relying on an optimal CFG scale for diverse domains. In contrast, ReDiFine leverages the increased conceptual diversity in the original training set, generating more reliable images across all datasets. To ensure a fair comparison, we control the number of training epochs to maintain consistent updates across experiments.

\subsection{Iterative Retraining}
\label{appendix:method:iterative_retraining}

\begin{figure}[h!]
    \centering
    \includegraphics[width=0.6\textwidth]{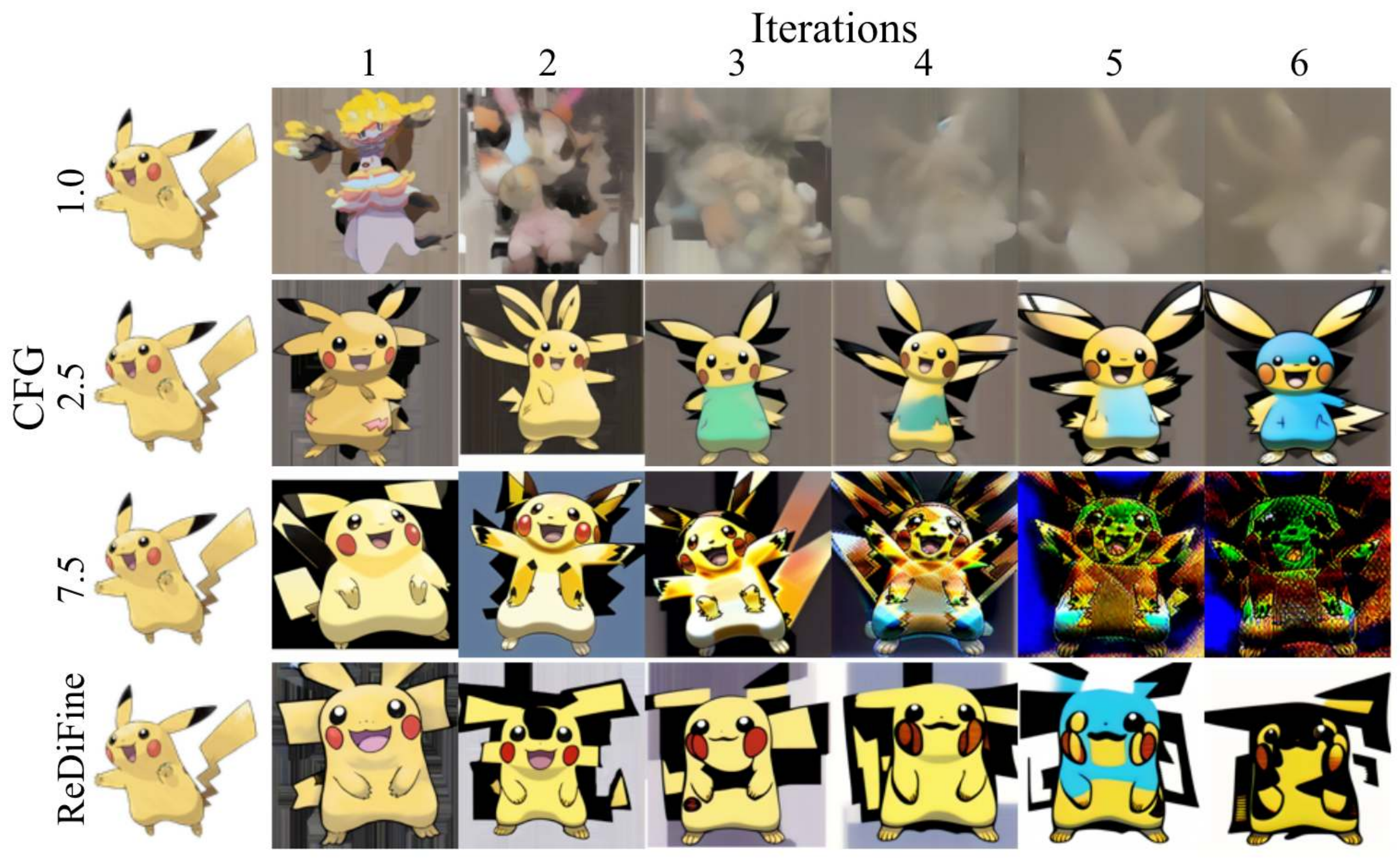}
    \caption{
    Chain of Diffusion where a single model is continually finetuned across multiple iterations. Model collapse is observed consistently at CFG scales of 1.0 and 7.5, while the baseline with a CFG scale of 2.5 and ReDiFine effectively mitigate model collapse. This demonstrates that model collapse is a universal phenomenon across different settings, and the effect of ReDiFine is effective robustly.
    }
    \label{figure:method_experiments:retraining}
\end{figure}

Some prior works on model degeneration examine scenarios in which a single model is continually trained on synthetic data it has generated. To adapt our Chain of Diffusion framework to this setting, we consider a setup where the same model is finetuned iteratively across multiple iterations. At each iteration, the model generates a fixed number of images using a predefined prompt set, and these generated images are then used to further finetune the model. Figure~\ref{figure:method_experiments:retraining} demonstrates how images degrade under this setting across different CFG scales and ReDiFine. While severe degradation is observed for low and high CFG scales, the optimal CFG scale and ReDiFine are able to mitigate model collapse, generating high-quality images. This indicates that the effect of ReDiFine is maintained even when a single model is continually finetuned.

\subsection{Quantitative results}
\label{appendix:method:quantitative}

\begin{figure}[t!]
    \centering
    \begin{subfigure}[b]{0.22\textwidth}
        \centering
        \includegraphics[width=1.0\textwidth]{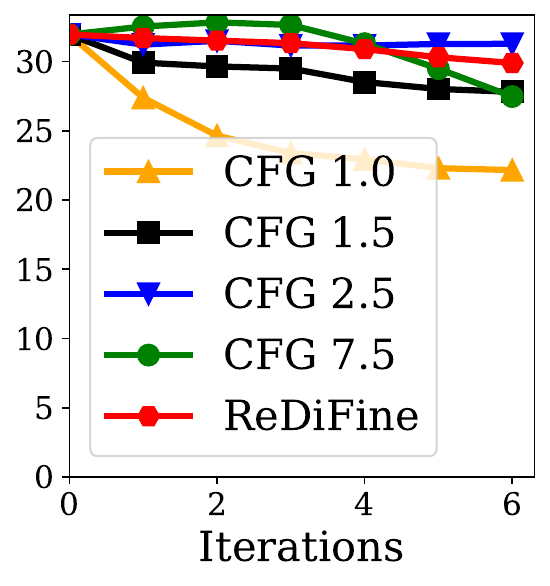}
        \subcaption{Pokemon, CLIP.}
        \label{figure:metric:pokemon:clip}
    \end{subfigure}
    \begin{subfigure}[b]{0.22\textwidth}
        \centering
        \includegraphics[width=1.0\textwidth]{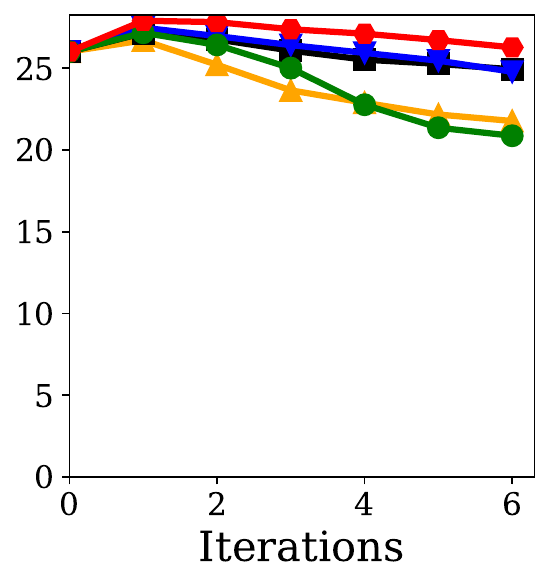}
        \subcaption{CelebA-1k, CLIP.}
        \label{figure:metric:celeba:clip}
    \end{subfigure}
    \begin{subfigure}[b]{0.22\textwidth}
        \centering
        \includegraphics[width=1.0\textwidth]{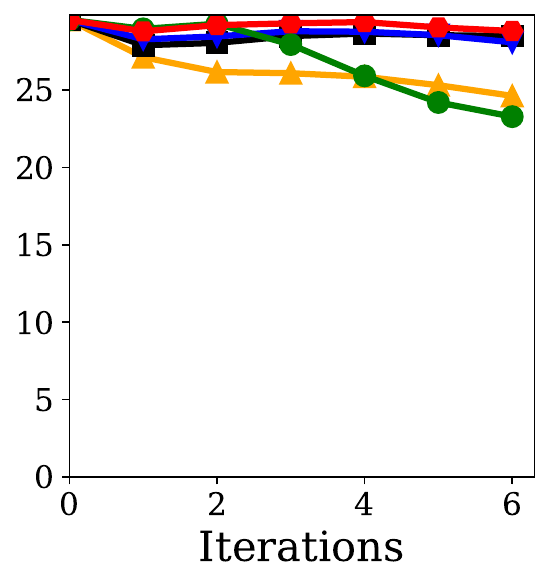}
        \subcaption{Kumapi, CLIP.}
        \label{figure:metric:kumapi:clip}
    \end{subfigure}
    \begin{subfigure}[b]{0.22\textwidth}
        \centering
        \includegraphics[width=1.0\textwidth]{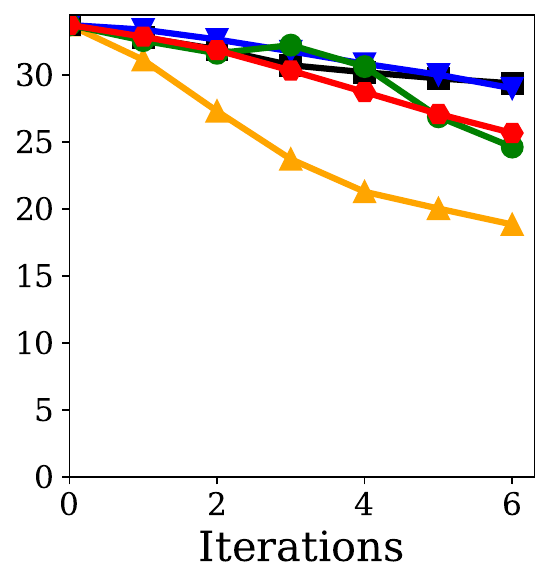}
        \subcaption{Butterfly, CLIP.}
        \label{figure:metric:butterfly:clip}
    \end{subfigure}
    \hfill
    \begin{subfigure}[b]{0.22\textwidth}
        \centering
        \includegraphics[width=1.0\textwidth]{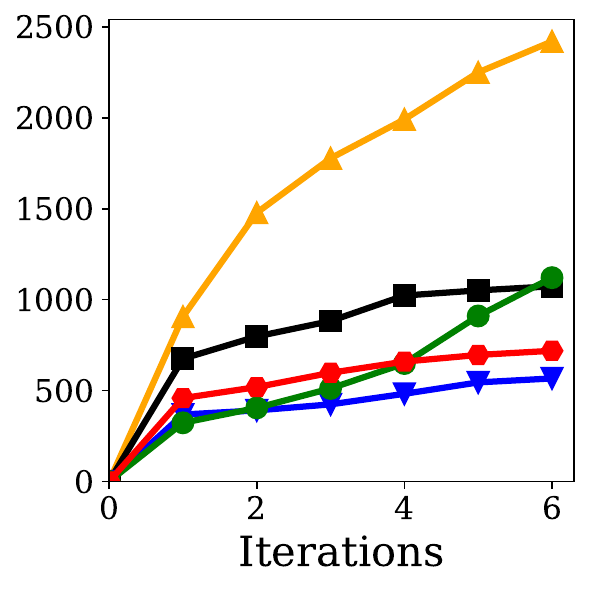}
        \subcaption{Pokemon, FID.}
        \label{figure:metric:pokemon:fid}
    \end{subfigure}
    \begin{subfigure}[b]{0.22\textwidth}
        \centering
        \includegraphics[width=1.0\textwidth]{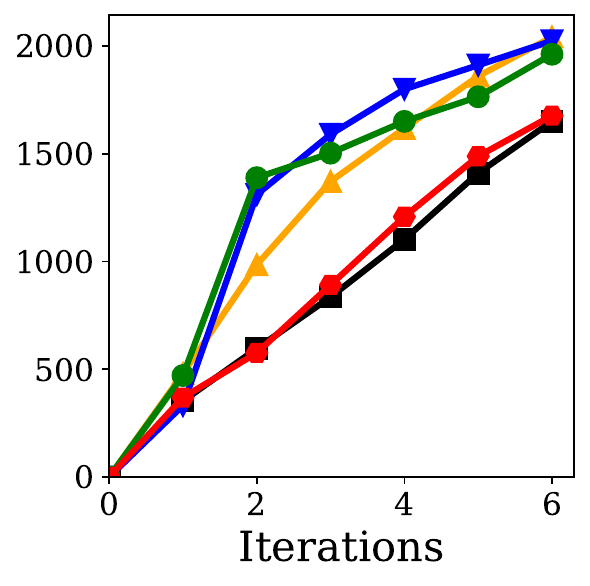}
        \subcaption{CelebA-1k, FID.}
        \label{figure:metric:celeba:fid}
    \end{subfigure}
    \begin{subfigure}[b]{0.22\textwidth}
        \centering
        \includegraphics[width=1.0\textwidth]{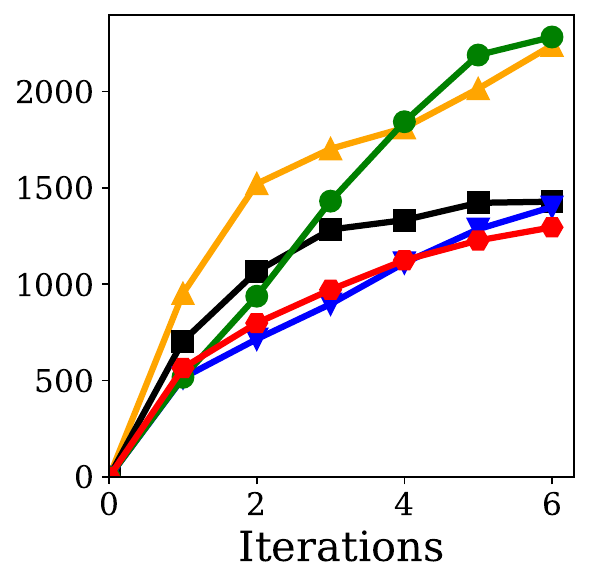}
        \subcaption{Kumapi, FID.}
        \label{figure:metric:kumapi:fid}
    \end{subfigure}
    \begin{subfigure}[b]{0.22\textwidth}
        \centering
        \includegraphics[width=1.0\textwidth]{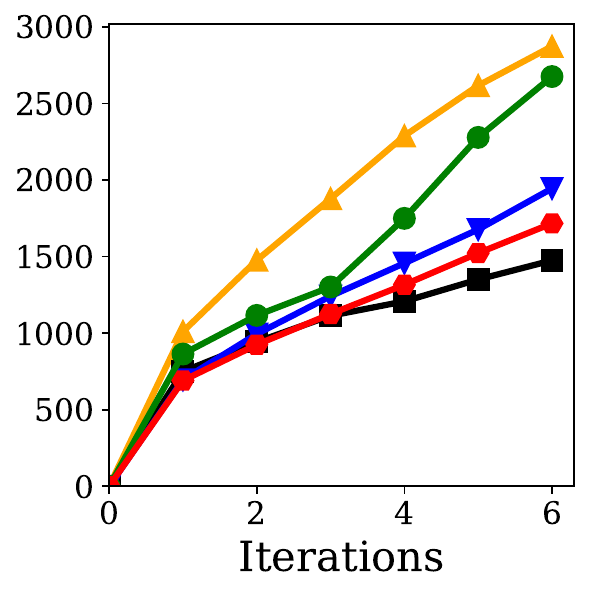}
        \subcaption{Butterfly, FID.}
        \label{figure:metric:butterfly:fid}
    \end{subfigure}
    \hfill
    \begin{subfigure}[b]{0.225\textwidth}
        \centering
        \includegraphics[width=1.0\textwidth]{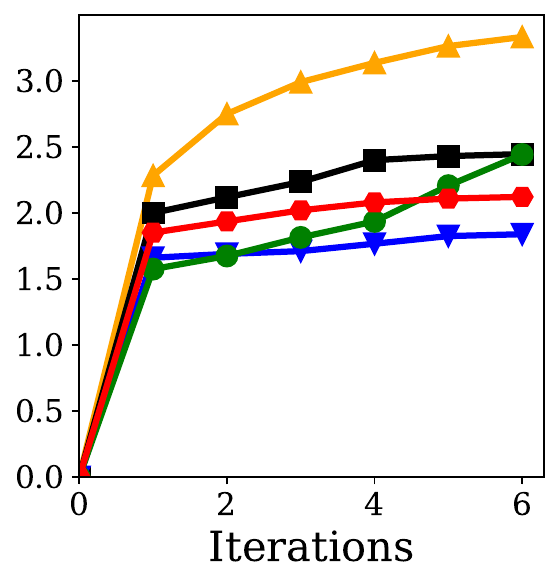}
        \subcaption{Pokemon, SFD.}
        \label{figure:metric:pokemon:sfd}
    \end{subfigure}
    \begin{subfigure}[b]{0.215\textwidth}
        \centering
        \includegraphics[width=1.0\textwidth]{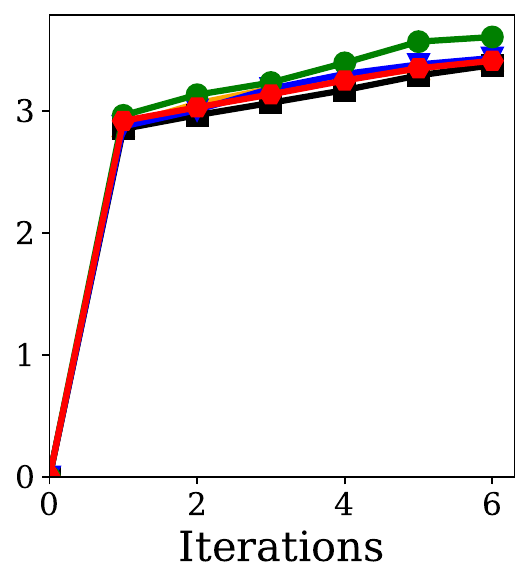}
        \subcaption{CelebA-1k, SFD.}
        \label{figure:metric:celeba:sfd}
    \end{subfigure}
    \begin{subfigure}[b]{0.225\textwidth}
        \centering
        \includegraphics[width=1.0\textwidth]{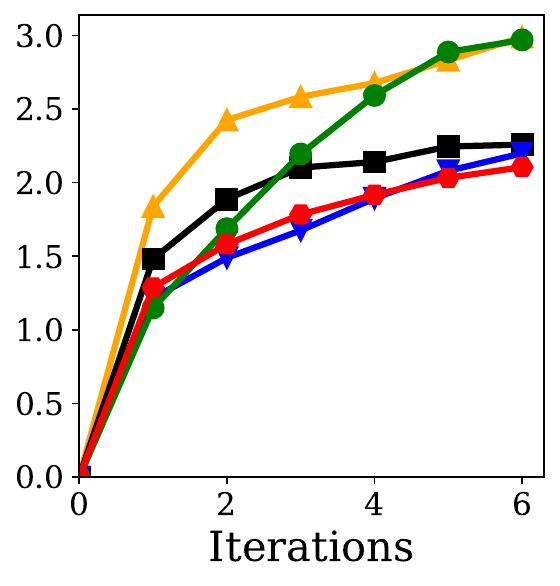}
        \subcaption{Kumapi, SFD.}
        \label{figure:metric:kumapi:sfd}
    \end{subfigure}
    \begin{subfigure}[b]{0.215\textwidth}
        \centering
        \includegraphics[width=1.0\textwidth]{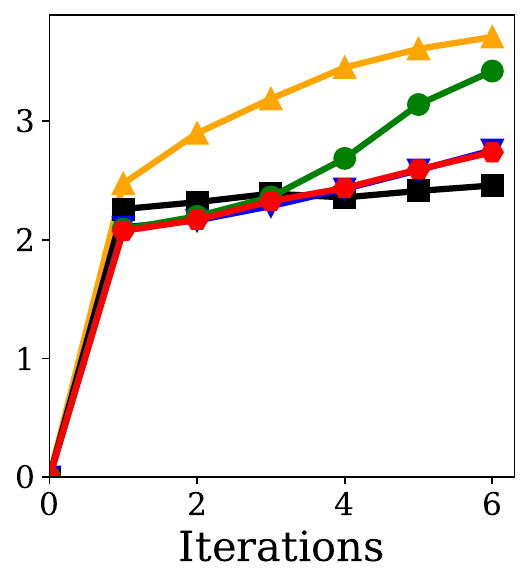}
        \subcaption{Butterfly, SFD.}
        \label{figure:metric:butterfly:sfd}
    \end{subfigure}
    \hfill
    \begin{subfigure}[b]{0.225\textwidth}
        \centering
        \includegraphics[width=1.0\textwidth]{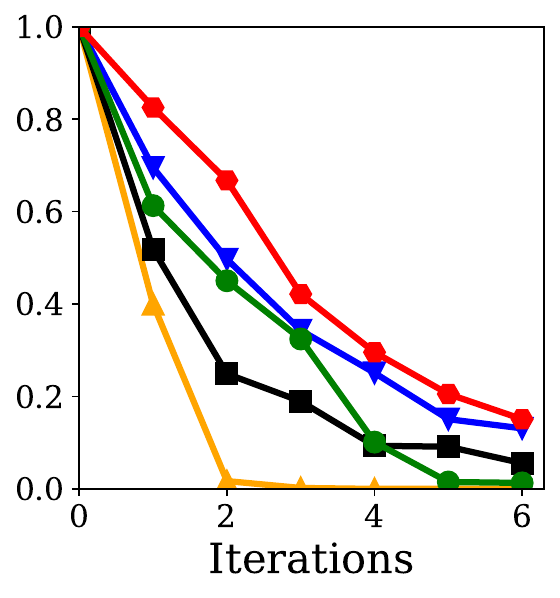}
        \subcaption{Pokemon, recall.}
        \label{figure:metric:pokemon:recall}
    \end{subfigure}
    \begin{subfigure}[b]{0.215\textwidth}
        \centering
        \includegraphics[width=1.0\textwidth]{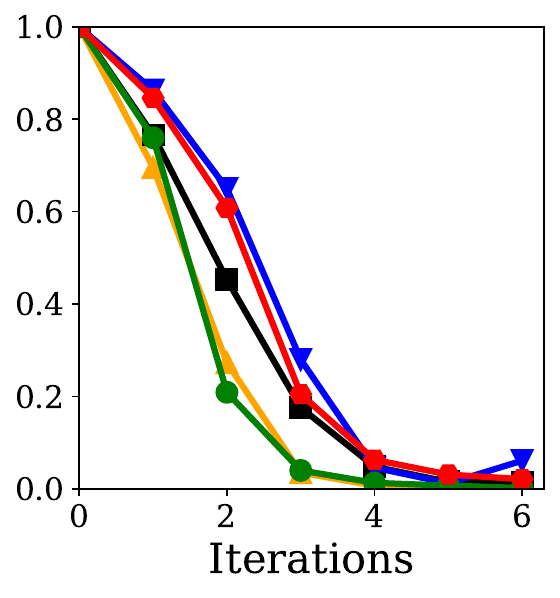}
        \subcaption{CelebA-1k, recall.}
        \label{figure:metric:celeba:recall}
    \end{subfigure}
    \begin{subfigure}[b]{0.225\textwidth}
        \centering
        \includegraphics[width=1.0\textwidth]{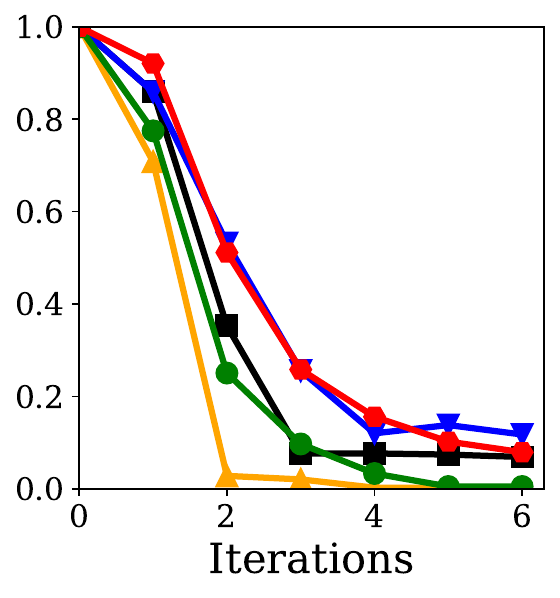}
        \subcaption{Kumapi, recall.}
        \label{figure:metric:kumapi:recall}
    \end{subfigure}
    \begin{subfigure}[b]{0.215\textwidth}
        \centering
        \includegraphics[width=1.0\textwidth]{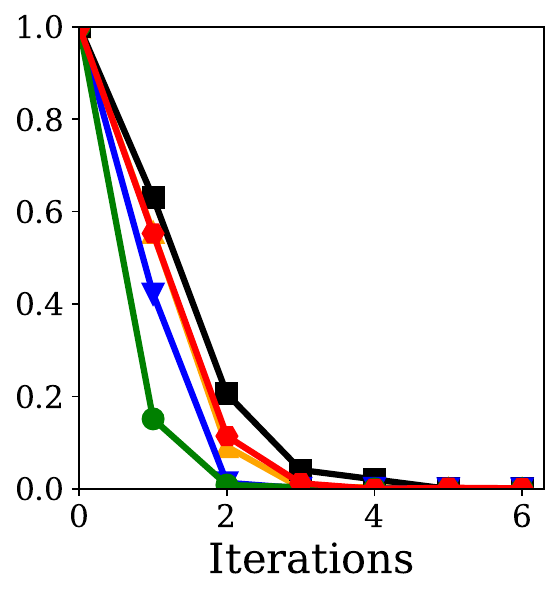}
        \subcaption{Butterfly, recall.}
        \label{figure:metric:butterfly:recall}
    \end{subfigure}
    \caption{
    Quantitative results of ReDiFine and baselines (different CFG scales).
    }
    \label{figure:metric:all}
\end{figure}

In addition to visual inspections for images generated using ReDiFine, we compare the quantitative results of ReDiFine to baselines with different CFG scales across FID, CLIP score, SFD, and recall. Details about each metric are provided in Appendix~\ref{appendix:experimental_setting:metrics}. DiNOv2 features are used to compute FID, recall, and SFD. We follow \cite{kynkaanniemi2019improved} to compute recall and set the number of neighbors for computing recall $5$. The results, shown in Figure~\ref{figure:metric:all}, demonstrate that ReDiFine achieves performance comparable to the optimal CFG scales ($2.5$ for Pokemon and Kumapi, $1.5$ for CelebA-1k and Butterfly) across different datasets and metrics.

\section{Ablation study} 
\label{appendix:ablation}

This section provides an ablation study to understand how condition drop finetuning and CFG scheduling contribute to the success of ReDiFine.

\subsection{Condition drop finetuning}
\label{appendix:ablation:condition_drop}

\begin{figure}[h!]
    \centering
    \includegraphics[width=0.7\textwidth]{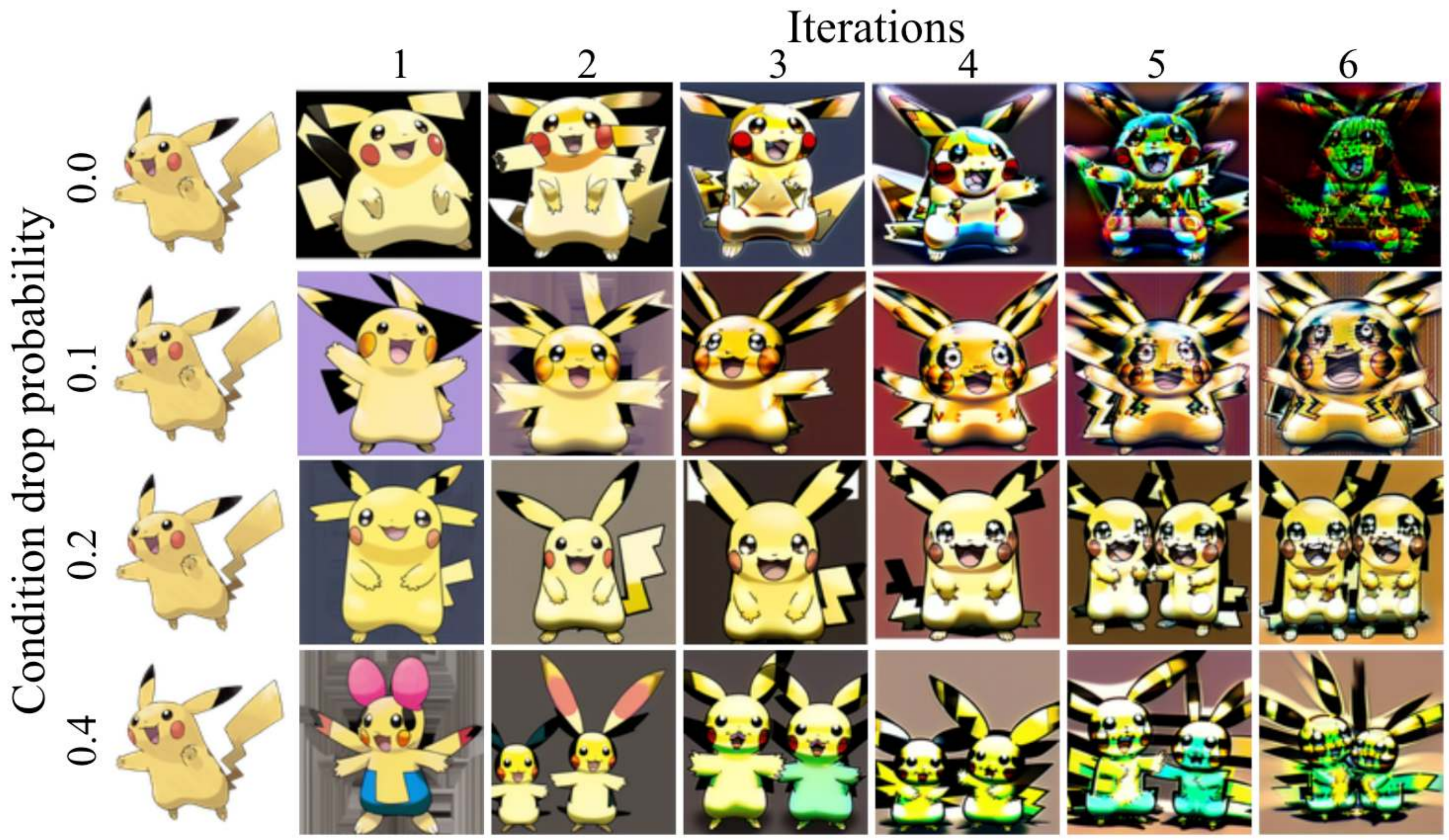}
    \caption{
    Chain of Diffusion with condition drop finetuning on Pokemon dataset.
    }
    \label{figure:ablation_experiments:condition_drop:pokemon}
\end{figure}

\begin{figure}[h!]
    \centering
    \includegraphics[width=0.7\textwidth]{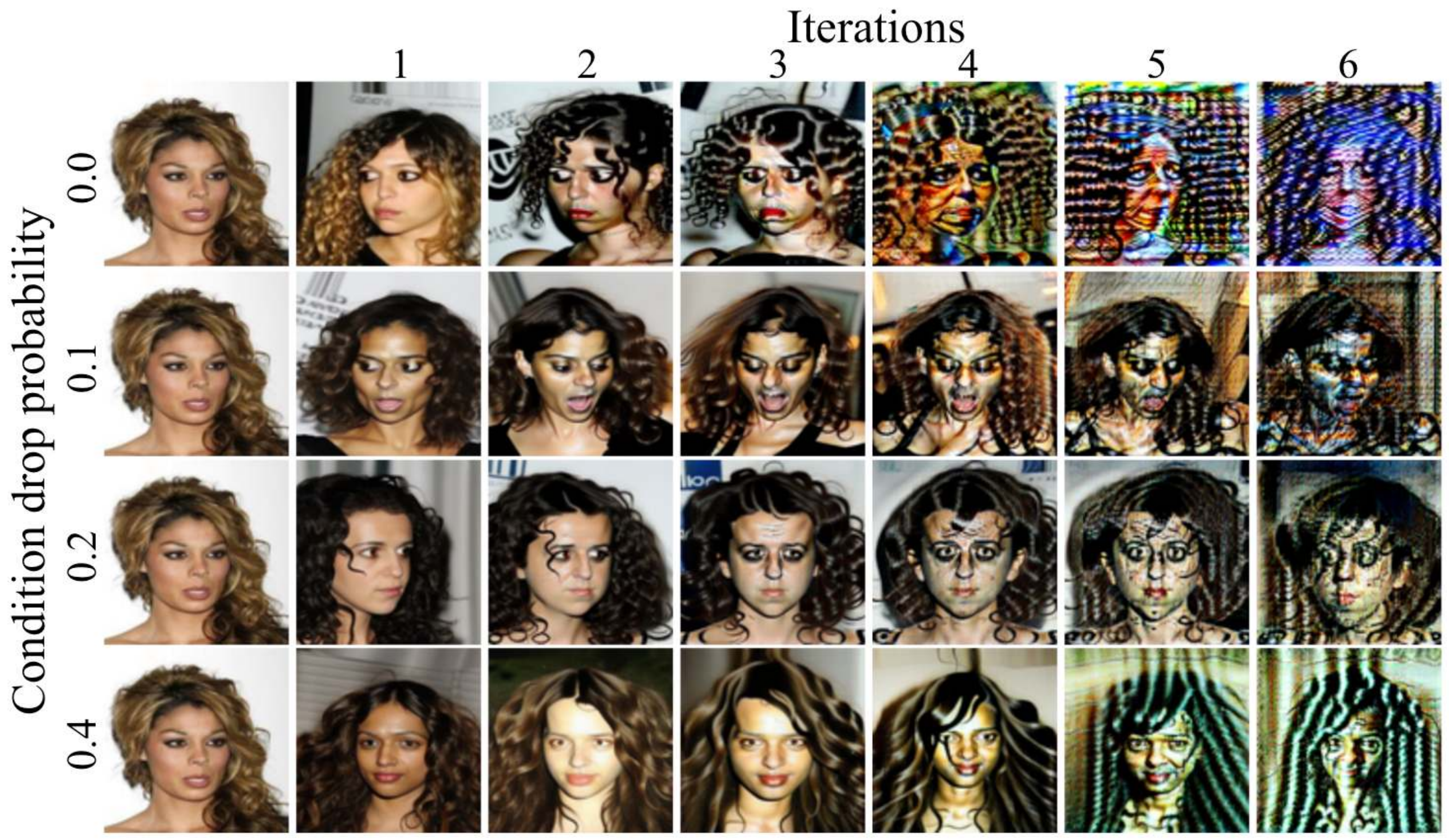}
    \caption{
    Chain of Diffusion with condition drop finetuning on CelebA-1k dataset.
    }
    \label{figure:ablation_experiments:condition_drop:celeba}
\end{figure}

We conducted an ablation study to understand how the probability of dropping text embedding during finetuning affects the image quality in the Chain of Diffusion. We examine $0.1$, $0.2$, and $0.4$ as Stable Diffusion is trained using $0.1$ or $0.2$. For both Pokemon and CelebA-1k datasets, a probability of 0.2 works the best, as shown in Figure~\ref{figure:ablation_experiments:condition_drop:pokemon} and Figure~\ref{figure:ablation_experiments:condition_drop:celeba}, respectively. Interestingly, condition drop finetuning helps to mitigate the color saturation problem, but its effect decreases with a higher probability. For both of these datasets, condition drop finetuning can mitigate image degradation to some degree, but still, there is a large quality degradation that needs to be improved.

\subsection{CFG scheduling}
\label{appendix:ablation:cfg_scheduling}

\begin{figure}[h!]
    \centering
    \includegraphics[width=0.7\textwidth]{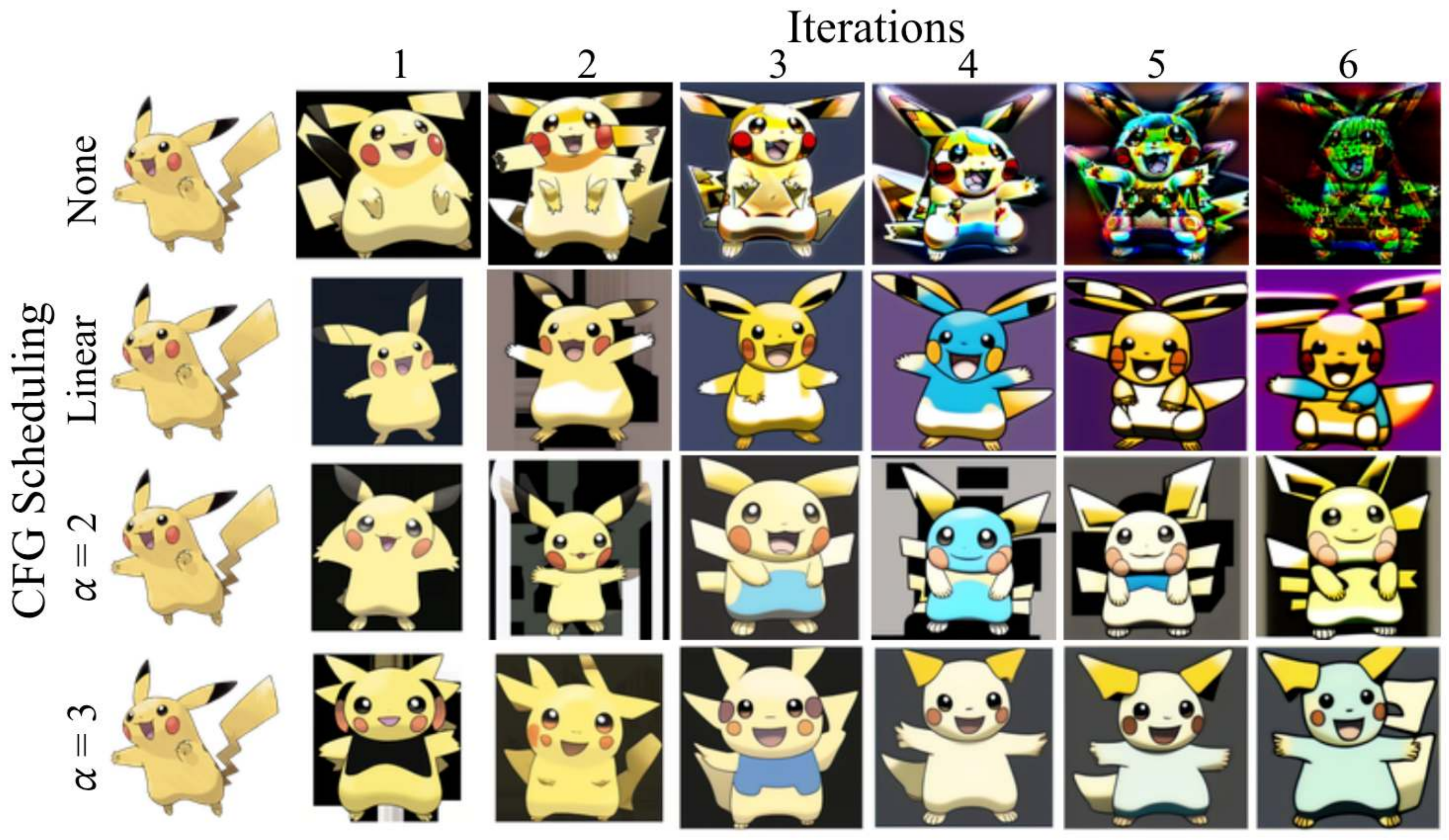}
    \caption{
    Chain of Diffusion with CFG scheduling on Pokemon dataset.
    }
    \label{figure:ablation_experiments:cfg_scheduling:pokemon}
\end{figure}

\begin{figure}[h!]
    \centering
    \includegraphics[width=0.7\textwidth]{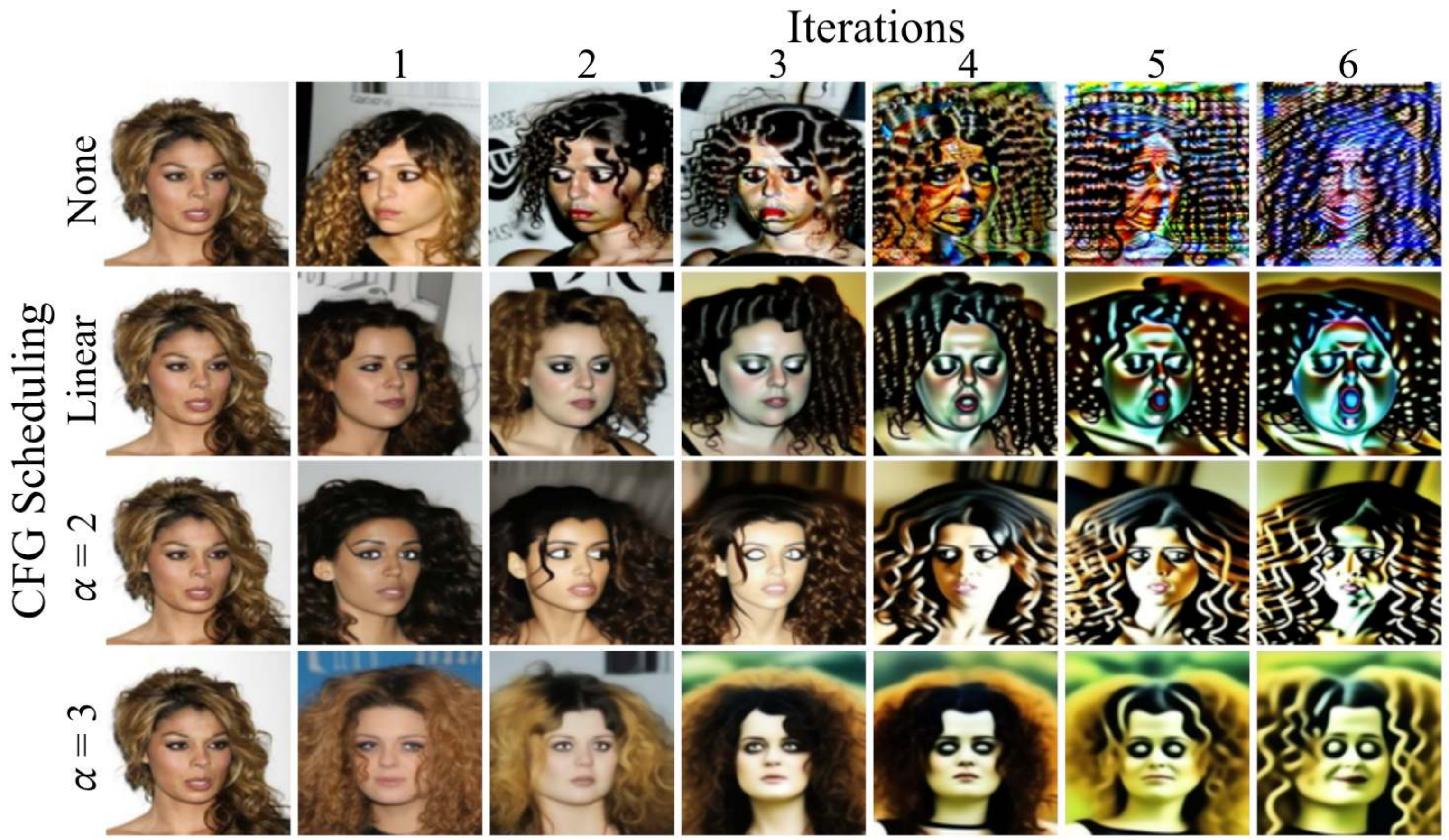}
    \caption{
    Chain of Diffusion with CFG scheduling on CelebA-1k dataset.
    }
    \label{figure:ablation_experiments:cfg_scheduling:celeba}
\end{figure}

We also evaluated how different CFG scale decreasing strategies impact image degradation in the Chain of Diffusion. We experimented with two different exponential decay rates and compared them with a linear decreasing strategy. Figure~\ref{figure:ablation_experiments:cfg_scheduling:pokemon} demonstrates that CFG scheduling is effective for Pokemon dataset, generating high-quality images comparable to those generated by ReDiFine. However, as shown in Figure~\ref{figure:ablation_experiments:cfg_scheduling:celeba}, it fails to enhance image quality on CelebA-1k dataset. This highlights the necessity of condition drop finetuning for achieving universal improvements in the Chain of Diffusion across various datasets.

\section{Analysis} 
\label{appendix:analysis}

In this section, we present a series of analyses of images generated through the Chain of Diffusion. Specifically, we examine the distribution of latent values and the differences between conditional and unconditional scores. Additionally, we analyze the power spectra of the images using 2D Fourier transforms and explore fingerprints through forensic analysis~\citep{corvi2023intriguing, corvi2023detection}.

\begin{figure}[h!]
    \centering
    \begin{subfigure}[b]{0.99\textwidth}
        \centering
        \includegraphics[width=1.0\textwidth]{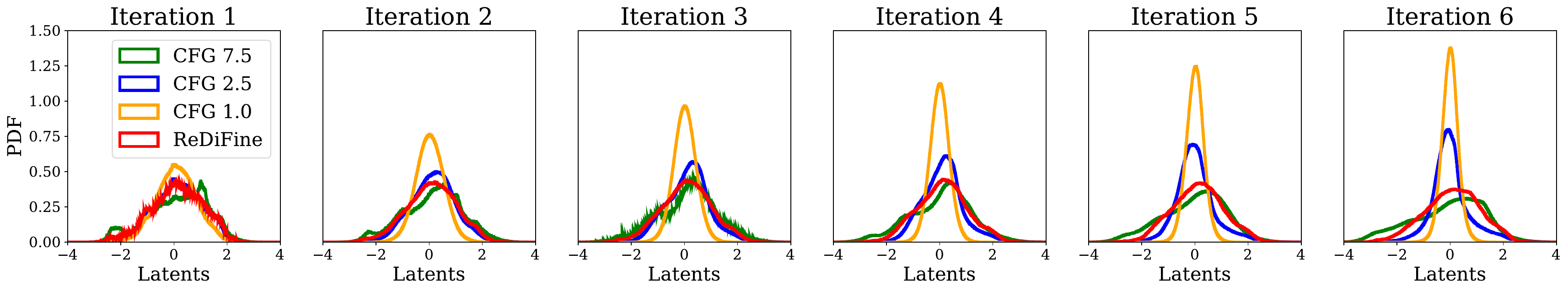}
        \subcaption{Probability density function (PDF) of values of latent vectors at the last diffusion step.}
        \label{figure:analysis:histogram}
    \end{subfigure}
    \hfill
    \begin{subfigure}[b]{0.99\textwidth}
        \centering
        \includegraphics[width=1.0\textwidth]{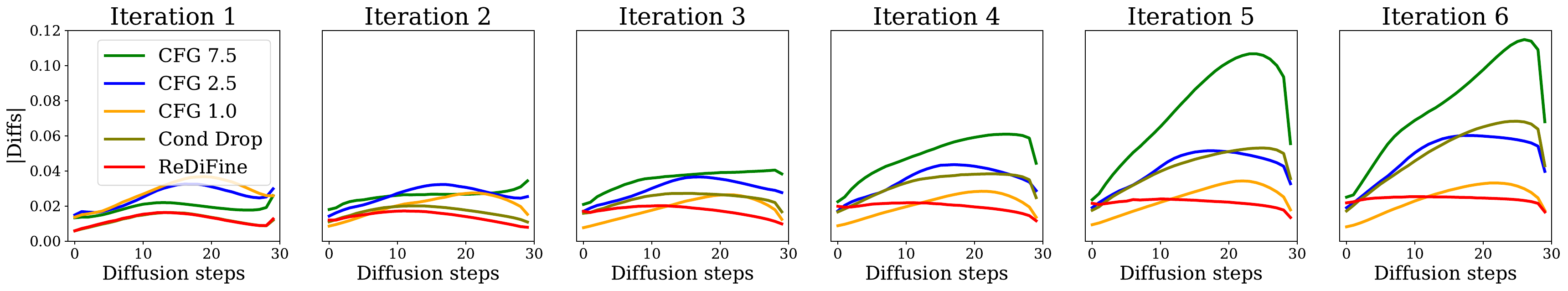}
        \subcaption{Norm of differences between conditional and unconditional scores during diffusion steps.}
        \label{figure:analysis:diffs}
    \end{subfigure}
    \caption{
    Histogram of latent values and Diffs during diffusion steps for Pokemon dataset.
    (a) 
    \textbf{Latent distribution shrinks over iteration for low CFG and expands with high CFG.}
    Larger values in latent vectors are more likely to occur with high CFG, gradually increasing the tail of the distribution.
    (b) 
    \textbf{Differences between conditional and unconditional scores increase as the training set is more degraded.}
    Especially, high differences in the later diffusion steps can be a cause of high-frequency degradation.
    }
    \label{figure:analysis:latents}
\end{figure}

\subsection{Latent analysis}
\label{appendix:analysis:latent}

Figure~\ref{figure:analysis:histogram} illustrates how the distribution of latent values evolves across different iterations. The histograms show the final latent vectors before decoding into pixel space, comparing various CFG scales and ReDiFine. For a CFG of $1.0$, the latent distribution rapidly converges into a Gaussian-like shape, with its variance shrinking over iterations. This behavior is consistent with previous work~\citep{bertrand2023stability,alemohammad2023self,dohmatob2024tale}, which theoretically predicted that the self-consuming loop progressively trims the tails of the distribution, reducing output diversity until it collapses to a single mode. We hypothesize that this narrowing in the latent space leads to blurrier, more homogeneous outputs in pixel space. Conversely, at a CFG scale of $7.5$, the latent distribution develops longer tails and tends toward a more uniform spread across space. A CFG scale of $2.5$, which demonstrates the best reusability among the three, better preserves the latent distribution over iterations. ReDiFine further enhances this preservation, maintaining the histogram from the first to the last iteration, thus achieving both high fidelity in the first iteration and better reusability.

\subsection{Differences between conditional and unconditional scores}
\label{appendix:analysis:scores}

Next, we plot the evolution of the average norm of $\text{Diff}$ ($= \text{Cond Score} - \text{Uncond Score}$) across diffusion steps for different iterations in Figure~\ref{figure:analysis:diffs}. In the first iteration, the highest Diff value is observed for CFG $1.0$, followed by CFG $2.5$ and CFG $7.5$. This behavior can be interpreted as the models' adaptive behavior to preserve the values added to the latent vectors, $\text{Diff}$ multiplied by CFG scale, at each step. However, this trend shifts in later iterations. The Diff value for CFG $7.5$ continues to grow with each iteration, and by iteration $6$, we see elevated Diff values throughout the entire diffusion steps, creating a significant gap compared to CFG $2.5$ and $1.0$. We conjecture that this accumulation of Diff is the responsible for the high-frequency degradation in images generated with CFG $7.5$. In contrast, the Diff value for CFG $1.0$ remains relatively stable or even decreases across iterations. The deviation of Diff among different iterations is minimized by ReDiFine, which explains its ability to preserve image quality in later iterations. While condition drop finetuning helps reduce the Diff in the earlier iterations, it fails to prevent accumulations in later iterations. This limitation is also evident in the ablation study, where condition drop finetuning alone was insufficient to prevent model collapse. Notably, ReDiFine produces significantly smaller Diff values compared to the baseline with CFG scale $2.5$, comparable to CFG scale $1.0$ even when using a high CFG scale $7.5$. This underscores the importance of combining condition drop finetuning with CFG scheduling.

\subsection{Power spectra of 2D Fourier transforms}
\label{appendix:analysis:spectra}

\begin{figure}[ht!]
    \centering
    \begin{subfigure}[b]{0.48\textwidth}
        \centering
        \includegraphics[width=1.0\textwidth]{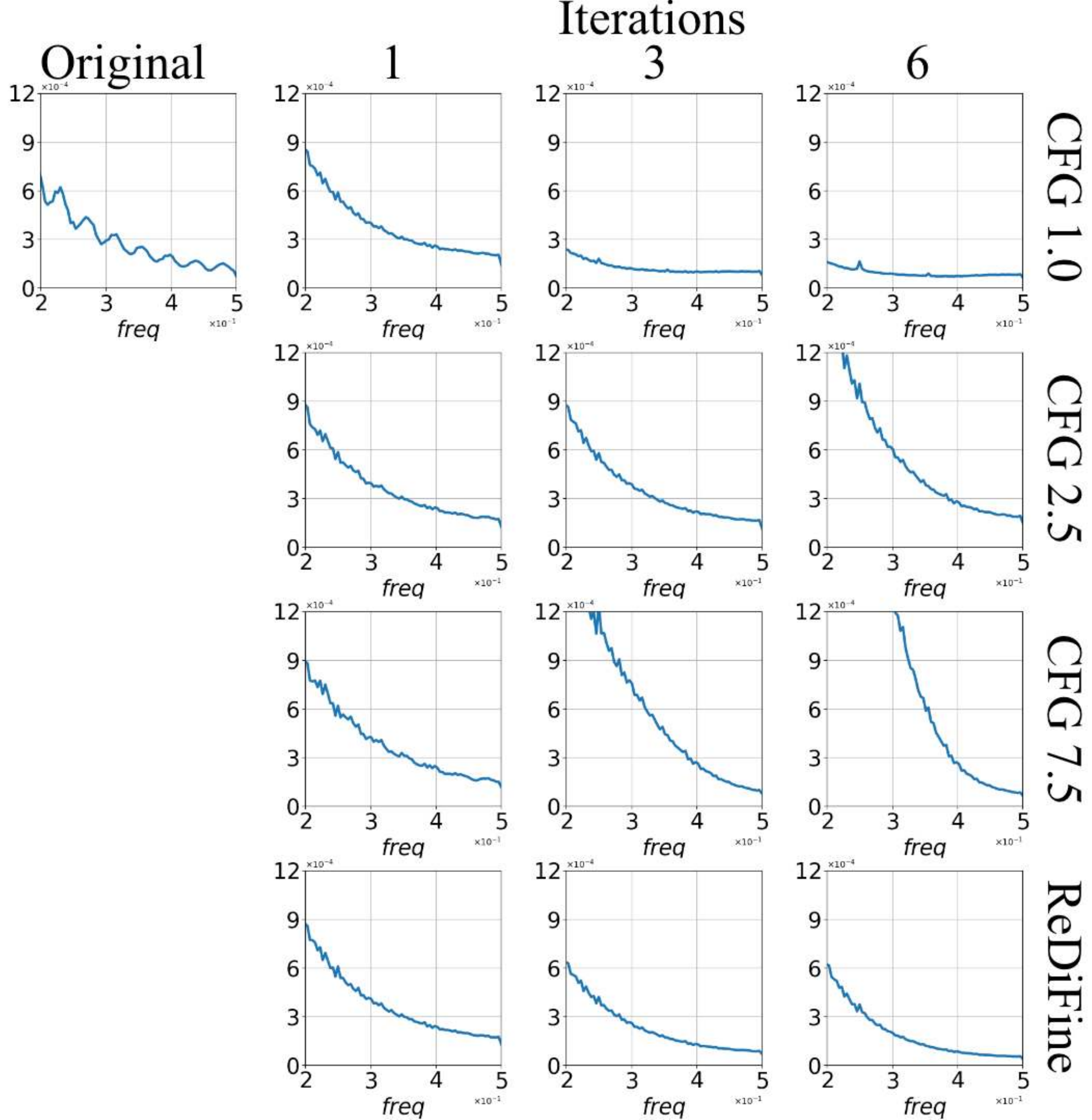}
        \subcaption{Radial spectrum power density.}
        \label{figure:analysis:radial}
    \end{subfigure}
    \hfill
    \begin{subfigure}[b]{0.50\textwidth}
        \centering
        \includegraphics[width=1.0\textwidth]{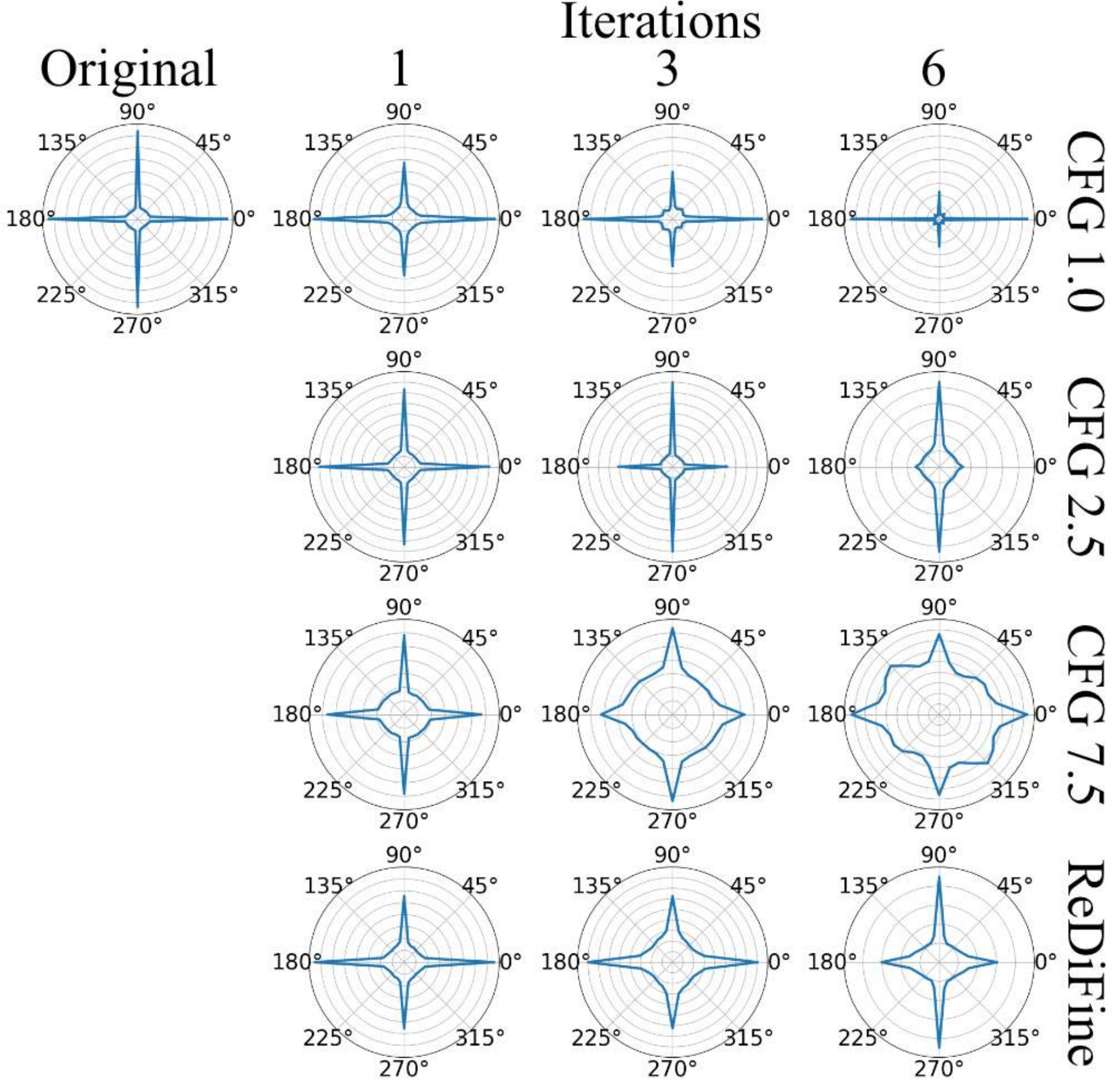}
        \subcaption{Angular spectrum power density.}
        \label{figure:analysis:angles}
    \end{subfigure}
    \caption{
    Power spectrum density of the original training set and synthetic sets for Pokemon dataset. \textbf{Images generated by ReDiFine maintain power density distribution during Chain of Diffusion while baselines fail. Even CFG scale $2.5$ cannot maintain the distribution for the last iteration.}
    (a) Radial spectrum power density. ReDiFine shows a density distribution similar to that of the original training set.
    (b) Angular spectrum power density. Power density of generated images by ReDiFine remains during the iterations while baselines cannot maintain angular distribution.
    }
    \label{figure:analysis:spectrum}
\end{figure}

Figure~\ref{figure:analysis:spectrum} demonstrates the radial and angular spectrum power density of both the original and synthetic images. It is evident that ReDiFine closely maintains the radial spectrum power density of the original training set, whereas even a CFG scale $2.5$ falls short. Additionally, ReDiFine demonstrates stable angular spectra throughout the Chain of Diffusion, even though they differ from those of the original training set. Pokemon dataset is used for power spectra analysis.

\subsection{Fingerprints for forensic analysis}
\label{appendix:analysis:fingerprint}

\begin{figure}[th!]
    \centering
    \begin{subfigure}[b]{0.49\textwidth}
        \centering
        \includegraphics[width=1.0\textwidth]{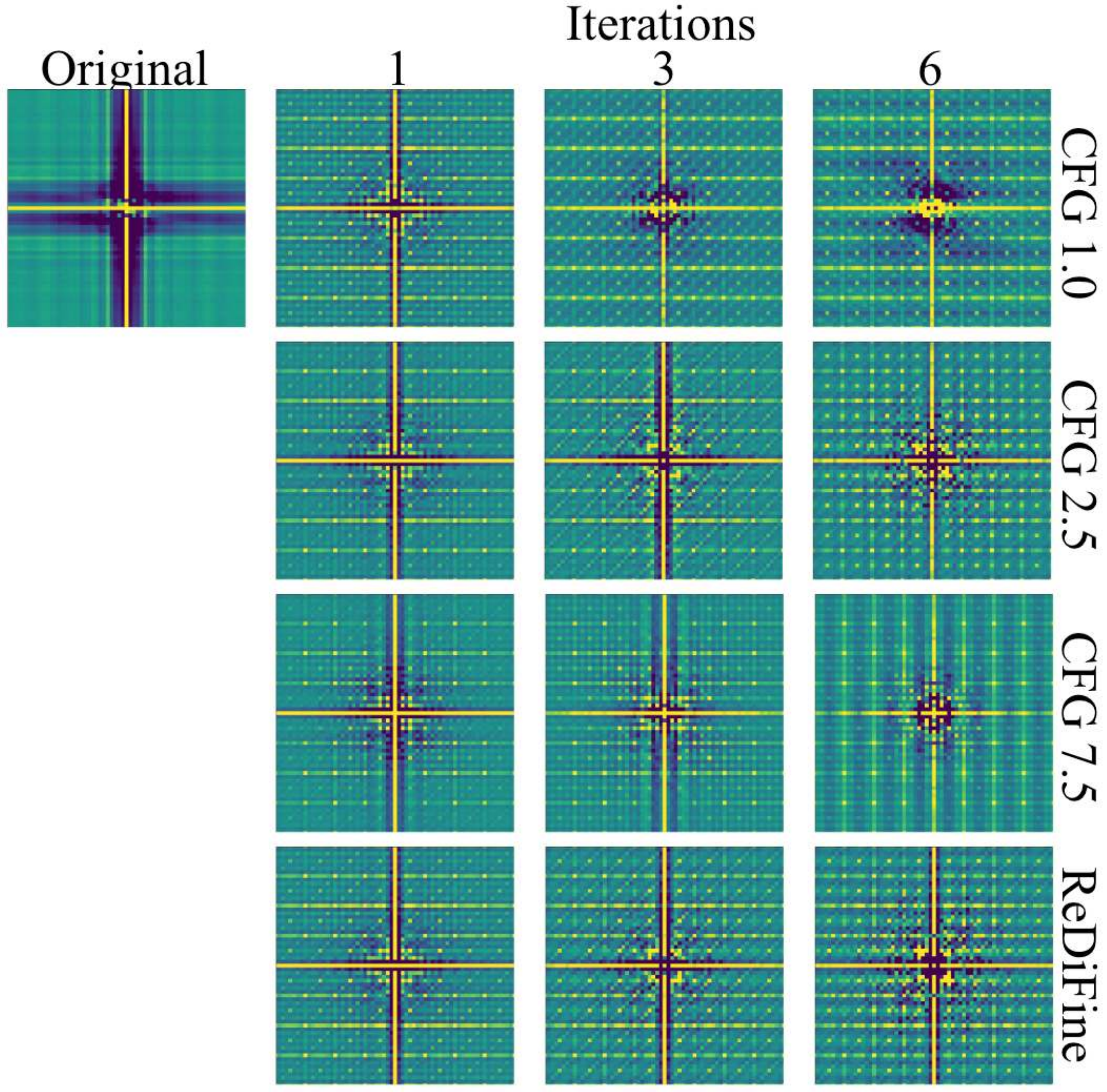}
        \subcaption{Autocorrelation.}
        \label{figure:analysis:auto}
    \end{subfigure}
    \hfill
    \begin{subfigure}[b]{0.49\textwidth}
        \centering
        \includegraphics[width=1.0\textwidth]{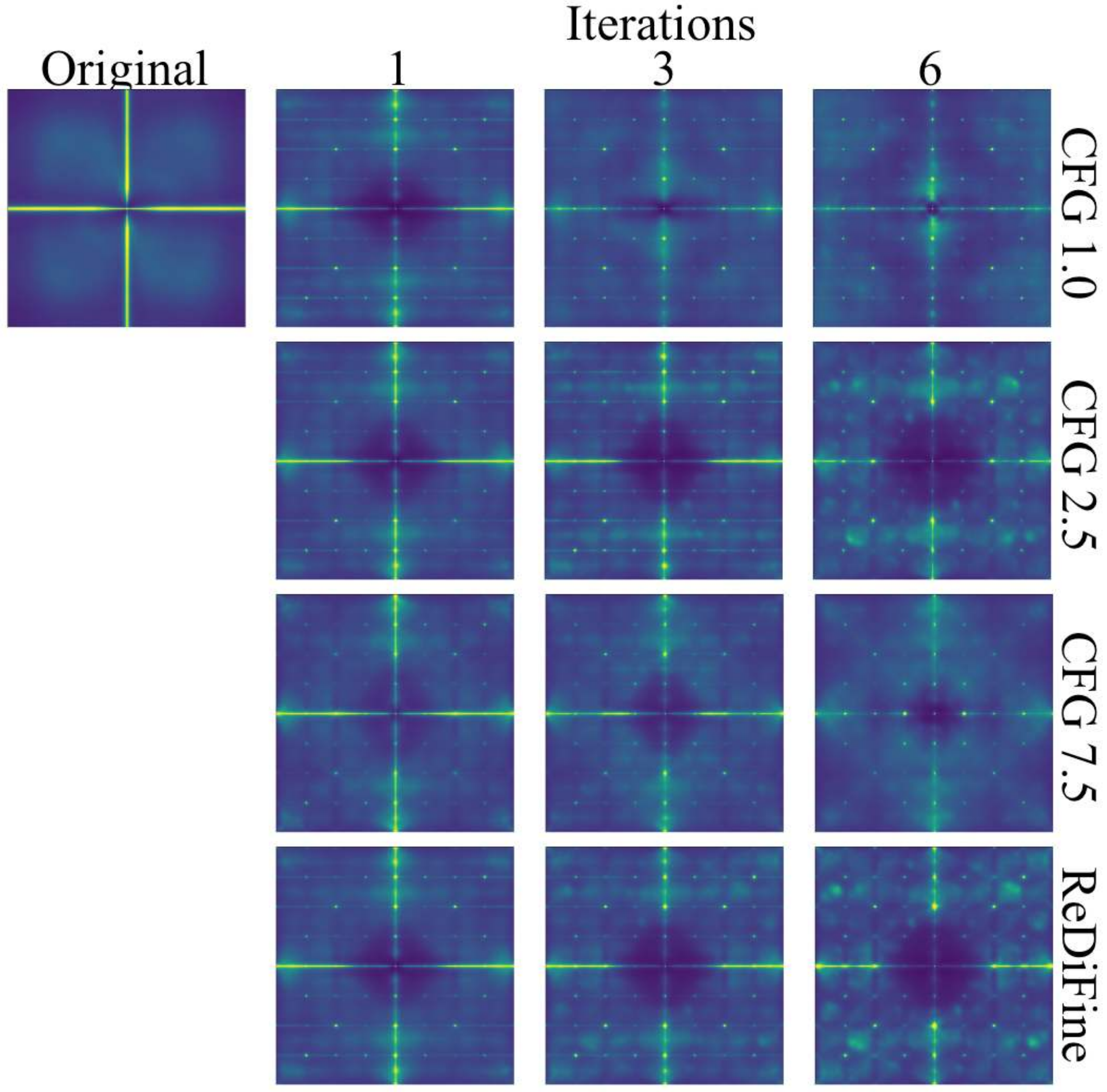}
        \subcaption{Average power spectra.}
        \label{figure:analysis:power}
    \end{subfigure}
    \caption{
    \textbf{The fingerprints of the original training set and synthetic sets show clear differences, and ReDiFine produces fingerprints similar to CFG scale 2.5.}
    (a) Autocorrelation of image fingerprints. Horizontal and vertical lines gradually disappear for CFG scales $1.0$ and $7.5$ while they are maintained for CFG scale $2.5$ and ReDiFine.
    (b) Average power spectra of images. Central regions are amplified or diminished for CFG scales $1.0$ and $7.5$, demonstrating low and high-frequency degradation.
    }
    \label{figure:analysis:fingerprints}
\end{figure}

Several works~\citep{corvi2023intriguing, corvi2023detection} aim to identify fingerprints of synthetic images. High-quality synthetic images from different generative models have clearly distinct fingerprints, showing the potential to be used for synthetic image detection. We analyze fingerprints of synthetic images for different CFG scales and iterations, and compare them to fingerprints of the original training set. Both autocorrelation and average power spectra show clear differences between the original training set and synthetic images, as shown in Figure~\ref{figure:analysis:fingerprints}. Moreover, how the fingerprints of synthetic images evolve throughout the Chain of Diffusion differ for ReDiFine and different CFG scales. Specifically, fingerprints of synthetic images from ReDiFine are similar to those of images from CFG scale $2.5$, while other CFG scales ($1.0$ and $7.5$) make fingerprints different from the first iteration as iterations proceed. Horizontal and vertical lines in autocorrelation gradually disappear and central regions in power spectra vary for further iterations. Also, the varying central regions in power spectra imply that low frequency features increase and decrease for CFG scale $1.0$ and $7.5$, respectively, aligning with visual inspections. Generating images with fingerprints similar to those of the original real images can be an interesting future direction to reduce the degradation in the Chain of Diffusion. Pokemon dataset is used for fingerprint analysis.


\end{document}